\documentclass[3p]{elsarticle}

\usepackage{hyperref}
\usepackage{amsmath}
\usepackage{amssymb}
\usepackage{float}
\usepackage{enumitem}
\usepackage{algorithm}
\usepackage{algpseudocode}
\usepackage{dirtytalk}
\usepackage{mathtools}
\usepackage{xcolor}
\usepackage{amsthm}
\usepackage{bm,natbib}
\usepackage[normalem]{ulem}

\newcommand{\R}{\mathbb{R}}
\newcommand{\ssymbol}[1]{^{\@fnsymbol{#1}}}
\newcommand{\norm}[1]{\|#1\|}

\newtheorem*{remark}{Remark}

\numberwithin{equation}{section}

\journal{Journal of Elsevier}

\begin{document}

\begin{frontmatter}

\title{Sparse Identification of Nonlinear Dynamical Systems via Reweighted $\ell_1$-regularized Least Squares}


\author[colorado1]{Alexandre Cortiella}
\ead{alexandre.cortiella@colorado.edu}
\author[colorado1]{Kwang-Chun Park}
\ead{kcpark@colorado.edu}
\author[colorado1]{Alireza Doostan\corref{mycorrespondingauthor}}
\cortext[mycorrespondingauthor]{Corresponding author}
\ead{alireza.doostan@colorado.edu}
\address[colorado1]{Smead Aerospace Engineering Sciences Department, University of Colorado, Boulder, CO 80309, USA}

\begin{abstract}
This work proposes an iterative sparse-regularized regression method to recover governing equations of nonlinear dynamical systems from noisy state measurements. The method is inspired by the Sparse Identification of Nonlinear Dynamics (SINDy) approach of {\it [Brunton et al., PNAS, 113 (15) (2016) 3932-3937]}, which relies on two main assumptions: the state variables are known {\it a priori} and the governing equations lend themselves to sparse, linear expansions in a (nonlinear) basis of the state variables. The aim of this work is to improve the accuracy and robustness of SINDy in the presence of state measurement noise. To this end, a reweighted $\ell_1$-regularized least squares solver is developed, wherein the regularization parameter is selected from the corner point of a Pareto curve. The idea behind using weighted $\ell_1$-norm for regularization -- instead of the standard $\ell_1$-norm  -- is to better promote sparsity in the recovery of the governing equations and, in turn, mitigate the effect of noise in the state variables. We also present a method to recover single physical constraints from state measurements. Through several examples of well-known nonlinear dynamical systems, we demonstrate empirically the accuracy and robustness of the reweighted $\ell_1$-regularized least squares strategy with respect to state measurement noise, thus illustrating its viability for a wide range of potential applications.
\end{abstract}


\end{frontmatter}


\section{Introduction} 
\label{sec:introduction}
Several branches in science and engineering represent systems that change over time as a set of differential equations that govern the underlying physical behavior. The structure of these differential equations is usually determined by observing the system and inferring relationships between variables, or derived from fundamental axioms and mathematical reasoning. Examples
of the empirical method include Johannes Kepler and Isaac Newton's approaches in deriving laws of planetary motion. The accurate measurements of planet trajectories by Tycho Brahe enabled Kepler to empirically determine the laws that govern the motion of elliptic orbits. Newton, in turn, was able to derive the law of universal gravitation by inductive reasoning. Solving models derived from fundamental laws, either analytically or numerically, has proven to be a useful approach in engineering to produce reliable systems. However, the derived models often rely on simplifying assumptions that may not explain complex phenomena, leading to a mismatch between predictions and observations. Moreover, parameters of these models may need to be estimated indirectly from system observables.

Recent advances in data acquisition systems along with modern data science techniques have fostered the development of accurate data-driven approaches, such as inverse modeling and system identification, in modeling physical or biological systems~\cite{Fassois2007}. In particular, system identification, i.e., deducting accurate mathematical models from measured observations, has experienced significant advances in recent years~\cite{ljung2001system,ljung2010perspectives}. System identification is a key to improved understanding of complex phenomena, dominant feature analysis, design of experiments, and system monitoring and control. Some promising applications include space weather modeling~\cite{balikhin2011using,boynton2013analysis}, development of material constitutive laws~\cite{mahnken2017identification}, turbulence modeling~\cite{brunton2015closed}, and model predictive control~\cite{morari1999model}, to name a few. 

In dynamical systems where the underlying physics is not fully understood or simplified for the interest of computational efficiency, system identification may be used to both determine a model structure, e.g., the governing equations, and estimate model parameters from data. When the model structure is unknown, one particular approach that has received increasing attention is to approximate the nonlinear dynamics in an over-complete basis of the state variables and eliminate the expansion terms that do not contribute to the dynamics. Examples of this approach include polynomial NARMAX~\cite{Leontaritis1985}, symbolic polynomial regression~\cite{Schmidt81,Bongard2007}, and sparse (polynomial) regression~\cite{Wang2011,Brunton2016} dubbed SINDy in~\cite{Brunton2016}. 

System identification via sparse (polynomial) regression employs techniques from compressed sensing -- specifically regularization via sparsity promoting norms, such as $\ell_0$- and $\ell_1$-norms, to identify an {\it a priori} unknown subset of the basis describing the dynamics. The idea behind sparsity-promoting norms is based on Occam's razor principle, also known as the law of parsimony, which states: \say{Of two competing theories, the simpler explanation of an entity is preferred}~\cite{OccamsRazor}. The identified model may then be further analyzed to understand the physical behavior of the dynamical system, and can be integrated in time to predict future state variables of the system. Least Absolute Shrinkage and Selection Operator (LASSO) \cite{tibshirani1996regression}, Least Angle Regression (LARS) \cite{Efron2004}, Sequentially Thresholded Least Squares (STLS)~\cite{Brunton2016}, and Basis Pursuit Denoising (BPDN) \cite{Chen2001} are some sparsity promoting algorithms that may be used for model recovery. In particular, the work by Wang et al.~\cite{Wang2011} proposed a compressive sensing approach to polynomial basis expansion with $\ell_1$-minimization of the coefficients in order to recover nonlinear dynamical systems. Similarly, Brunton et al.~\cite{Brunton2016} proposed SINDy to recover the governing equations of dynamical systems. In their approach, an overdetermined system of equations is solved using a Sequentially Thresholded Least Squares scheme -- instead of $\ell_1$-minimization -- to enforce sparsity in the expansion coefficient and mitigate the impact of state measurement noise.

A major challenge in model recovery via SINDy is the identification of accurate models when the input data, i.e., state measurements or their time derivatives, are corrupted with noise. Noisy measurements may lead to the identification of incorrect basis terms and inaccurate estimation of the parameters of the model. While algorithms based on standard $\ell_1$-regularization or thresholded least squares lead to accurate recovery of the governing equations for small noise levels, they may produce inaccurate results or wrong sparsity patterns as the noise in the data increases \cite{zhang2018robust,kaheman2020sindy}. An exception is a class of chaotic dynamical systems, e.g., Lorenz 63, satisfying certain ergodicity property \cite{tran2017exact}. In these cases, as shown in \cite{tran2017exact}, the governing equation can be recovered exactly even when the state data is highly noisy and a large fraction of data is corrupted with outliers.  

\subsection{Contribution of this work}
\label{sec:contribution}

The focus of this article is to improve the accuracy and robustness of SINDy in the presence of state measurement noise. We advocate the utility of a weighted $\ell_1$-norm to regularize SINDy's regression problem. In the fields of sparse regression and compressed sensing, weighted $\ell_1$-minimization has been shown to empirically outperform $\ell_1$-minimization in recovering sparse solutions to certain under-determined linear systems~\cite{zou2006adaptive,Candes2008}. Similar observations have been made in approximating multivariate functions in orthogonal polynomial bases~\cite{Yang13,Peng14,adcock2017infinite}. The weights can be generated iteratively and inversely proportional to the values of the coefficients from the previous iteration \cite{Candes2008,Yang13}, or based an approximate value of the coefficients \cite{Peng14} or the $\ell_\infty$-norm of the basis functions~\cite{adcock2017infinite}. In this work, we adopt a reweighted version of the Basis Pursuit Denoising algorithm (WBPDN) to solve an overdetermined system with the aim of mitigating the effects of noise, and thereby recover the governing equations of dynamical systems more accurately. In WBPDN, the sparsity of the recovered model and its accuracy in generating the training data, specifically, the time derivatives of the state variables are balanced via a regularization parameter. We propose the selection of the regularization parameter based on the corner point of a Pareto curve. Additionally, we present a method based on low-rank matrix factorization via interpolative decomposition \cite{Cheng2005} to identify single constraints of dynamical systems, such as conservation of total energy, from state measurements that are polluted with low noise levels. Beyond learning constraints, the identification and removal of redundant information introduced by the constraints improve the conditioning of SINDy's regression problem. 

A related approach to improving the accuracy and robustness of SINDy is the utility of data associated with short trajectories of the state variables \cite{schaeffer2018extracting,wu2019numerical}. The trajectories correspond to multiple (random) initializations of the system. In the present study, however, we assume we have access to only single, but long, trajectories of the system state variables, an assumption that is more relevant when state measurements are obtained experimentally. 

We begin, in the next section, by presenting a background on recovering dynamical system equations from state measurements using sparsity promoting regression techniques. In Section \ref{sec:approach}, we present the WBPDN approach along with discussions on the selection of regularization parameter and the calculation of time derivatives of state variables from their noisy measurements. In Section \ref{subsec:constrainedsystems}, we introduce our approach for the identification of single constraints from state data. In Section \ref{sec:numericalexamples}, the performance of the WBPDN method is assessed through several numerical examples of well-known dynamical systems with distinct features. Finally, in Section \ref{sec:conclusion}, we draw conclusions and discuss relevant aspects of the proposed method, and provide directions for future improvement.
\section{Problem Statement and Background}
\label{sec:statement}

Throughout this work, we assume that a dynamical system has the form
\begin{equation}
    \dot{\mathbf{x}}(t) = \frac{\mathrm{d}\mathbf{x}(t)}{\mathrm{d}t} = \mathbf{f}(\mathbf{x}(t)), \quad \mathbf{x}(0) = \mathbf{x}_0,\label{eq:DynamicalSystem}
\end{equation}
where $\mathbf{x}(t)\in\R^n$ are the known and measurable state variables of the system at time $t \in [0,T]$ and $\mathbf{f}(\mathbf{x}(t))\colon \R^n \to \R^n$ is a state-dependent unknown  vector that describes the motion of the system. An important observation is that in many systems $\mathbf{f}(\mathbf{x}):=\mathbf{f}(\mathbf{x}(t))$ is a {\it simple} function of the state variables $\mathbf{x}:=\mathbf{x}(t)$ in that only a small set of state-dependent quantities, e.g., damping or inertial forces, contribute to the dynamics. Given that $\mathbf{f}(\mathbf{x})$ is unknown and following~\cite{Wang2011, Brunton2016}, we assume that each state dynamics $\dot{x}_j:=\dot{x}_j(t)$ or, equivalently, $f_j(\mathbf{x})$, $j = 1,\dots,n$, is spanned by a set of $p$ candidate nonlinear (in the state variables) basis functions $\phi_{i}(\mathbf{x})$ weighted by unknown coefficients $\xi_{ji}$,
\begin{equation}\label{eq:dynExpansion}
\dot{x}_j = \sum_{i = 1}^p \xi_{ij} \phi_{i}(\mathbf{x}),\,\,\,\,\,j = 1,\dots,n.
\end{equation}
As the true dynamics may be described by only a subset of the the considered basis $\{\phi_{i}(\mathbf{x})\}$, the unknown coefficients $\xi_{ij}$ are sparse. Exploiting this sparsity in identifying $\mathbf{f}(\mathbf{x})$ is the key idea behind SINDy algorithms. The selection of the basis is crucial as the true $\dot{\mathbf{x}}_j$, while unknown, are assumed to be either exactly or approximately in the span of the basis. For arbitrary dynamical systems, such a selection is non-trivial and physical insight or a trial and error approach must be exercised. One approach to ease the difficulty of basis selection is to build an overcomplete basis, perhaps by concatenating different types of basis, e.g., polynomials and harmonics. However, this may negatively affect the accuracy of the computed $\xi_{ij}$, especially in the presence of large levels of state measurement noise. 

To determine the governing equations $\mathbf{f}(\mathbf{x})$ via (\ref{eq:dynExpansion}), we assume that the state variables $\mathbf{x}$ are known and can be measured at discrete times $t_k$, $k = 1,\dots,m$, where $m$ is the number of measurements. Hence, (\ref{eq:dynExpansion}) may be written in matrix form as
\begin{equation}\label{eq:matrix_main_system}
    \dot{\mathbf{x}}_j = \bm{\Phi}(\mathbf{x})\bm{\xi}_j ,\,\,\,\,\,j = 1,\dots,n,
\end{equation}
where,
\begin{gather*}
\dot{\mathbf{x}}_j = 
    \begin{bmatrix}
    \dot{x}_j(t_1), & \dot{x}_j(t_2), & \hdots \,, &\dot{x}_j(t_m)
    \end{bmatrix}^T \in \R^{m};\\ \\
    \bm{\Phi}(\mathbf{x}) =
    \begin{bmatrix}
    \phi_{1}(\mathbf{x}(t_1)) & \phi_{2}(\mathbf{x}(t_1)) & \hdots & \phi_{p}(\mathbf{x}(t_1))\\
    \phi_{1}(\mathbf{x}(t_2)) & \phi_{2}(\mathbf{x}(t_2)) & \hdots & \phi_{p}(\mathbf{x}(t_2))\\
    \vdots         & \vdots         & \ddots & \vdots\\
    \phi_{1}(\mathbf{x}(t_m)) & \phi_{2}(\mathbf{x}(t_m)) & \hdots & \phi_{p}(\mathbf{x}(t_m))\\
    \end{bmatrix} \in \R^{m \times p};\,\,\,\,\text{and}\\ \\
    \bm{\xi}_j =
    \begin{bmatrix}
    \xi_{1j}, & \xi_{2j}, & \hdots \,, & \xi_{pj}
    \end{bmatrix}^T \in \R^{p}.
\end{gather*}
Hereafter, we refer to $\bm{\Phi}(\mathbf{x})$ as the {\it measurement} matrix. As we shall describe in Section~\ref{subsec:numericaldifferentiation}, we estimate the dynamics $\dot{\mathbf{x}}_j$ in (\ref{eq:matrix_main_system}) via time derivatives of the state variables $x_j$, which may require access to a large number of state measurements. Therefore, the present work focuses on over-determined systems (\ref{eq:matrix_main_system}), where the number of measurements is larger than the number of candidate functions, i.e., $m > p$. This assumption may be relaxed when $\dot{\mathbf{x}}_j$ is directly measured. For the interest of a simpler notation, we henceforth drop the subscript $j$ from $\dot{\mathbf{x}}_j$ and $\bm{\xi}_j$ in (\ref{eq:matrix_main_system}). Unless otherwise stated, $\dot{\mathbf{x}}$ refers to the measurements of $\dot{x}_j$ and not the dynamics $\dot{\mathbf{x}}$ in (\ref{eq:DynamicalSystem}).

The coefficients $\bm{\xi}$ are computed from (\ref{eq:matrix_main_system}), for each $j$, subject to a sparsity constraint on $\bm{\xi}$. Wang et al. \cite{Wang2011} achieve this via $\ell_1$-minimization or basis pursuit denoising (BPDN), a widely-used compressed sensing technique, 
\begin{equation}\label{eq:L1min}
   \min_{\bm{\xi}}\norm{\bm{\xi}}_1 \quad \text{subject to} \quad \norm{\bm{\Phi}(\mathbf{x})\bm{\xi} - \dot{\mathbf{x}}}_2 \leq \delta,
\end{equation}
where $\Vert\bm{\xi}\Vert_1=\sum_{i=1}^p\vert\xi_i\vert$ is the $\ell_1$-norm of $\bm\xi$ and $\delta\ge 0$ is some tolerance parameter to avoid over-fitting. The unconstrained formulation of (\ref{eq:L1min}) is given by the second order cone program,
\begin{equation}\label{eq:BPDN}
\text{(BPDN)}\qquad \min_{\bm{\xi}} \norm{\bm{\Phi}(\mathbf{x})\bm{\xi} - \dot{\mathbf{x}}}_2^2 + \lambda \norm{\bm{\xi}}_1,
\end{equation}
which coincides with the unconstrained LASSO~\cite{tibshirani1996regression}, and is also referred to as $\ell_1$-regularized least squares \cite{Kim2007}. In (\ref{eq:BPDN}), the regularization parameter $\lambda>0$ creates a trade-off between the accuracy of satisfying (\ref{eq:matrix_main_system}) and the sparsity of the solution. 

The SINDy algorithm of~\cite{Brunton2016} proposes a Sequentially Thresholded Least Squares (STLS) algorithm, which iteratively solves a least squares regression problem and hard-thresholds the coefficients to promote sparsity and thereby regularize the regression problem. The procedure is repeated on the non-zero entries of $\bm{\mathbf{\xi}}$ until the solution converges or the algorithm reaches a maximum number of iterations. In more details, let $\mathcal{S}(\bm\xi):=\{i:\ \xi_i\neq 0\}$ denote the support of an instance of $\bm\xi$. At the $(k+1)$th iteration of STLS, $\bm\xi^{(k+1)}$ is computed from a least squares problem over $\mathcal{S}(\bm\xi^{(k)})$ and its components smaller than some threshold parameter $\gamma>0$ are set to zero,
\begin{align}\label{eq:STLS}
\text{(STLS)}\qquad    &\bm\xi^{(k+1)}\longleftarrow\underset{\bm\xi}{\mathrm{argmin}}\left\{\norm{\bm{\Phi}(\mathbf{x})\bm{\xi} - \dot{\mathbf{x}}}_2^2 \quad \text{subject to} \quad \mathcal{S}(\bm\xi) = \mathcal{S}(\bm\xi^{(k)})\right\}\\
    &\bm\xi^{(k+1)}\longleftarrow \mathcal{T}(\bm\xi^{(k+1)};\gamma),\nonumber
\end{align}
where the thresholding operator $\mathcal{T}(\cdot;\gamma)$ is defined as
\begin{equation}\label{eq: threshold_operator}
    \mathcal{T}_i(\bm{\xi};\gamma) = 
    \begin{cases}
    \xi_i& \text{if }\ |\xi_i| > \gamma\\
    0              & \text{otherwise}
\end{cases}
,\quad i = 1,\dots,p. 
\end{equation}

The choice of the threshold parameter $\gamma$ remains a challenge since the magnitude of each entry of $\bm{\mathbf{\xi}}$ is unknown and depends on the selected candidate basis. Mangan et al.~\cite{Mangan2017} suggest the Akaike Information Criteria (AIC) for selecting $\gamma$; however, their approach is computationally expensive since many validation sets are needed to produce accurate results, and they use exact state derivatives instead of computing them from noisy state variables. Later, Rudy et al.~\cite{Rudy2017} and Quade et al.~\cite{Quade2018} proposed a Sequential Thresholded Ridge Regression (STRidge) to solve for $\bm\xi$. STRidge solves a Ridge regression problem -- a Tikhonov regularized and an improved variant of least squares regression -- and thresholds the coefficients with magnitude smaller than $\gamma$. The aim of STRidge it thus to improve the condition number of the linear system arising from the least squares problem (\ref{eq:STLS}) in STLS. The threshold parameter is chosen -- from several candidates --  based on cross validation or a Pareto curve \cite{Brunton2016}. 

In practice, BPDN, STLS, and STRidge lead to accurate recovery of $\bm\xi$ for small state measurement noises. However, as we shall illustrate in the examples of Section \ref{sec:numericalexamples}, the accuracy of the recovered coefficients for general dynamical systems may deteriorate considerably when the state variables and their time derivatives are polluted with relatively large noise levels. One reason for this lack of robustness to noise is that the measurement matrix $\bm\Phi(\mathbf{x})$ does not in general satisfy the incoherence \cite{Bruckstein09,Doostan11a,Hampton15a} or restricted isometry property~\cite{Candes08c, rauhut2012sparse, peng2016polynomial} conditions in the under-sampled case, $m<p$, or the incoherence property~\cite{Cohen13a,Hampton15b} in the over-sampled case, $m\ge p$. The reason for this is two fold: Firstly, $\mathbf{f}(\mathbf{x})$ may be sparse in a basis $\{\phi_{i}(\mathbf{x})\}$ that is not orthonormal, e.g., monomials. Secondly, unlike in standard compressed sensing or least squares regression settings, $\phi_{i}(\mathbf{x})$ are sampled at the state variables $\mathbf{x}$, which follow the dynamics of the system, as opposed to random (or experimentally designed) samples that would lead to well-conditioned measurement matrices; see, e.g.~\cite{Hadigol18}. 

\begin{remark}
Notice that, in practice, only noisy measurements of state variables are available and not their true values. Similarly, the time derivatives of state variables are approximated from noise-contaminated state measurements. Therefore, the linear system (\ref{eq:matrix_main_system}) is indeed a perturbed variant of the true but unattainable system. More precisely, (\ref{eq:matrix_main_system}) may be written as $\dot{\mathbf{x}}^* + \delta\dot{\mathbf{x}}= \bm{\Phi}(\mathbf{x}^* + \delta\mathbf{x})\bm{\xi}$, where $\mathbf{x}^*$ and $\dot{\mathbf{x}}^*$ are the exact state variables and state derivatives, respectively. Here, $\delta\mathbf{x}$ are deviations from the exact state variables caused by the measurement noise, and $\delta\dot{\mathbf{x}}$ are the errors in the exact state derivatives due to the numerical differentiation of noisy state variables.
\end{remark}

\section{Approach: Reweighted $\ell_1$-regularized Least Squares}
\label{sec:approach}

To improve the robustness of SINDy with respect to the state and state derivative noise, we propose regularizing the regression problem involving (\ref{eq:matrix_main_system}) via weighted $\ell_1$-norm of $\bm\xi$, 
\begin{equation}
\Vert\mathbf{W}\bm\xi\Vert_1 = \sum_{i=1}^p w_i\vert\xi_i\vert.    
\end{equation}
Here, $\mathbf{W}\in\mathbb{R}^{p\times p}$ is a diagonal matrix with diagonal entries $w_i>0$, $i=1,\dots,p$. Our approach is inspired by the work in~\cite{zou2006adaptive, Candes2008,Yang13,Peng14,adcock2017infinite}
from the statistics, compressed sensing, and function approximation literature, where weighted $\ell_1$-norm has been shown to outperform the standard $\ell_1$-norm in promoting sparsity, especially in the case of noisy measurements or when the solution of interest is not truly sparse, i.e., many entries of $\bm\xi$ are near zero~\cite{Candes2008,Yang13,Peng14,adcock2017infinite}. Depending on the choice of $\mathbf{W}$, $\Vert\mathbf{W}\bm\xi\Vert_1$ gives a closer approximation to the $\ell_0$-norm of $\bm\xi$, $\Vert\bm\xi\Vert_0:=\#\{i:\xi_i\neq 0\}$, than $\Vert\bm\xi\Vert_1$, and thus better enforces sparsity in $\bm\xi$. 

More specifically, we solve the weighted variant of the BPDN problem (\ref{eq:BPDN}),
\begin{equation}\label{eq:WBPDN}
(\text{WBPDN})\qquad    \min_{\bm{\xi}} \norm{\bm{\Phi}(\mathbf{x})\bm{\xi} - \dot{\mathbf{x}}}_2^2 + \lambda \norm{\mathbf{W}\bm{\xi}}_1,
\end{equation}
which coincides with the adaptive LASSO approach of~\cite{zou2006adaptive}. The problem in (\ref{eq:WBPDN}) may be solved via BPDN solvers for standard $\ell_1$-minimization with the simple transformations $\tilde{\bm\xi}:=\mathbf{W}\bm\xi$ and $\tilde{\bm{\Phi}}(\mathbf{x}) = \bm{\Phi}(\mathbf{x})\mathbf{W}^{-1}$, i.e.,
\begin{equation}\label{eq:transformed-BPDN}
    \min_{\tilde{\bm{\xi}}} \norm{\tilde{\bm{\Phi}}(\mathbf{x})\bm{\xi} - \dot{\mathbf{x}}}_2^2 + \lambda \norm{\tilde{\bm{\xi}}}_1.\nonumber
\end{equation}
Given the solution $\tilde{\bm\xi}$ to (\ref{eq:transformed-BPDN}), $\bm\xi$ is then computed from $\bm\xi = \mathbf{W}^{-1}\tilde{\bm\xi}$. In what follows, we describe the selection of the weight matrix $\mathbf{W}$ and regularization parameter $\lambda$. 

\subsection{Setting weights $\mathbf{W}$}
\label{sec:weights}

The main goal of using a weighted $\ell_1$-norm -- instead of its standard counterpart -- is to place a stronger penalty on the coefficients $\xi_i$ that are anticipated to be small (or zero). The obvious choice is to set $w_i$ inversely proportional to $\vert\xi_i\vert$, which is of course not possible as $\xi_i$ is unknown. An alternative approach, proposed first in \cite{zou2006adaptive, Candes2008}, is to use approximate values of $\vert\xi_i\vert$ to set $w_i$. In details, an iterative approach is devised where, at iteration $k+1$, the WBPDN solution $\bm\xi^{(k)}$ from iteration $k$ is used to generate the weights according to
\begin{equation}\label{eq:WL1weigths}
    w_i^{(k+1)} = \frac{1}{|\xi_i^{(k)}|^q + \epsilon},
\end{equation}
where $q>0$ represents the strength of the penalization and $\epsilon$ is a small value to prevent numerical issues when $\xi_i$ is zero. In our numerical experiments, we set $q=2$ and $\epsilon = 10^{-4}$ as they consistently produce better solution; however, {\it optimal} values of $q$ and $\epsilon$ may be selected along with $\lambda$ using the approach discussed in Section \ref{subsec:modelselection}. The iterations are started by solving the standard BPDN problem (\ref{eq:BPDN}) to compute $\bm\xi^{(0)}$. Algorithm \ref{alg:WBPDN}, adopted from \cite{Candes2008}, outlines the steps involved in WBPDN.

Similar to BPDN, WBPDN possesses the properties listed below~\cite{Kim2007}:
\begin{itemize}
    \item Nonlinearity: WBPDN yields a solution $\bm{\xi}$ that is nonlinear in $\dot{\mathbf{x}}$.
    \item Limiting behavior as $\lambda \rightarrow 0$: the WBPDN solution tends to the ordinary least squares solution as $\lambda \rightarrow 0$.
    \item Finite convergence to zero as $\lambda \rightarrow \infty$: the WBPDN solution converges to zero for a finite value of $\lambda$ defined as $\lambda \geq \lambda_{\max} = \|2\bm{\Phi}(\mathbf{x})^T \dot{\mathbf{x}}\|_{\infty}$, where $\|\cdot\|_{\infty} \coloneqq \max|\cdot|$ is the infinity norm.
    \item Regularization path: there are values $0 = \lambda_k < \dots < \lambda_1 = \lambda_{\max}$ such that the solution is a piece-wise linear curve on $\R^p$.
\end{itemize}
\begin{algorithm}[H]
\caption{Iteratively Reweighted Basis Pursuit Denoising adopted from \cite{Candes2008}}\label{alg:WBPDN}
\begin{algorithmic}[1]
\Procedure {WBPDN}{$\bm{\Phi}(\mathbf{x})$, $\dot{\mathbf{x}}$, $\lambda$, $q$, $\epsilon$}

\State Set the iteration counter to $k = 0$ and $w_i^{(0)} = 1, \quad i = 1,\dots,p$.
\While{\text{not converged} \text{or} $k < k_{\max}$}

    \State Solve the WBPDN problem (\ref{eq:WBPDN}) for a specific $\lambda$
    \begin{equation*}
    \bm{\xi}^{(k)} = \underset{\bm\xi}{\mathrm{argmin}}\left\{\norm{\bm{\Phi}(\mathbf{x})\bm{\xi} - \dot{\mathbf{x}}}_2^2 + \lambda \norm{\mathbf{W}^{(k)}\bm{\xi}}_1\right\}.
    \end{equation*}

    \State Update the weights for $i = 1,\dots,p$
    
    \begin{equation*}
    w_i^{(k+1)} = \frac{1}{|\xi_i^{(k)}|^q + \epsilon}.
    \end{equation*}

    \State $k = k + 1$.
\EndWhile
\EndProcedure
\end{algorithmic}
\end{algorithm}
\subsection{Selection of $\lambda$ via Pareto curve}
\label{subsec:modelselection}

\begin{figure}[H]
    \centering
    \includegraphics[width=0.6\textwidth]{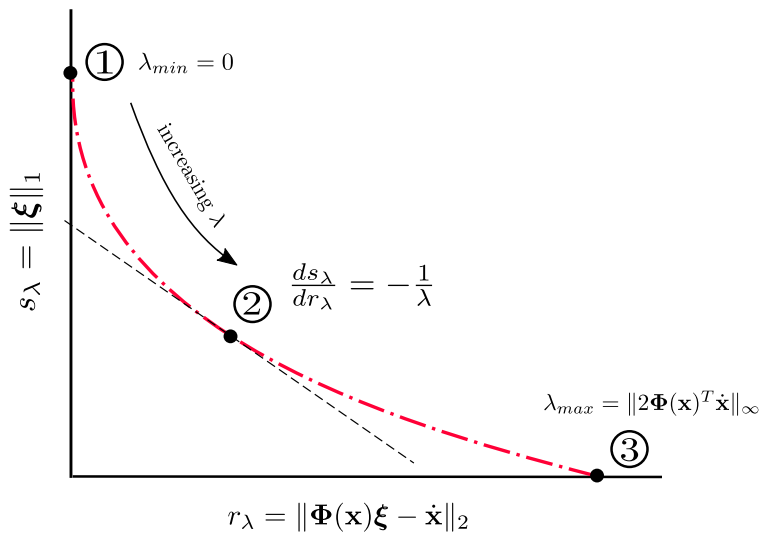}
    \caption{Illustration of the $\ell_1$-norm Pareto curve. Point 1 represents the unregularized least squares solution with infinite slope. Point 2 represents any point along the curve with slope $-1/\lambda$. Point 3 represents the upper bound on $\lambda$ that gives the zero solution.}
    \label{fig:ParetoCurve}
\end{figure}

The Pareto curve is a graph that traces the trade-off between the residual and the regularization constraint by varying the parameter $\lambda$ in (\ref{eq:WBPDN}). In Tikhonov regularization, where $\ell_2$-norm is used as a regularizer, the Pareto curve is referred as L-curve when a log-log scale is employed~\cite{Hansen1992Lcurve}. In this work, the Pareto curve is defined as the graph generated by solving (\ref{eq:WBPDN}) for different $\lambda$ values in the ($\norm{\bm{\Phi}(\mathbf{x})\bm{\xi} - \dot{\mathbf{x}}}_2$,$\norm{\mathbf{W}^{(k)}\bm{\xi}}_1$) space. Recall that $\lambda$ controls the sparsity of the solution $\bm{\xi}$; setting $\lambda = 0$ yields the ordinary least squares solution, which is not sparse. As $\lambda$ increases, the non-sparse solutions are increasingly penalized. In contrast to the $\ell_2$-norm regularization, where the solution tends to zero as $\lambda$ tends to infinity, the $\ell_1$-norm regularization yields an upper bound on the regularization parameter given by $\lambda_{\max} = \norm{2\bm{\Phi}(\mathbf{x})^T \dot{\mathbf{x}}}_{\infty}$ (i.e. a $\lambda \geq \lambda_{\max}$ yields the zero solution).
As proven in~\cite{VanDenBerg2008}, the $\ell_1$-norm Pareto curve is convex, continuously differentiable and non-increasing. Its slope at each point is given by $-1/\lambda$, as shown in Figure~\ref{fig:ParetoCurve}. Therefore, the regularization parameter that yields an approximation close to the exact $\bm{\xi}$, within a noise-dependent distance, must live within $\lambda_{\min}$ and $\lambda_{\max}$. For $\ell_2$-regularized least squares, Hansen~\cite{Hansen1992Lcurve} suggests the corner point criterion to select $\lambda$ from the L-curve. The underlying idea is that the L-curve has an L-shaped corner located where the solution changes from being dominated by regularization errors, corresponding to the steepest part of the curve, to being dominated by noise errors, where the curve becomes flat. The corner point corresponds to an {\it optimal} balance between the sparsity of the solution and the quality of the fit. The present work adopts this corner point criterion to select the $\lambda$ parameter from the log-log $\ell_1$-norm Pareto curve.
For continuous Pareto curves,~\cite{Hansen1992Lcurve} suggests defining the corner point as the point with maximum curvature. For discrete curves, however, it becomes more complicated to define the location of the corner point. Hansen et al.~\cite{Hansen2007} highlight the difficulties in computing the corner point from discrete L-curves built using a finite number of $\lambda$ values. The discrete curves may contain small local corners other than the global one that may give sub-optimal regularization parameters. To alleviate this issue, they propose an adaptive pruning algorithm, where the best corner point is computed by using candidate corners from curves at different resolutions. Since the L-curves must be sampled at different scales, the pruning algorithm can be computationally expensive. We instead compute the corner points using a simpler method proposed in Cultrera et al.~\cite{Cultrera2016}. The algorithm iteratively employs the Menger curvature of a circumcircle and the golden section search method to efficiently locate the point on the curve with maximum curvature. The benefit of this algorithm is that it does not require computing the solution for a large set of $\lambda$ values.

Cross-validation (CV) is a popular approach for choosing the best regularization parameter $\lambda$ in regression problems. However, when applied to (\ref{eq:BPDN}), it is known that CV prefers smaller values of $\lambda$ that are associated with smaller residual errors and less sparse coefficients. While CV was proposed to select the regularization parameter of adaptive LASSO, as we shall illustrate in the numerical examples of Section \ref{sec:numericalexamples}, the Pareto curve method produces more accurate results. Similarly, in \cite{tehrani2012l1}, the Pareto method has been shown to outperform CV for sparse approximation using LASSO. 

\subsection{Numerical computation of state time derivatives $\dot{\mathbf{x}}$}
\label{subsec:numericaldifferentiation}

In most practical applications, only the state trajectories are available, and their derivatives $\dot{\mathbf x}$ must be estimated numerically. To generate the data $\dot{\mathbf x}$ in the WBPDN problem (\ref{eq:WBPDN}), we seek to estimate the discrete time derivatives $\dot{{x}}_j(t_i)$, $i = 1,\dots,m$, at each time instances $t_i$ from a set of discrete, noisy state trajectories ${x}_j(t_i)$.

Finite difference method is a common approach to compute derivatives from discrete data. However, if the data is corrupted with noise, the error in the derivative approximation by finite difference will generally be amplified. Several alternatives have been proposed to mitigate the effect of noise when computing derivatives: locally weighted regression~\cite{William1979}, Tikhonov regularization~\cite{Cullum1971}, total variation regularization~\cite{Chartrand2011}, to name a few. Following the work by Knowles et al.~\cite{Cullum1971,Knowles2014}, we employ the second-order Tikhonov regularization approach, wherein the derivative is formulated as the solution to an integral equation and solved for using a Tikhonov-regularized least squares problem. The detail of our implementation is described in Appendix A.

\section{Identifying Single Constraints from Data}
\label{subsec:constrainedsystems}

Apart from satisfying governing equations of the form (\ref{eq:DynamicalSystem}), some dynamical systems evolve under additional constraints on the state variables arising from certain conservation laws. For example, fluids are often assumed to satisfy the incompressibility condition or the total energy of a conservative system is constant. In the case of a single constraint, the dynamical systems can be represented as
\begin{equation}\label{eq:ConstrDynamicalSystem}
    \dot{\mathbf{x}} = \mathbf{f}(\mathbf{x}) \quad \text{subject to} \quad g(\mathbf{x}) = 0,
\end{equation}
where $g\colon \R^n \to \R$ is the constraint function. In this article, we focus on learning from data single constraints that are functions of the state variables with no explicit dependence on time. In the noise-free case, assuming that the constraint can be expressed in the same basis $\{\phi_k(\mathbf{x})\}$ for recovering the governing equations in (\ref{eq:dynExpansion}), $g(\mathbf{x})$ is given by
\begin{equation}\label{eq:Constraint}
   g(\mathbf{x}) = \sum_{k=1}^{p}\eta_{k} \phi_k(\mathbf{x})=0,
\end{equation}
for some unknown coefficients $\eta_k$. As an example, consider an undamped single degree of freedom spring-mass system. The conservation of energy can be put in implicit form as $g(\mathbf{x}) = 2\mathcal{E} - kx_1^2 - mx_2^2$, where $\mathcal{E}$ is the total energy of the system, $k$ the spring stiffness, $m$ the mass, and $x_1$ and $x_2$ the displacement and velocity of the system, respectively. In this case, the active bi-variate monomial basis functions in (\ref{eq:Constraint}) are $\phi_1(\mathbf{x}) = 1$, $\phi_4(\mathbf{x}) = x_1^2$, $\phi_6(\mathbf{x}) = x_2^2$ and the corresponding coefficients are respectively $\eta_1 = -2\mathcal{E}$, $\eta_4 = k$ and $\eta_6 = m$.

Evaluating (\ref{eq:Constraint}) at the state samples and letting $\bm{\eta} =[\eta_{1}, \dots, \eta_{p}]^T \in \R^{p}$, we arrive at
\begin{equation}\label{eq:Constraint_matrix}
    \bm{\Phi}(\mathbf{x})\bm{\eta}=\mathbf{0},
\end{equation}
which implies that $\bm{\Phi}(\mathbf{x})$ has a non-empty null-space, i.e., is rank-deficient, and the solution $\bm{\xi}$ is non-unique. In practice, state measurements contain noise and the constraints are not satisfied exactly. As a result, the columns of $\bm{\Phi}({\mathbf{x}})$ are {\it nearly} linearly dependent and $\bm{\Phi}({\mathbf{x}})$ may have a large condition number. This, in turn, results in high sensitivity of the solution $\bm\xi$ to the noise in $\dot{\mathbf x}$. In addition to learning the constraint $g(\mathbf{x})$, the identification of dependent columns of $\bm{\Phi}({\mathbf{x}})$ provides a means to improve its condition number. 

To this end, we perform a low-rank factorization of $\bm{\Phi}({\mathbf{x}})$ via interpolative decomposition (ID)~\cite{Cheng2005} in order to identify a subset of the columns of $\bm{\Phi}({\mathbf{x}})$ that form a basis for its range space. In detail, using a rank-revealing QR factorization (RRQR)~\cite{Golub1996}, $\bm{\Phi}(\mathbf{x})$ is decomposed to
\begin{equation}
    \bm{\Phi}({\mathbf{x}})\mathbf{P} \approx \mathbf{Q}\mathbf{R},
\end{equation}
where $\mathbf{P} \in \R^{p \times p}$ is a permutation matrix, $\mathbf{Q} \in \R^{m \times r}$ has $r$ orthogonal columns, and $\mathbf{R} \in \R^{r \times p}$ is an upper triangular matrix. Let $^{\dagger}$ denote the pseudoinverse of a matrix. Partitioning $\mathbf{R}$ into $\mathbf{R}_{1} \in \R^{r \times r}$, and $\mathbf{R}_{2} \in \R^{r \times (p - r)}$ and assuming the relation $\mathbf{R}_{2} \approx \mathbf{R}_{1} \mathbf{Z}$ for $\mathbf{Z}=\mathbf{R}_{1}^{\dagger}\mathbf{R}_{2}$ yields
\begin{equation}\label{eq:rankrfact}
    \bm{\Phi}({\mathbf{x}})\mathbf{P} \approx \mathbf{Q}\mathbf{R}_{1}[\mathbf{I}\,\,|\,\,\mathbf{Z}],
\end{equation}
where $\mathbf{I} \in \R^{r \times r}$ is the identity matrix. The rank-$r$ factorization (\ref{eq:rankrfact}) can be rewritten as
 \begin{equation}\label{eq:rankrfactSC}
    \bm{\Phi}({\mathbf{x}})\approx \mathbf{S}[\mathbf{I}\,\,|\,\,\mathbf{Z}]\mathbf{P}^T = \mathbf{S}\mathbf{C},
\end{equation}
where $\mathbf{S} = \mathbf{Q}\mathbf{R}_{1}$ contains $r$ columns of $\bm{\Phi}({\mathbf{x}})$ -- specifically, the first $r$ columns of $\bm{\Phi}({\mathbf{x}})\mathbf{P}$ -- and $\mathbf{C} = [\mathbf{I}\,\,|\,\,\mathbf{Z}]\mathbf{P}^T$ is the coefficient matrix specifying the linear combinations of those columns that approximate $\bm{\Phi}({\mathbf{x}})$. Let $\mathcal{I} = \{1,\dots,p\}$ and $\mathcal{J} = \{j_1,\dots,j_r\}\subset \mathcal{I}$ denote, respectively, the set of indices of columns of $\bm{\Phi}({\mathbf{x}})$ and the subset of those forming $\mathbf{S}$. Given the ID approximation (\ref{eq:rankrfactSC}), any column $\bm{\phi}_l(\mathbf{x})$ of $\bm{\Phi}({\mathbf{x}})$ with index $l \in \mathcal{I}\backslash\mathcal{J}$ therefore satisfies the approximate relation
\begin{equation}
\label{eq:constraint_approx}
    \sum_{m=1}^r C_{m,l}\bm{\phi}_{j_m}(\mathbf{x}) - \bm{\phi}_l(\mathbf{x}) \approx \mathbf{0},\quad l \in \mathcal{I}\backslash\mathcal{J},
\end{equation}
in which $C_{m,l}$ is the entry $(m,l)$ of $\mathbf{C}$. Rewriting (\ref{eq:constraint_approx}) in the form of (\ref{eq:Constraint_matrix}), the coefficients $\eta_k$ in (\ref{eq:Constraint_matrix}) are approximated by
\begin{equation}\label{eq:eta_recovery}
    \eta_k = \begin{cases}
C_{m,l} & \text{if }\ \quad k = j_m,\quad m= 1,\dots,r,\\
-1 & \text{if }\ \quad k = l,\\
0  & \text{otherwise}.
\end{cases}
\end{equation}
Notice that the $\mathbf{C}$ matrix may contain non-zero but small elements due to the noise in the data. We therefore propose a thresholding step by setting to zero the elements of $\mathbf{C}$ whose magnitude is smaller than a threshold parameter $\tau$.

To improve the conditioning of $\bm{\Phi}({\mathbf{x}})$ in order to calculate $\bm\xi$ several paths may be taken. We may remove columns with indices in $\mathcal{I}\backslash\mathcal{J}$, i.e., replace $\bm{\Phi}({\mathbf{x}})$ by $\mathbf{S}$, or remove columns with indices in $\mathcal{I}$ to which the columns in $\mathcal{I}\backslash\mathcal{J}$ depend on. In doing these, for the  interest of arriving at a {\it simpler} model, we may remove the monomial basis functions of highest degree. If we happen to remove a basis function $\phi_i(\mathbf{x})$ that appears in the original dynamical system, the corresponding components of $\bm\xi$ will not be recovered. However, still a correct $\mathbf{f}(\mathbf{x})$ is computed.  The procedure to learn single constraints $g(\mathbf{x})$ using ID of $\bm{\Phi}({\mathbf{x}})$ and remove dependent columns of  $\bm{\Phi}({\mathbf{x}})$ is outlined in Algorithm \ref{alg:ID}.

\begin{algorithm}[H]
\caption{Identification of constraints via ID}\label{alg:ID}
\begin{algorithmic}[1]
\State Compute the singular values of $\bm{\Phi}({\mathbf{x}})$.

\State Identify a gap on the singular values and find the numerical rank $r$ of $\bm{\Phi}({\mathbf{x}})$.

\State Perform rank-$r$ ID of $\bm{\Phi}({\mathbf{x}}) \approx \mathbf{S}\mathbf{C}$ given in (\ref{eq:rankrfactSC}) and obtain $\mathcal{J} = \{j_1,\dots,j_r\}$.
\State Threshold $\mathbf{C}$ by setting entries of $\mathbf{C}$ smaller than $\tau$ to zero.
\State Given $\mathbf{C}$ and $\mathcal{J} = \{j_1,\dots,j_r\}$, select any column $l \in  \mathcal{I}\backslash\mathcal{J}$ and set the $\eta_k$ coefficients in (\ref{eq:Constraint}) via (\ref{eq:eta_recovery}).

\State Remove dependent columns of $\bm{\Phi}({\mathbf{x}})$, e.g., set $\bm{\Phi}({\mathbf{x}}) \longleftarrow \mathbf{S}$, for calculating $\bm\xi$.
\end{algorithmic}
\end{algorithm}

\begin{remark}
Notice that the ID in (\ref{eq:rankrfactSC}) requires the knowledge of the numerical rank of $\bm{\Phi}({\mathbf{x}})$, $r$, which may be detected by identifying a gap in the magnitude of the singular values of $\bm{\Phi}({\mathbf{x}})$. However, depending on the level of noise in state measurements, such a gap may not exist or clearly identifiable, thus rendering the proposed constraint learning approach inaccurate. In Section \ref{sec:numericalexamples}, we provide empirical results clarifying this remark.  
\end{remark}

\begin{remark} In case the state variables are constrained by several implicit functions, there is no guarantee we are able to recover each individual one using Algorithm \ref{alg:ID}. This is because the state trajectory is confined to the intersection of the constraints, which leads to rank deficiency or ill-conditioning of $\bm{\Phi}(\mathbf{x})$. In the case of exact state measurements, removing the linearly dependent columns of $\bm{\Phi}(\mathbf{x})$, addresses the rank deficiency issue and reveals the constraint intersection,  as  illustrated in the numerical example of  Section \ref{subsec:EulerRBD}.
\end{remark}

\section{Numerical examples}
\label{sec:numericalexamples}
In this section, we present numerical examples to assess the performance of the WBPDN method to recover the ordinary differential equations (ODEs) of five nonlinear dynamical systems. In all cases, we assume no prior knowledge about the governing equations that generated the data, except that they can be sparsely expressed in a multivariate polynomial basis in the state variables with known dimension. We only have access to the noisy state measurements at discrete times $t_k = k \Delta t$ sampled every $\Delta t$ units of time. The exact state variables are computed by integrating the nonlinear ODEs using the fourth-order explicit Runge-Kutta (RK4) integrator implemented in MATLAB 2018b with a tolerance of $10^{-10}$. We then corrupt the exact state variables by adding different levels of noise. In this work, we assume that the state variables are contaminated with independent zero-mean additive white Gaussian noise with variance $\sigma^2$. The noise model is given by
\begin{eqnarray}
    x_j(t_k) = x_j^*(t_k) + \eta_j(t_k),
\end{eqnarray}
where $x_j^*(t_k)$, $j = 1,\dots,n$, denotes the exact state variable at time $t_k,\,\,k = 1,\dots,m$, and $\eta_j(t_k) \sim \mathcal{N}(0,\sigma)$. The noise levels are varied from $\sigma_{\min} $ to $\sigma_{\max}$ depending on the magnitude of the signal for each case. To measure the signal magnitude relative to the noise level, we provide the signal-to-noise ratio (SNR) for each state $j$. The SNR, expressed in decibels, is defined as
\begin{equation}
    (\text{SNR}_j)_{\text{dB}} =
    10~\text{log}_{10}\bigg(\frac{\sum_{k = 1}^{m}x_j(t_k)^2}{\sigma^2}\bigg).
\end{equation}
The sampling time is fixed to $\Delta t = 0.01$, and the number of samples depends on the simulation time used to capture the essential behavior of each dynamical system. As discussed in Section \ref{subsec:numericaldifferentiation} and Appendix A, we use Tikhonov regularization differentiation with a second-order differential operator as a smoothing constraint to compute the time derivative of state variables. Following the remark in Appendix A, the state time derivatives are computed over an interval that is 5\% (from each side) larger than the intended training time span, but only the data over the original training time is retained to compute $\bm\xi$. To show the quality of the resulting derivatives, we report the relative error
\begin{equation}\label{eq:relderivativeerror}
    e_{\dot{x}} = \frac{\|\dot{\mathbf{x}} - \dot{\mathbf{x}}^*\|_2}{\|\dot{\mathbf{x}}^*\|_2},
\end{equation}
where $\dot{\mathbf{x}}$ and $\dot{\mathbf{x}}^*$ is the computed and true value of state derivatives, respectively.

We approximate the governing equations of each example by a multivariate monomial basis of total degree $d$ in $n$ state variables. That is, $\phi_i(\mathbf{x})$ in (\ref{eq:dynExpansion}) (and (\ref{eq:Constraint})) are given by
\begin{equation}
\phi_i(\mathbf{x})=\prod_{j=1}^{n}x_j^{i_j}\qquad \text{such that}\qquad \sum_{j=1}^n i_j \le d, \quad i_j \in \mathbb{N}\cup\{0\},
\end{equation}
where the non-negative integer $i_j$ denotes the degree of the monomial in state variable $x_j$. The size of the approximation basis is then $p = \binom{n + d}{n}$. For all test cases, we set the total degree of basis $d$ to one more than the highest total degree monomial present in the governing equations. Our primary goal is to demonstrate the consistency and robustness to noise of the WBPDN algorithm to identify an accurate approximation of the governing equations, as well as to enable the accurate prediction of future states of the system. We run the WBPDN optimization procedure using solveBP routine from the open-source SparseLab 2.1 package~\cite{Donoho2009sparselab,donoho2007sparselab} and report the relative solution error defined as
\begin{equation}\label{eq:relsolerror}
    e_{\xi} = \frac{\|\bm{\xi} - \bm{\xi}^*\|_2}{\|\bm{\xi}^*\|_2},
\end{equation}
where $\bm{\xi}$ and $\bm{\xi}^*$ are the approximate and exact solution vectors for each state variables, respectively.

The considered nonlinear dynamical systems are as as follows: the Lorenz 63 system as a base model for identifying chaotic dynamics, the Duffing and Van der Pol oscillators as nonlinear stiffness and damping models, and the harmonic single degree of freedom spring-mass system, and Euler Rigid Dynamics equations as conservative models satisfying physical constraints.

\subsection{Lorenz 63 system}
\label{subsec:Lorenz63}

The Lorenz 63 system is a canonical model for nonlinear chaotic dynamics that was developed by Lorenz  as a simplified model for atmospheric convection \cite{Lorenz1963}. This system of nonlinear ODEs has been fundamental for the study of chaotic systems, wherein the future states of the system are highly sensitive to initial conditions. The state trajectories are three-dimensional, chaotic, deterministic, non-periodic and confined within a butterfly-shaped attractor, making them hard to predict. The Lorenz 63 model is given by the following system of first-order equations
\begin{subequations}\label{eq:Lorenz63}
\begin{alignat}{3}
    \dot{x} = \gamma(y - x), &\quad x(0) = x_0,\label{eq:Lorenz63_a}\\
    \dot{y} = x(\rho - z) - y, &\quad y(0) = y_0,\label{eq:Lorenz63_b}\\
    \dot{z} = xy - \beta z, &\quad z(0) = z_0,\label{eq:Lorenz63_c}
\end{alignat}
\end{subequations}
where the parameter values are set to $\gamma = 10$, $\rho = 28$ and $\beta = 8/3$, and the initial condition is $(x_0,y_0,z_0) = (-8,7,27)$. Note that the first state derivative is linear in the state variables, and the second and third ones contain quadratic nonlinearities. Assuming a degree $d=3$ expansion for the right-hand-side of (\ref{eq:Lorenz63}), is described exactly by seven of the $p = 20$ monomials. 

\begin{figure}[H]
    \centering
    \includegraphics[trim = 0 4 0 1, clip,width=0.47\textwidth]{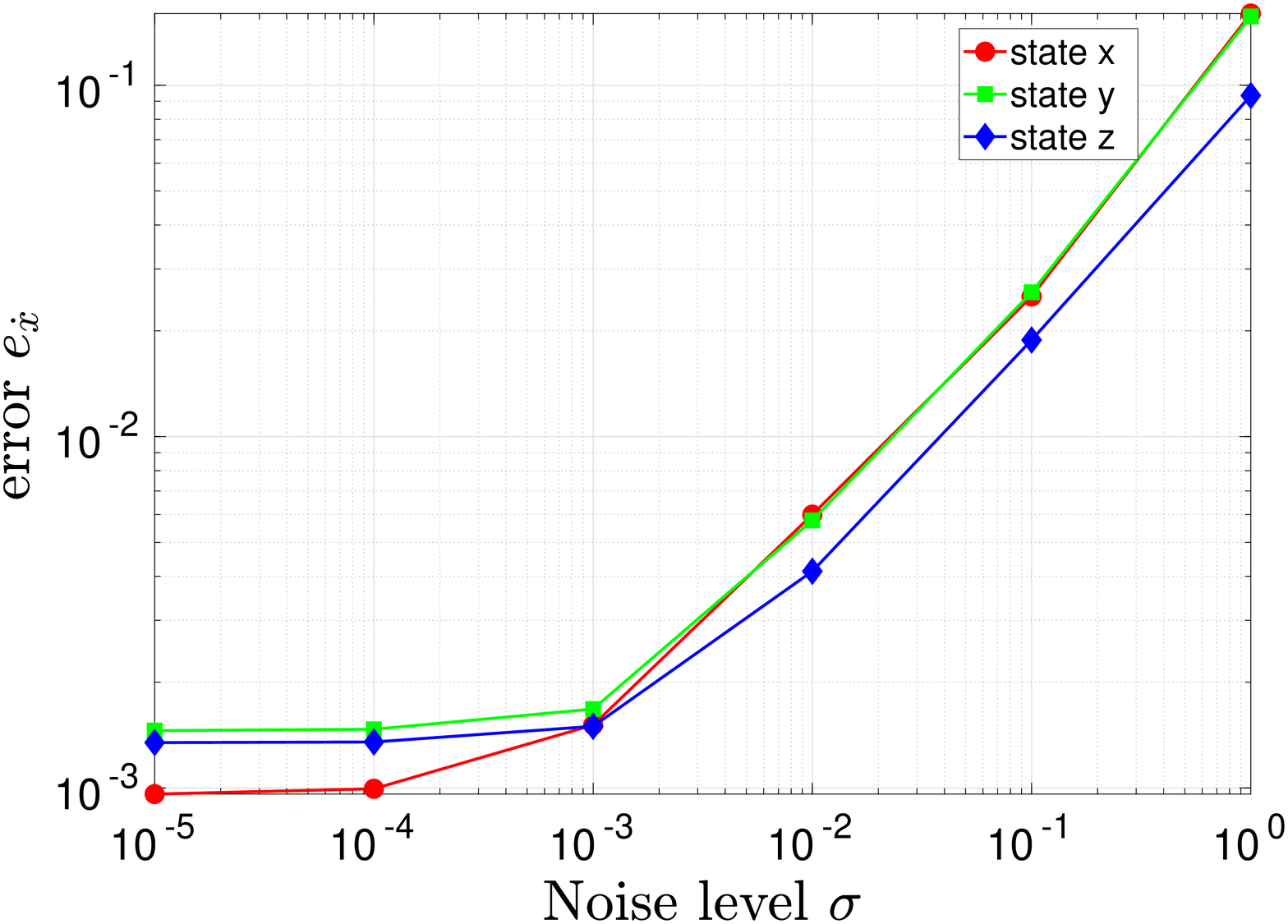}
    \includegraphics[trim = 0 4 0 1, clip,width=0.47\textwidth]{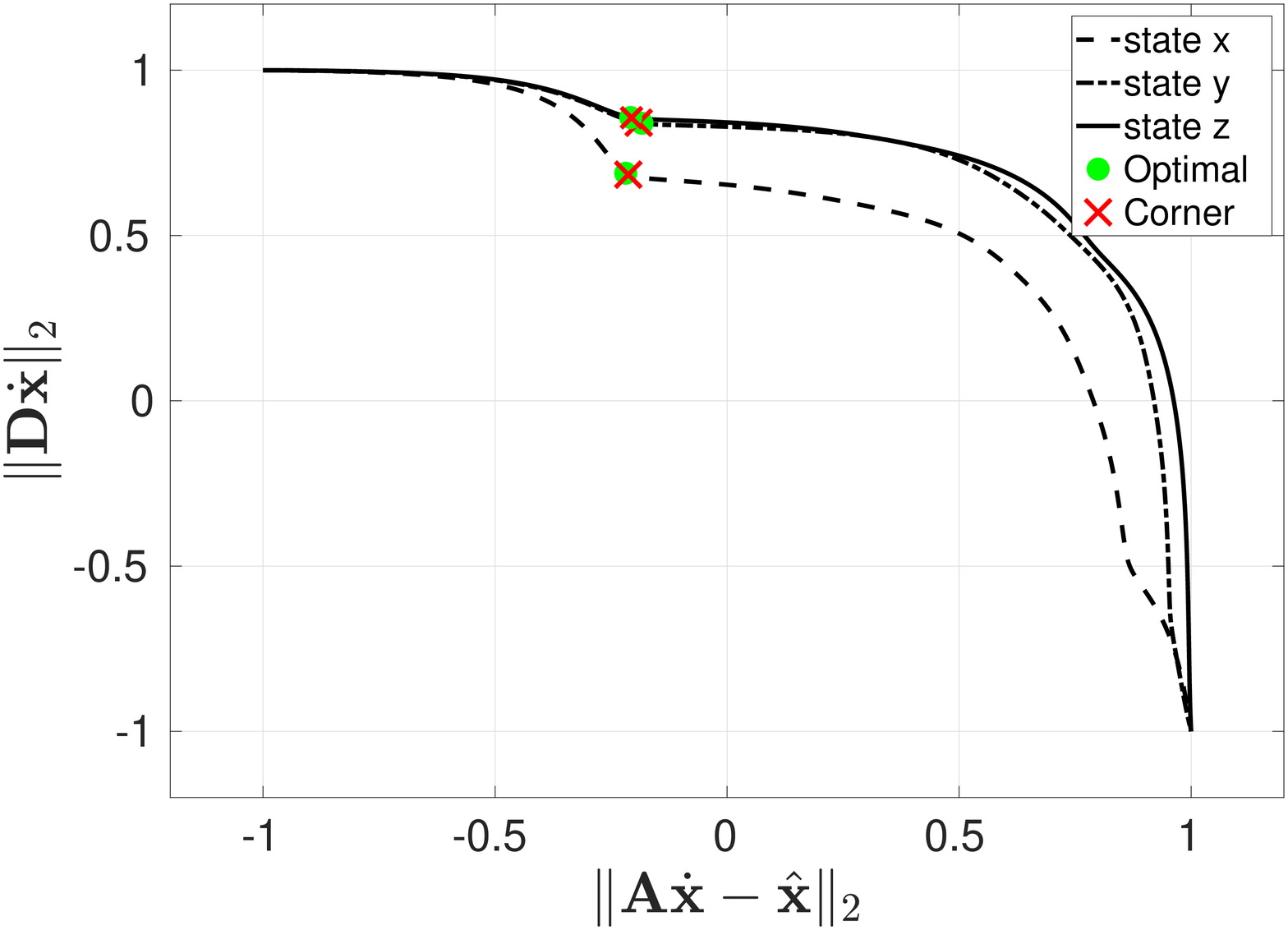}
    \caption{Left: Relative error of Lorenz 63 state derivatives with respect to different noise levels. Right: Pareto curves for each state at noise level $\sigma = 10^{-2}$. The signal-to-noise ratios are: $\text{SNR}_x = 58.11~\text{dB}$, $\text{SNR}_y =  59.46~\text{dB}$, and $\text{SNR}_z =  67.80~\text{dB}$.}
    \label{fig:LorenzTikDiff}
\end{figure}

We simulated the Lorenz 63 system from $t=0$ to $t=2.2$ time units to obtain the state trajectories. We then sampled the exact state variables at $\Delta t = 0.01$ resulting in $m=220$ samples, and perturbed them with noise at different levels $\sigma$. The first step to recover the governing equations is to numerically compute state derivatives from noisy measurements. We performed Tikhonov-regularized numerical differentiation and truncated the discrete state trajectory and resulting state derivatives yielding 200 samples from $t = 0.1$ to $t = 2.1$ time units. Figure \ref{fig:LorenzTikDiff} (left) shows the relative error in the state derivatives with respect to different noise levels. The flat region on Figure \ref{fig:LorenzTikDiff} (left) is dominated by time discretization errors $\sim \mathcal{O}(\Delta t^2)$ (independent of noise) when computing derivatives via numerical differentiation, whereas the steep region is dominated by noise. The Pareto curves are computed for each state at a noise level $\sigma = 10^{-2}$ in Figure~\ref{fig:LorenzTikDiff} (right). For clarity, the curves were normalized between $[-1,1]$. The locations of the computed corner points match well with the optimal points corresponding to which the solution error is minimal. 

\begin{figure}[H]
    \centering
    \includegraphics[trim = 0 0 0 0, clip,width=.49\textwidth]{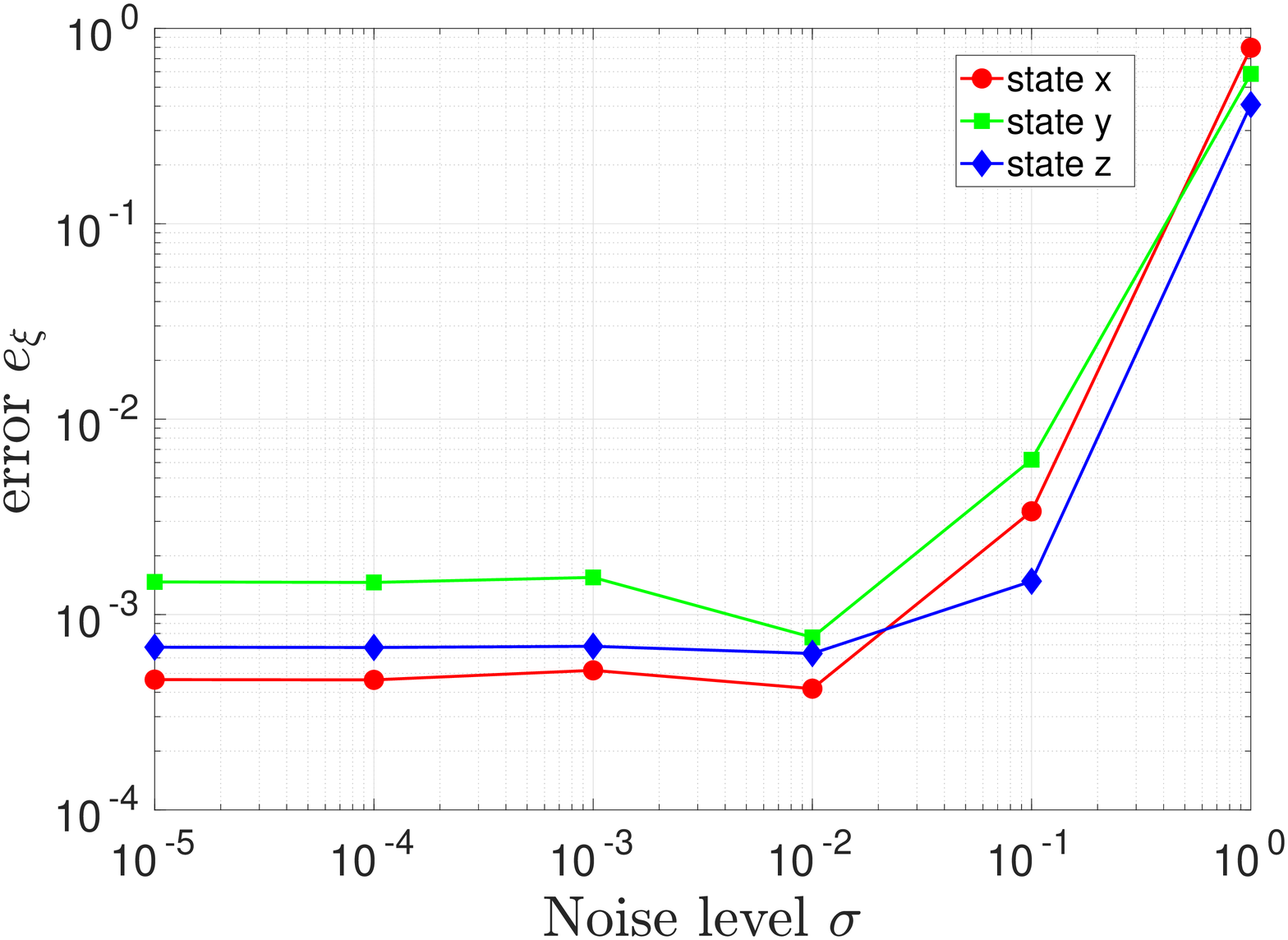}
    \includegraphics[trim = 0 0 0 0, clip,width=.49\textwidth]{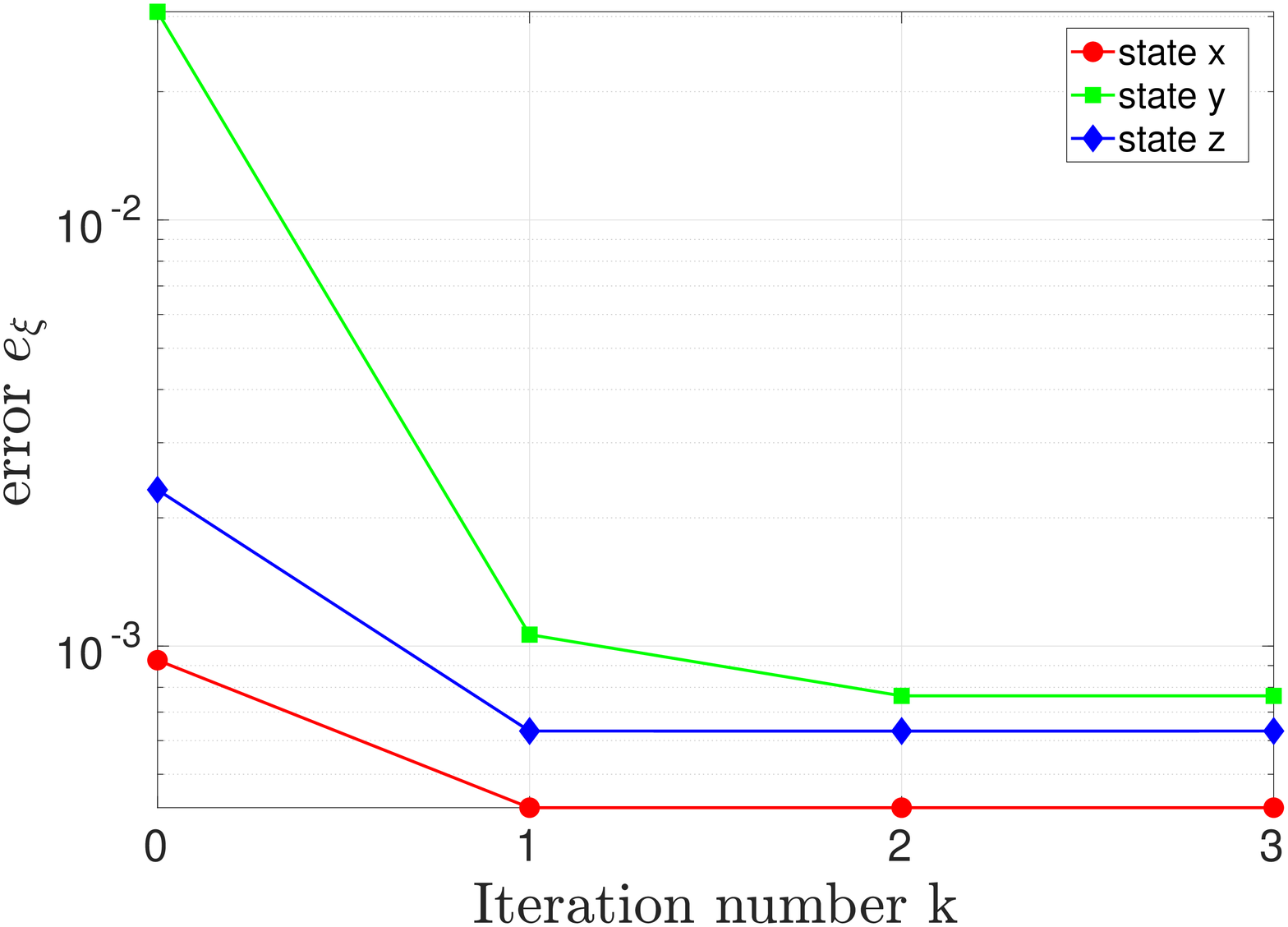}

    \caption{Left: Relative solution error of Lorenz 63 with respect to different noise levels $\sigma$. Right: Relative solution error of Lorenz 63 with respect to the iteration count $k$ at noise level $\sigma = 10^{-2}$.}
    \label{fig:Lorenzerrorcoeff}
\end{figure}

\begin{figure}[H]
    \centering
    \includegraphics[trim = 0 0 0 0, clip,width=0.6\textwidth]{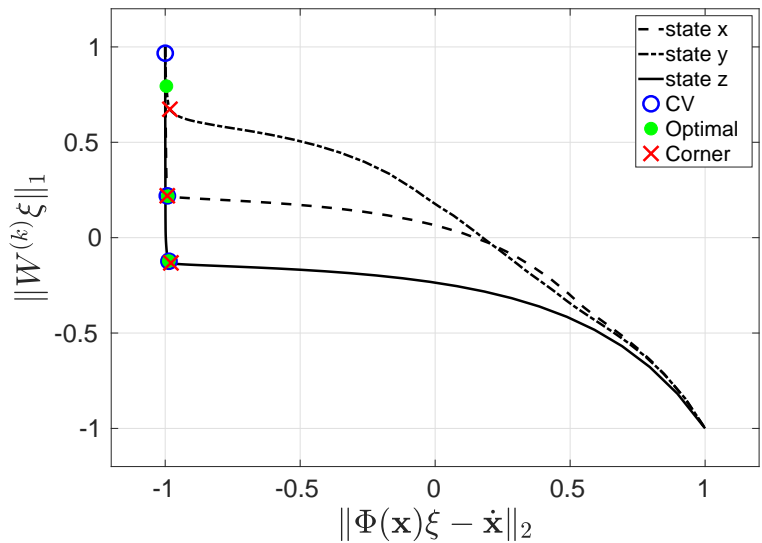}
    \caption{Pareto curves for each state of Lorenz 63 at noise level $\sigma = 10^{-2}$ and iteration 0.}
    \label{fig:LorenzWBPDNLcurve_it0}
\end{figure}

\begin{figure}[H]
    \centering
    \includegraphics[trim = 0 0 0 0, clip,width=0.54\textwidth]{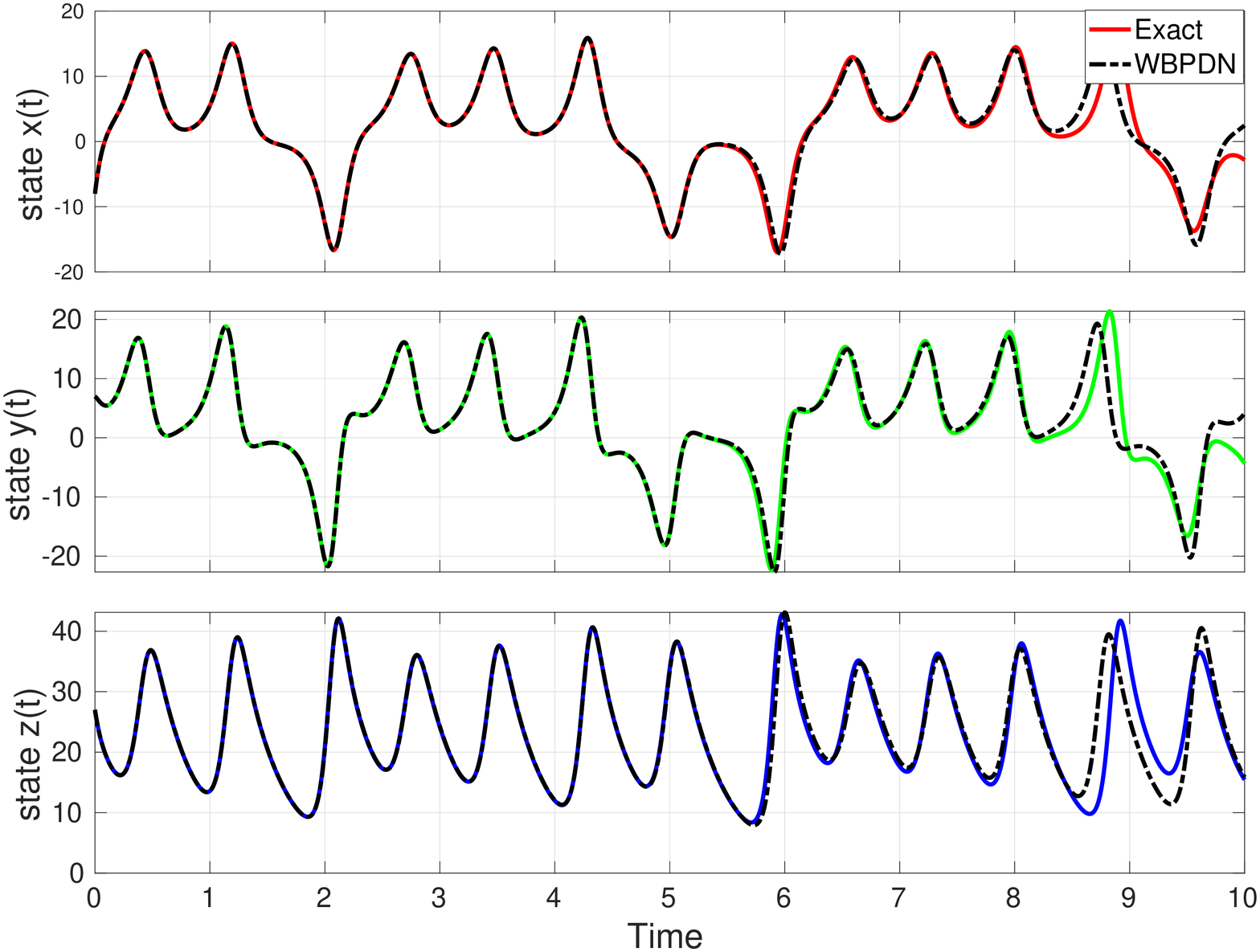}
    \includegraphics[trim = 0 0 0 0, clip,width=0.44\textwidth]{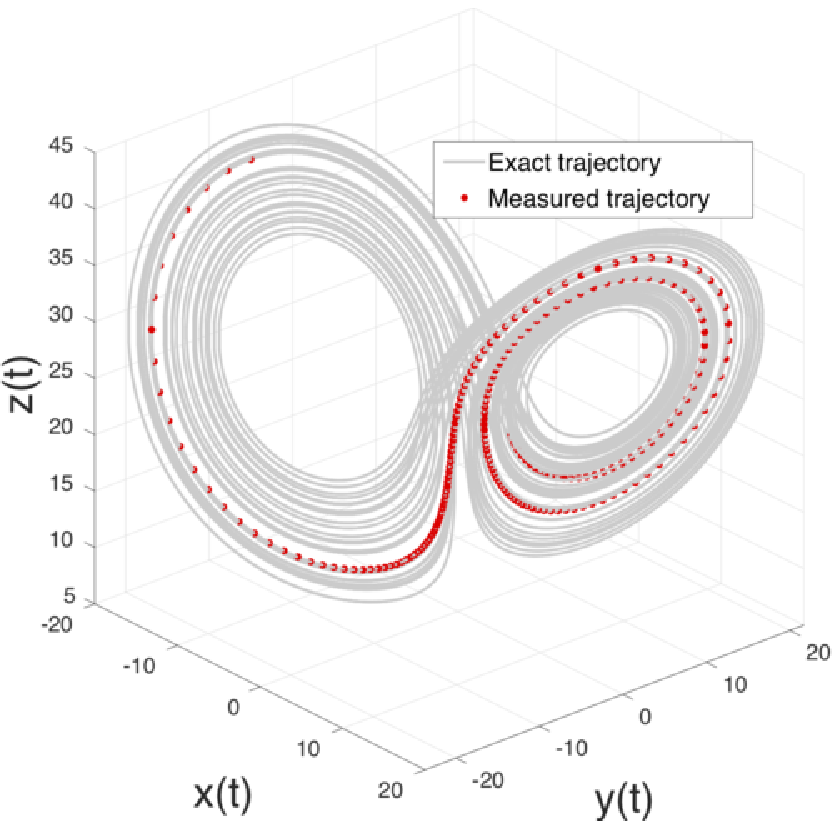}
    \caption{Left: Prediction of Lorenz 63 state trajectory by the identified model at noise level $\sigma = 10^{-2}$. Right: Exact and measured state trajectory of the Lorenz 63 system.}
    \label{fig:LorenzWBPDNPredicted}
\end{figure}

Figure~\ref{fig:Lorenzerrorcoeff} (left) illustrates the relative solution error as a function of the noise level $\sigma$. For small noise levels, the curve is almost flat and the solution error starts to rapidly increase around $\sigma \approx 10^{-2}$. Figure~\ref{fig:Lorenzerrorcoeff} (right), shows the effectiveness of iteratively reweighting the coefficients; few iterations are required (one in this example) to converge and considerably reduce the error in the computed coefficients. Notice that iteration $0$ of WBPDN is basically the standard BPDN solution.  
The Pareto curves for each state at iteration 0 and noise level $\sigma = 10^{-2}$ are shown in Figure~\ref{fig:LorenzWBPDNLcurve_it0}. We observe that the corner point criterion to select $\lambda$ yields near optimal regularization. Here, we also compare the corner point criterion with K-fold CV with $K = 5$ to select the regularization parameter $\lambda$. As seen, $\lambda$ obtained with CV matches the optimal and corner point for the $x$ and $y$ state variables, but is quite suboptimal for the $z$ state. The reason for this discrepancy is that CV hinges upon selecting the regularization parameter that minimizes the mean residual error over the folds, and does not take into account the $\ell_1$-norm of the solution. In our experience, the CV function (mean residual error vs. $\lambda$) is usually flat for a wide range of $\lambda$, meaning that there exist many $\lambda$ values that produce similar mean residual errors.
We now assess the performance of WBPDN by examining the prediction accuracy. Figure~\ref{fig:LorenzWBPDNPredicted} (left) compares the exact trajectory of the Lorenz 63 system with the predicted trajectory of the identified model for $\sigma = 10^{-2}$. The exact trajectory is computed by integrating~(\ref{eq:Lorenz63}), whereas the predicted one is computed using the identified model starting at the same initial conditions. The training set used to identify the system -- red points in Figure~\ref{fig:LorenzWBPDNPredicted} (right) -- ranges from $t=0.1$ to $t=2.1$, and the total simulation time is until $t=10$. As shown, the predicted trajectory agrees with the exact one for a relatively long time span, even though the system is chaotic. 

\begin{figure}[H]
    \centering
    \includegraphics[trim = 0 0 0 0, clip,width=0.8\textwidth]{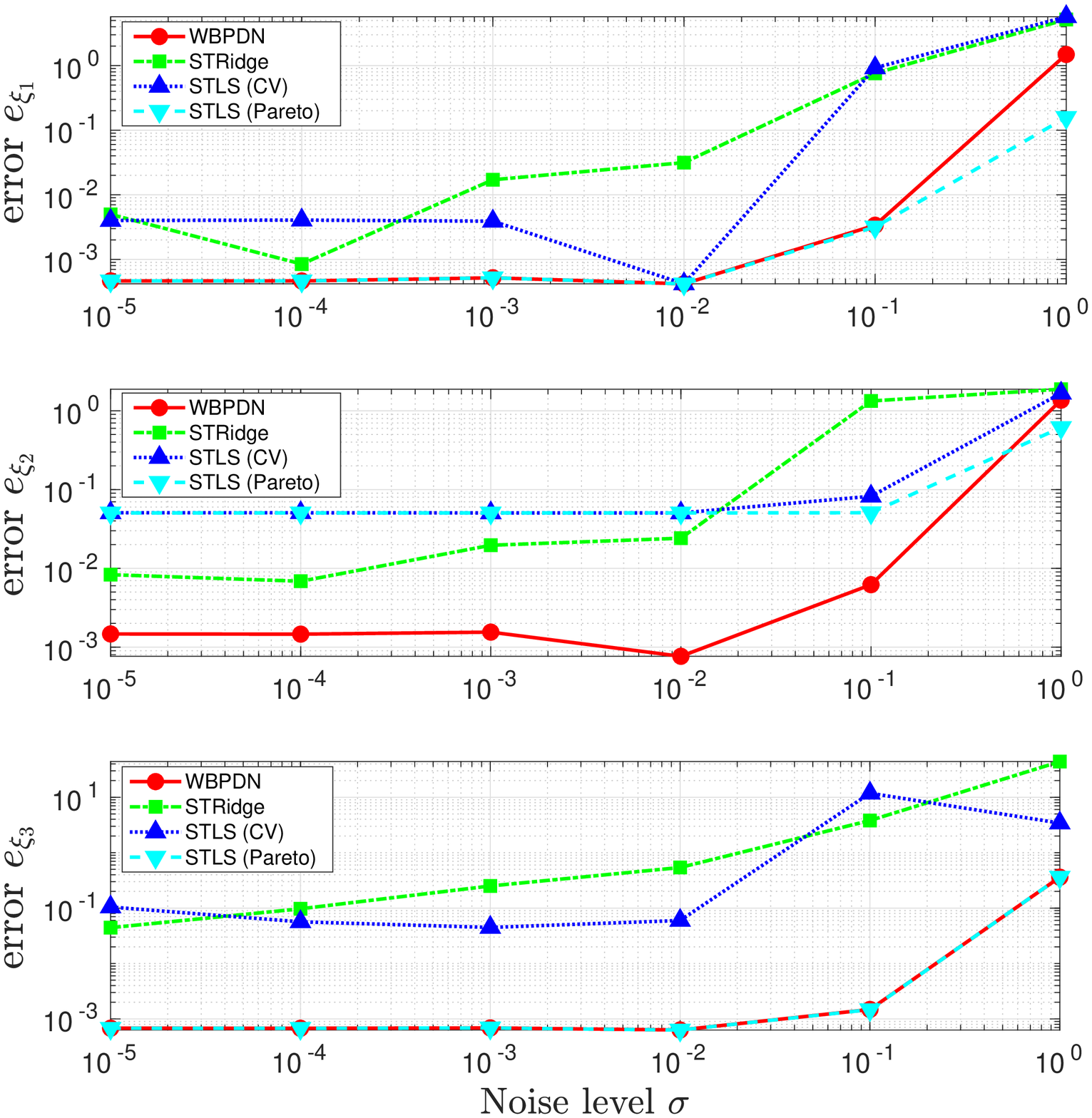}
    \caption{Comparison between WBPDN,  STLS with CV and STLS with Pareto curve algorithms for the Lorenz 63 system for each state variable. Relative solution error with respect to different noise levels $\sigma$.}
    \label{fig:Lorenz63_WBPDNvsSTR}
\end{figure}

Lastly, we compare WBPDN with STLS and STRidge algorithms proposed in~\cite{Rudy2017, Quade2018}. The implementation details and original STLS\footnote{STLS (MATLAB): http://faculty.washington.edu/sbrunton/sparsedynamics.zip} and STRidge\footnote{STRidge (Python): https://github.com/snagcliffs/PDE-FIND} codes used in this article are publicly available online. In the case of STLS, we used two model selection criteria: K-fold CV with $K = 5$ and the $\ell_0$-based Pareto curve, i.e., residual error versus $\ell_0$-norm of $\bm\xi$, and corner point criterion to select the threshold parameter, as suggested in the supporting information for~\cite{Brunton2016}. For CV, we used 200 samples (80\% training and 20\% validation split) over 2 time units, and default tolerances and maximum number of iterations. The state derivatives were computed using the same Tikhonov regularization differentiation algorithm. The results are illustrated in Figure~\ref{fig:Lorenz63_WBPDNvsSTR}. All three algorithms exhibit similar solution error trends with respect to the noise level; a flat region for low noise levels and an error increase for high noise levels. Except for $\sigma = 10^0$ where all methods fail to recover the equations, WBPDN  outperforms STLS and STRidge for all states and noise levels. STLS with $\ell_0$-based Pareto curve produced the closest solution errors to WBPDN; however, it yielded considerable error for the $y$ state.

\subsection{Duffing oscillator}
\label{subsec:Duffing}

\begin{figure}[H]
    \centering
    \includegraphics[trim = 0 0 0 0, clip,width=0.49\textwidth]{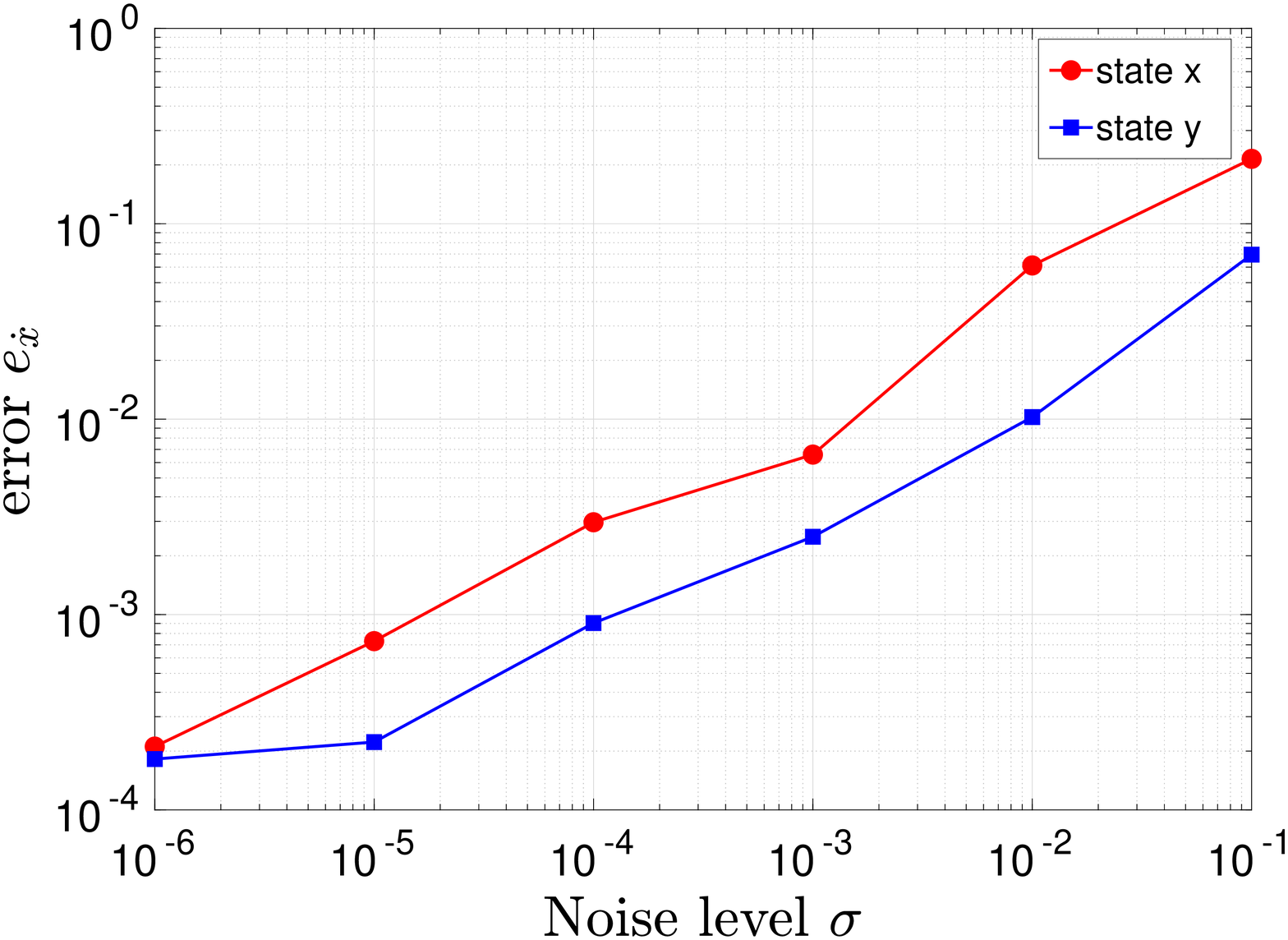}
    \includegraphics[trim = 0 4 0 1, clip,width=0.49\textwidth]{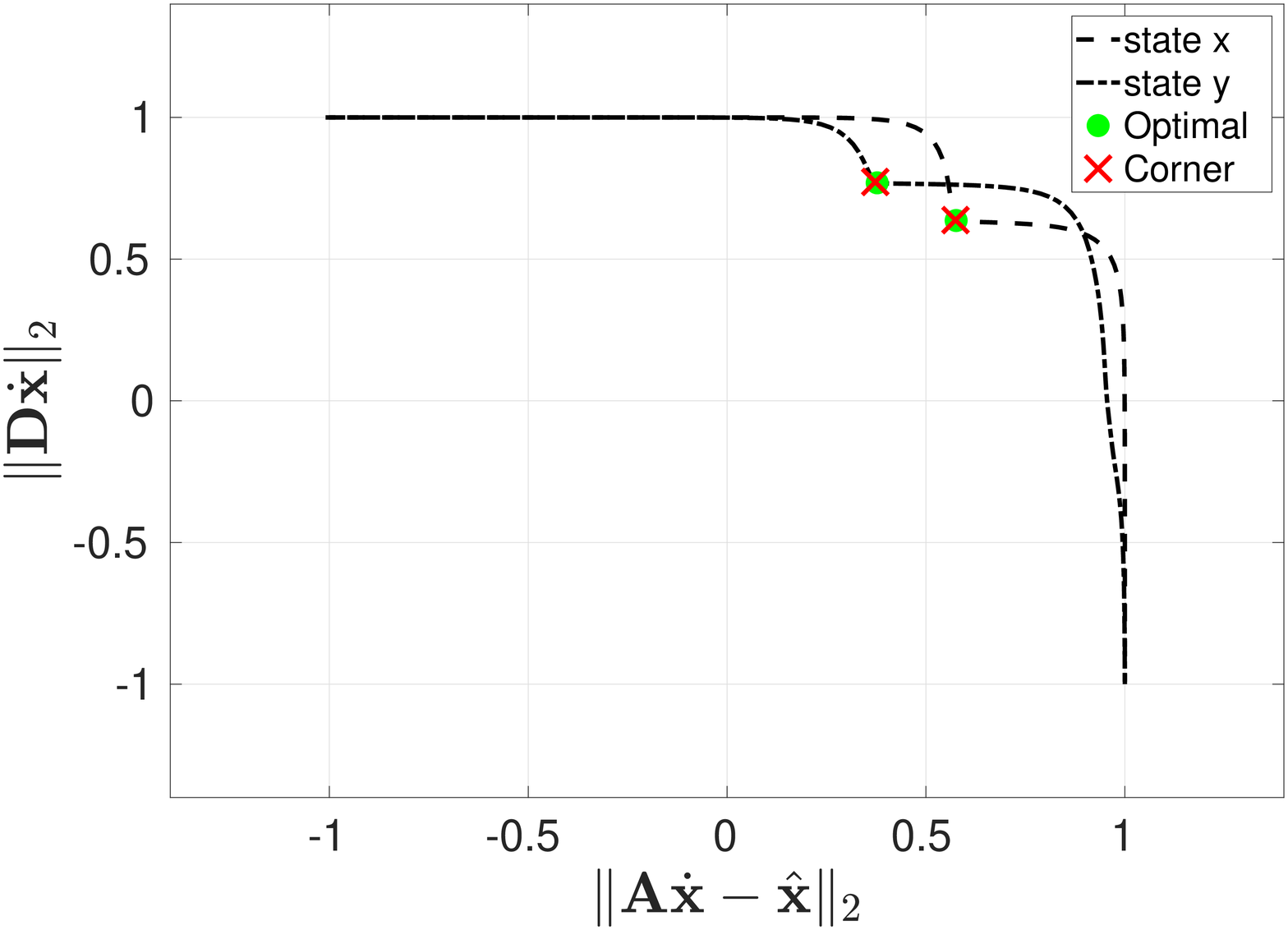}
    \caption{Left: Relative error of the Duffing oscillator state derivatives with respect to different noise levels. Right: Pareto curves for each state of the Duffing oscillator at noise level $\sigma = 10^{-3}$. The signal-to-noise ratios are: $\text{SNR}_x =  52.58~\text{dB}$ and $\text{SNR}_y =  57.44~\text{dB}$.}
    \label{fig:DuffingTikDiff}
\end{figure}

\begin{figure}[H]
    \centering
    \includegraphics[trim = 0 0 0 0, clip,width=.49\textwidth]{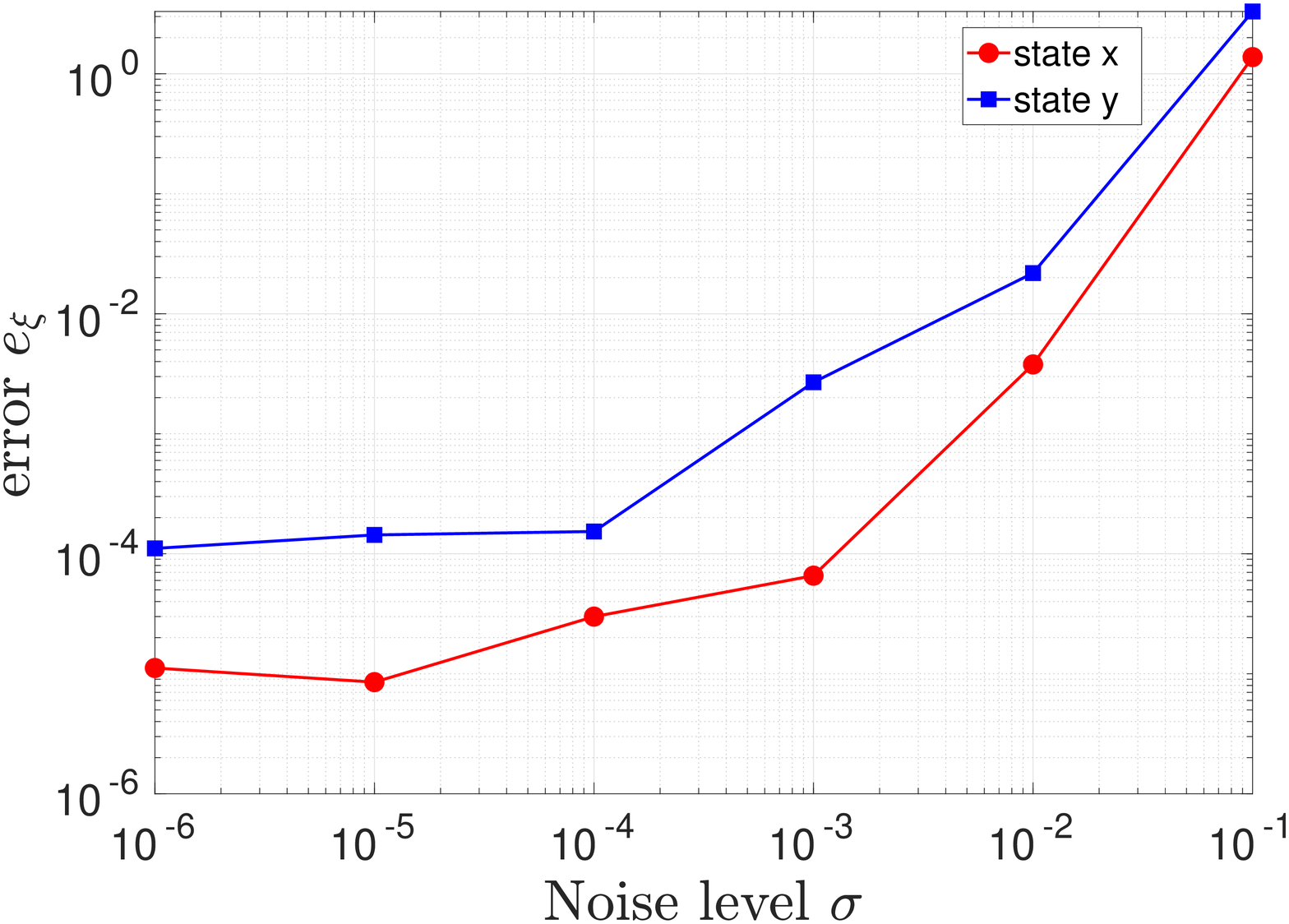}
    \includegraphics[trim = 0 4 0 1, clip,width=.49\textwidth]{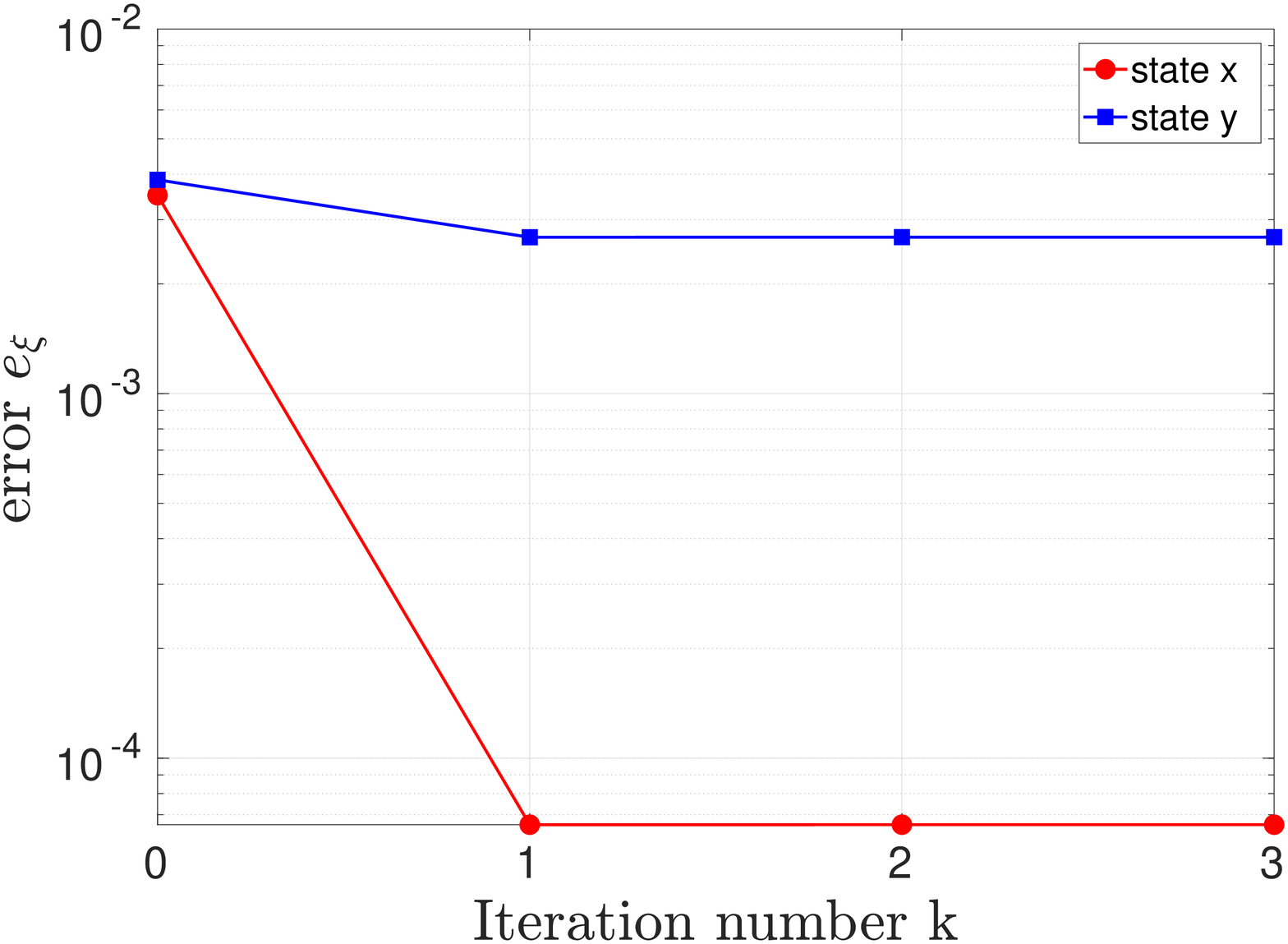}
    \caption{Left: Relative solution error of the Duffing oscillator with respect to different noise levels $\sigma$. Right: Relative solution error of the Duffing oscillator with respect to iteration $k$ at noise level $\sigma = 10^{-3}$.}
    \label{fig:Duffingerrorcoeff}
\end{figure}

\begin{figure}[H]
    \centering
    \includegraphics[trim = 0 0 0 0, clip,width=.49\textwidth]{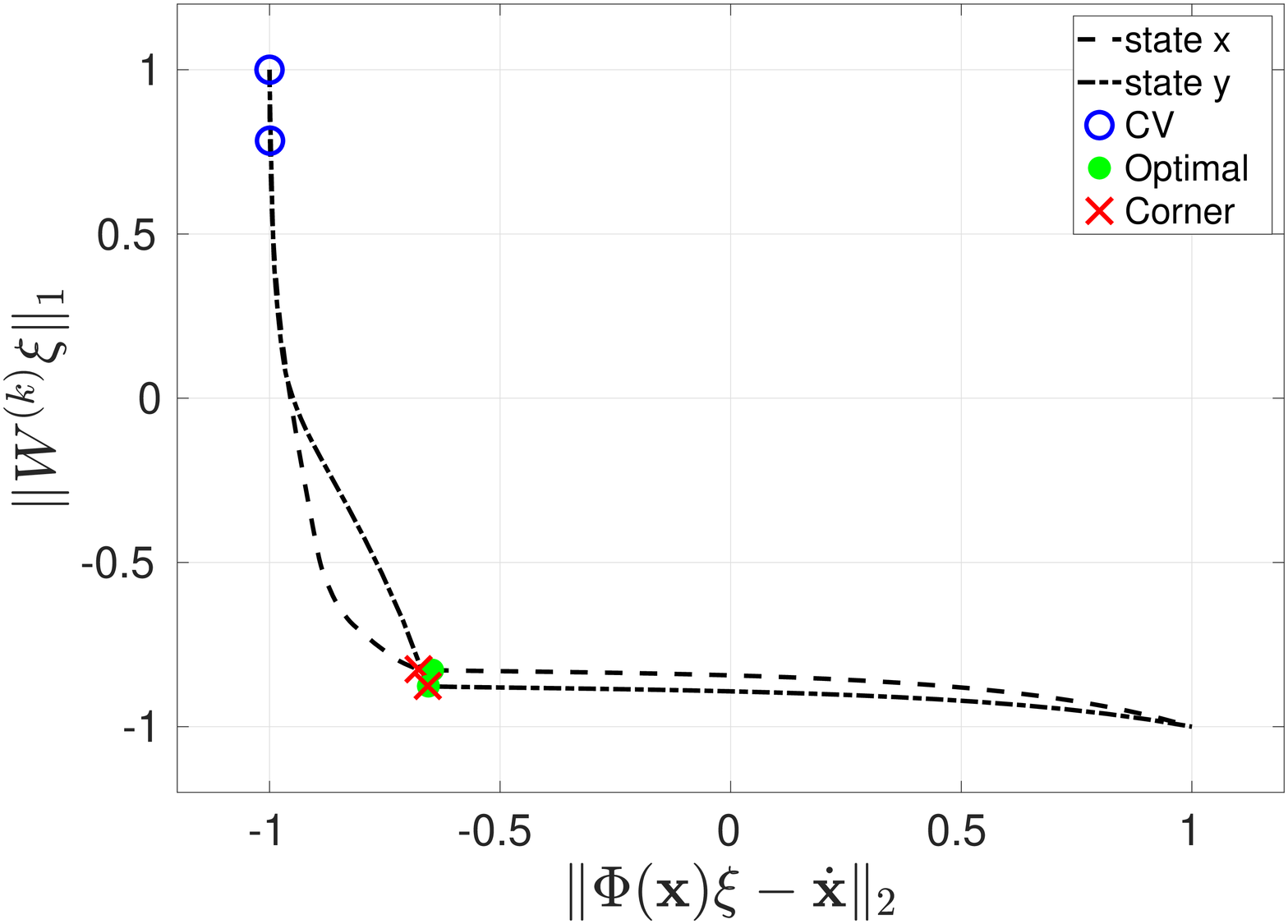}
    \includegraphics[trim = 0 0 0 0, clip,width=.49\textwidth]{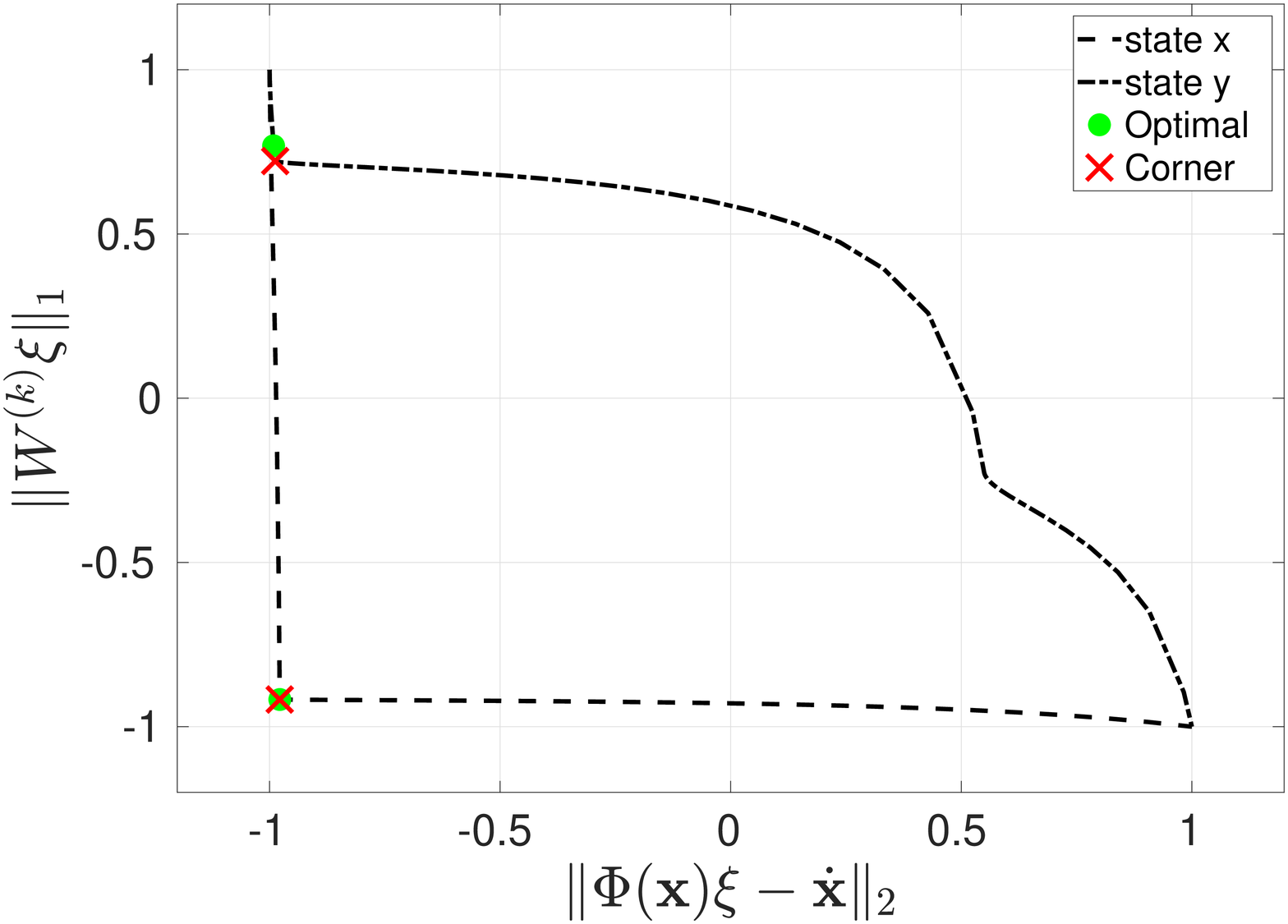}
    \caption{Left: Pareto curves for each state of the Duffing oscillator at noise level $\sigma = 10^{-3}$ and iteration 0. Right: Pareto curves for each state of the Duffing oscillator at noise level $\sigma = 10^{-3}$ and iteration 1.}
    \label{fig:DuffingWBPDNLcurve_it0}
\end{figure}

\begin{figure}[H]
    \centering
    \includegraphics[trim = 0 0 0 0, clip,width=0.63\textwidth]{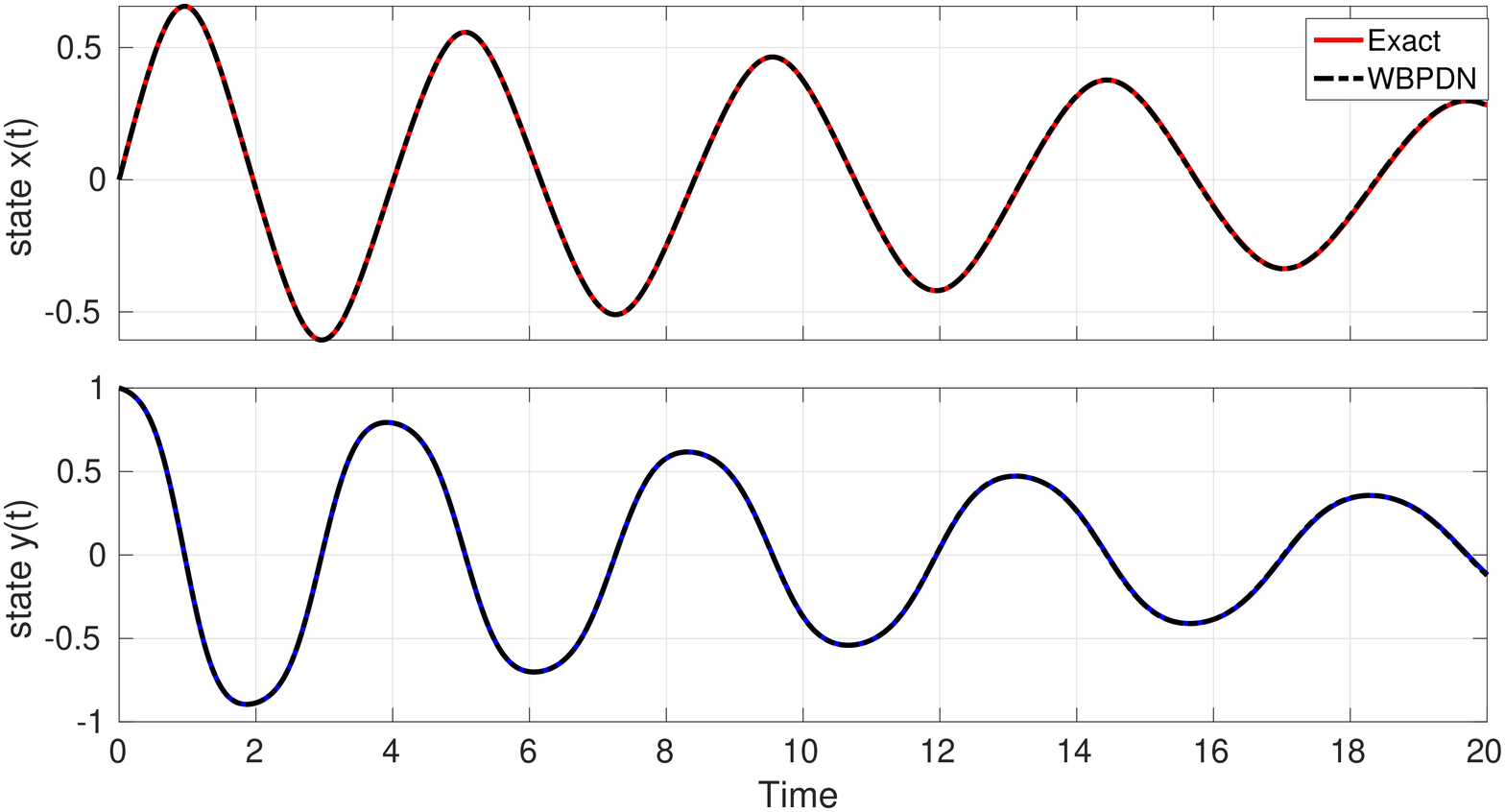}
    \includegraphics[trim = 0 3 0 0, clip,width=0.35\textwidth]{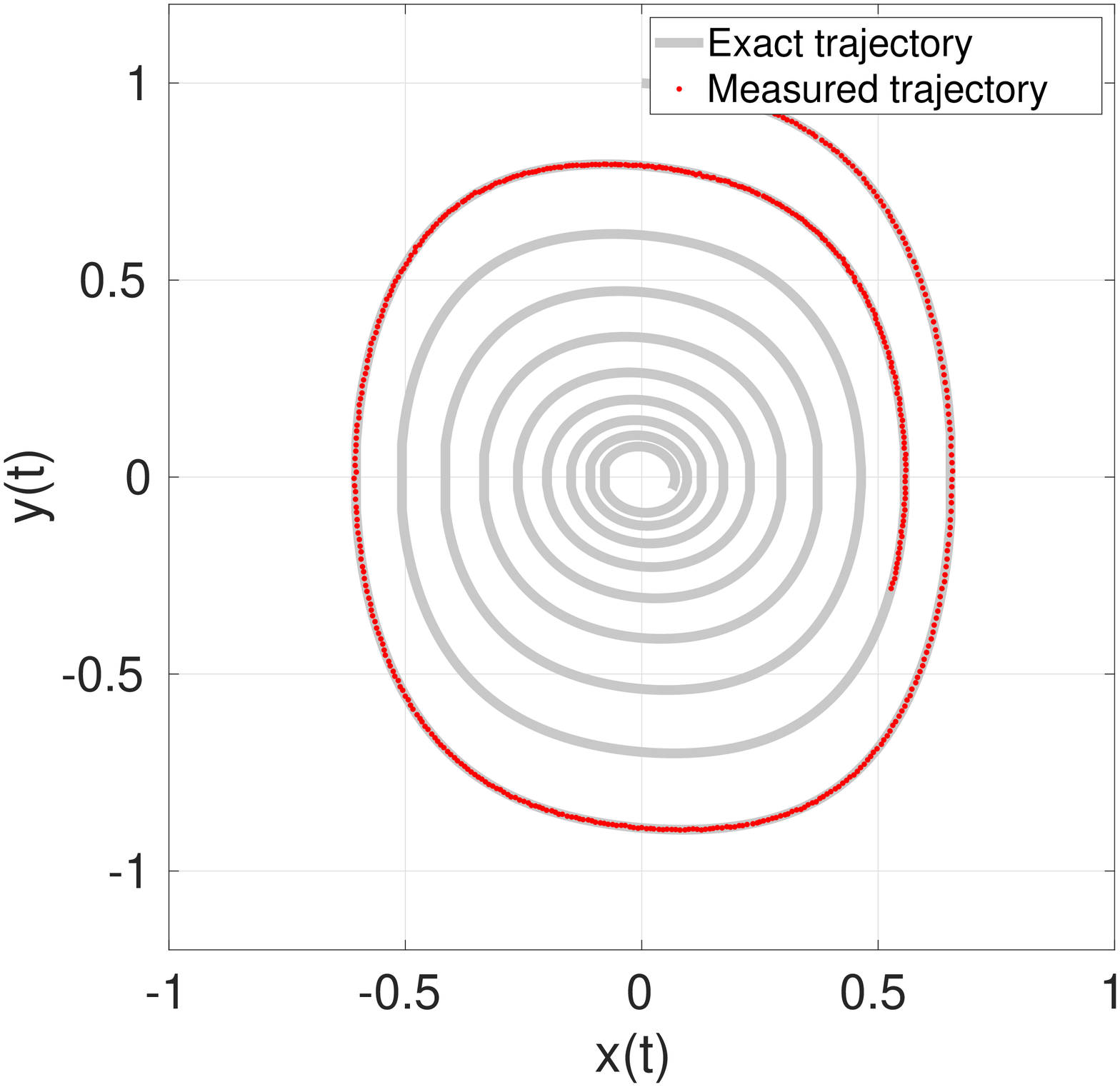}
    \caption{Prediction of Duffing state trajectory by the identified model at noise level $\sigma = 10^{-3}$.
    Left: Prediction of Duffing state trajectory by the identified model at noise level $\sigma = 10^{-3}$. Right: Exact and measured state trajectory of the Duffing system.}
    \label{fig:DuffingWBPDNPredicted}
\end{figure}

The Duffing oscillator  features a cubic nonlinearity and can exhibit chaotic behavior.  Physically, it models a spring-damper-mass system with a spring whose restoring force is $F(\zeta) = -\kappa \zeta - \varepsilon \zeta^3$. When $\varepsilon > 0$, it represents a \textit{hard spring}. Conversely, for $\varepsilon < 0$ it represents a \textit{soft spring}.  The unforced Duffing equation is given by
\begin{equation}\label{eq:DuffingSys}
    \ddot{\zeta} + \gamma \dot{\zeta} + (\kappa + \varepsilon \zeta^2)\zeta  = 0,
\end{equation}
which can be transformed into a first-order system by setting $x = \zeta$ and $y = \dot{\zeta}$, giving
\begin{subequations}
\begin{alignat}{2}
    \dot{x} = y, &\quad x(0) = x_0,\label{eq:DuffingSys_a}\\
    \dot{y} = -\gamma y - \kappa x - \varepsilon x^3, &\quad y(0) = y_0.\label{eq:DuffingSys_b}
\end{alignat}
\end{subequations}
The parameters of the system ($\ref{eq:DuffingSys}$) are set to $\kappa = 1$, $\gamma = 0.1$ and $\varepsilon = 5$, and the initial conditions to $(x_0,y_0) = (1,0)$. For these parameter values, the Duffing oscillator does not present chaotic behavior. The number of state variables is $n = 2$ and the degree of the polynomial basis is set to $d = 4$, yielding $p = 15$ monomial terms. Out of these, only 4 describe the dynamics. In this case, the displacement $x$ and velocity $y$ are measured and we used 200 samples over 2 time units, from $t$ = 0.1 to $t$ = 2.1, to recover the system. The errors of the state variable derivatives and the Pareto curves are shown in the left and right panels of Figure \ref{fig:DuffingTikDiff}, respectively.
As shown, the $\lambda$ associated with the corner point agrees with the optimal one, yielding accurate state derivatives for each noise level. The effect of noise on the solution error, presented in Figure \ref{fig:Duffingerrorcoeff}, is consistent with previous results. As in the Lorenz problem, the WBPDN converges in only one iteration for $\sigma = 10^{-3}$. 
Figure \ref{fig:DuffingWBPDNLcurve_it0} shows the effectiveness of the corner point criterion to find a regularization parameter close to the optimal, as opposed to the CV approach, which yields quite suboptimal $\lambda$ estimates. The predicted trajectory matches the exact one, as illustrated in Figure~\ref{fig:DuffingWBPDNPredicted}.

\begin{figure}[H]
    \centering
    \includegraphics[trim = 0 0 0 0, clip,width=0.8\textwidth]{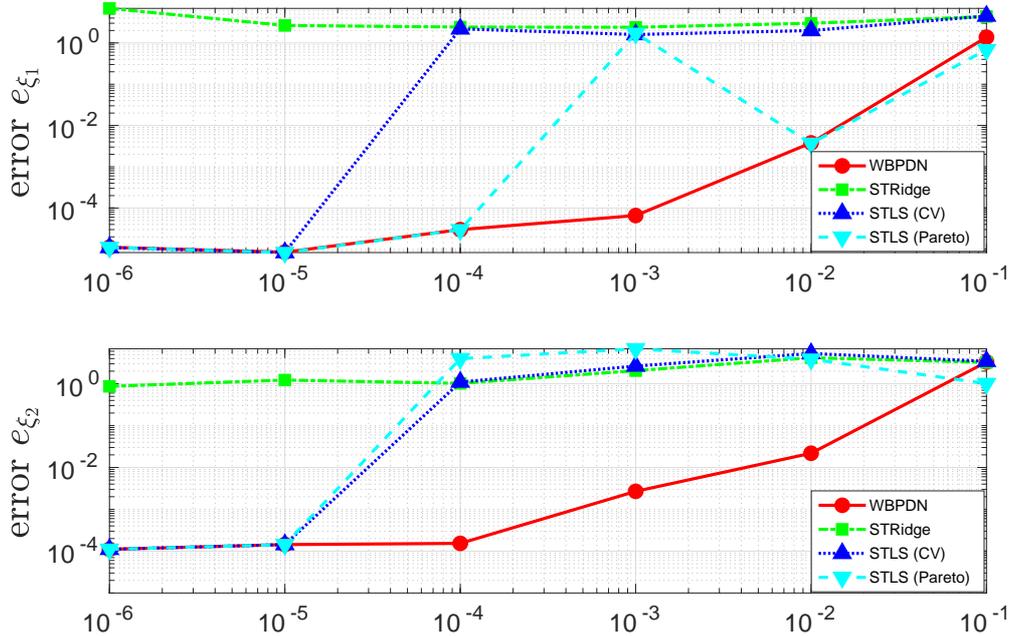}
    \caption{Comparison between WBPDN,  STLS with CV and STLS with Pareto curve algorithms for the Duffing oscillator for each state variable. Relative solution error with respect to different noise levels $\sigma$.}
    \label{fig:Duffing63_WBPDNvsSTRvsSTLS}
\end{figure}

In this example, we also compared the performance of STLS, STRidge and WBPDN in recovering the coefficient vector $\bm{\xi}$. As in the Lorenz 63 example, we used an 80\% training and 20\% validation data split for both STLS with CV and STRidge. Figure~\ref{fig:Duffing63_WBPDNvsSTRvsSTLS} displays the error in $\bm\xi$ at different noise levels. Overall, WBPDN outperforms both STLS and STRidge. In the case of STRidge, the solution happens to be unstable even for low noise levels. This may be because the default regularization parameters of the algorithm are not suitable for this problem and need careful tuning.
In the STLS case, we also used CV and the Pareto criterion to set the threshold parameter $\gamma$. On the one hand, we noticed that the CV function, i.e., mean residual versus regularization parameter, is flat for the region where the regularization parameter with minimum solution error was located. This yields a minimum mean residual regularization parameter far from the optimal one, and therefore, inaccurate solution. On the other hand, we noticed that the corner point of the Pareto curve may not be well defined. Therefore, any algorithm trying to find the corner of the Pareto curve is deemed to fail. This fact may explain the sudden jumps in Figure~\ref{fig:Duffing63_WBPDNvsSTRvsSTLS} for the STLS (Pareto) case.

\subsection{Van der Pol oscillator}
\label{subsec:VanderPol}
\begin{figure}[H]
    \centering
    \includegraphics[trim = 0 0 0 0, clip,width=0.49\textwidth]{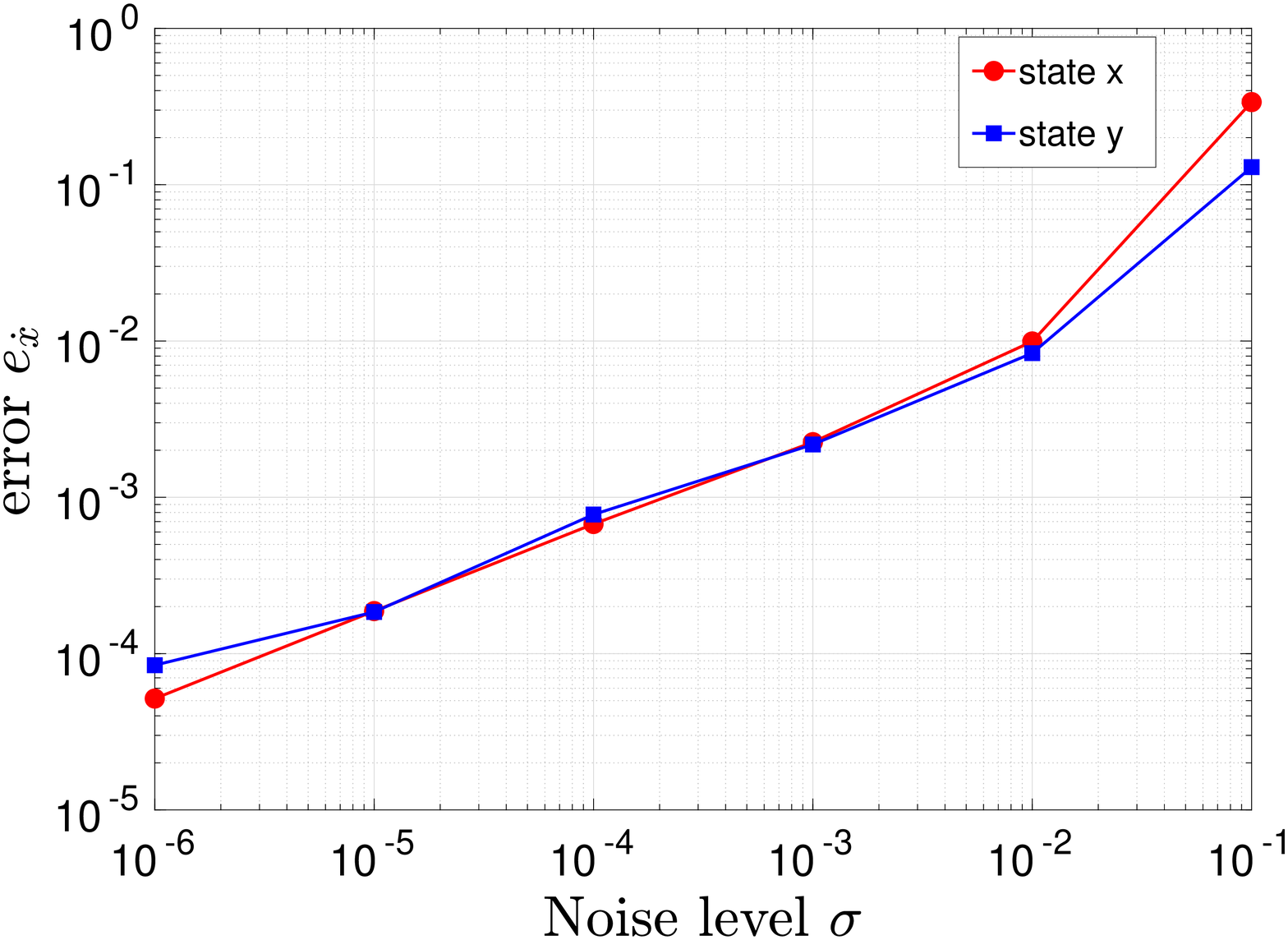}
    \includegraphics[trim = 225 25 275 50, clip,width=0.49\textwidth]{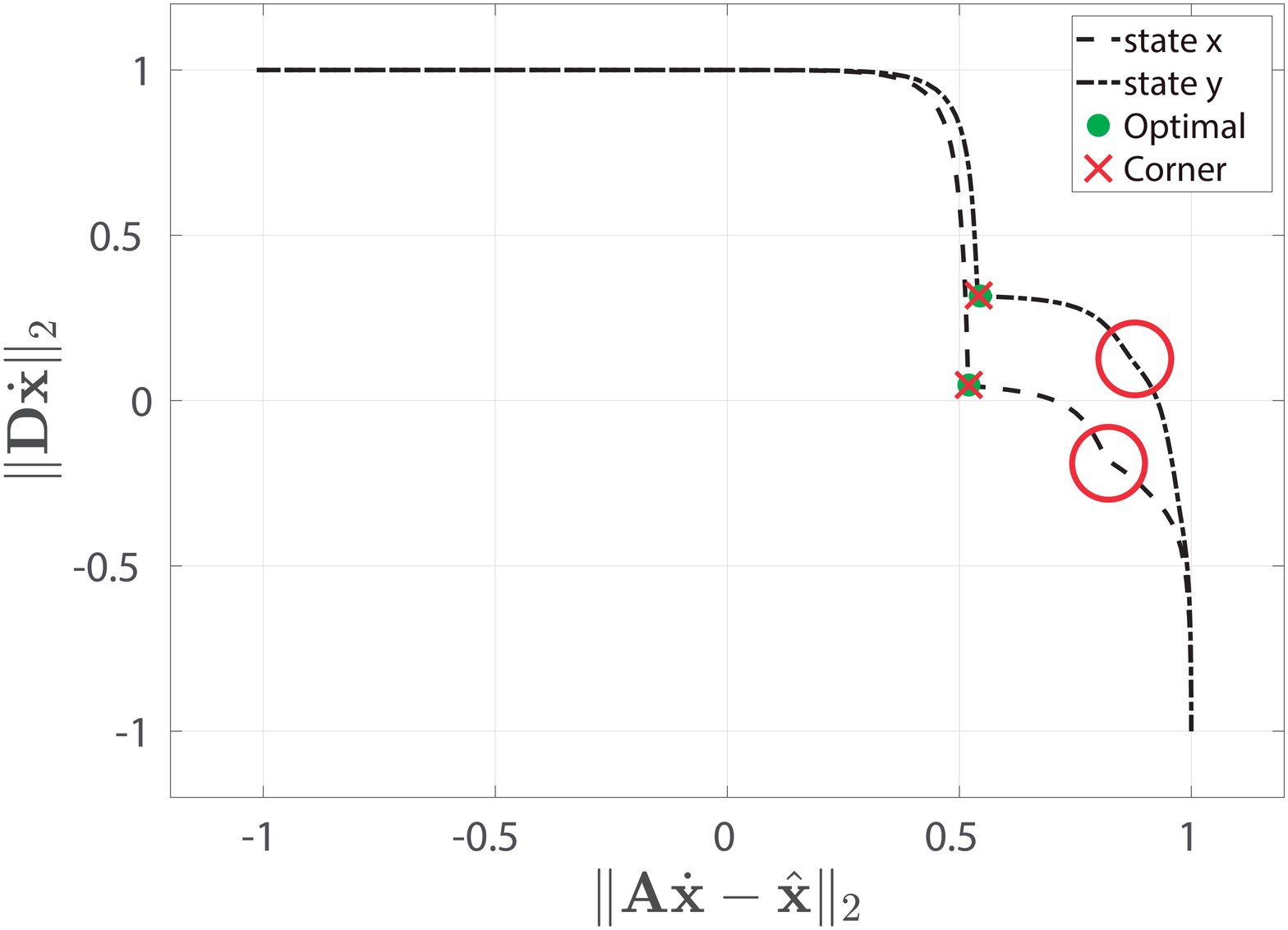}
    \caption{Left: Relative error of the Van der Pol state derivatives with respect to different noise levels. Right: Pareto curves for each state of the Van der Pol oscillator at noise level $\sigma = 10^{-3}$. The red circles indicate additional corner points. The signal-to-noise ratios are: $\text{SNR}_x = 62.41~\text{dB}$ and $\text{SNR}_y = 61.15~\text{dB}$.}
    \label{fig:VanderPolTikDiff}
\end{figure}

\begin{figure}[H]
    \centering
    \includegraphics[trim = 0 0 0 0, clip,width=.49\textwidth]{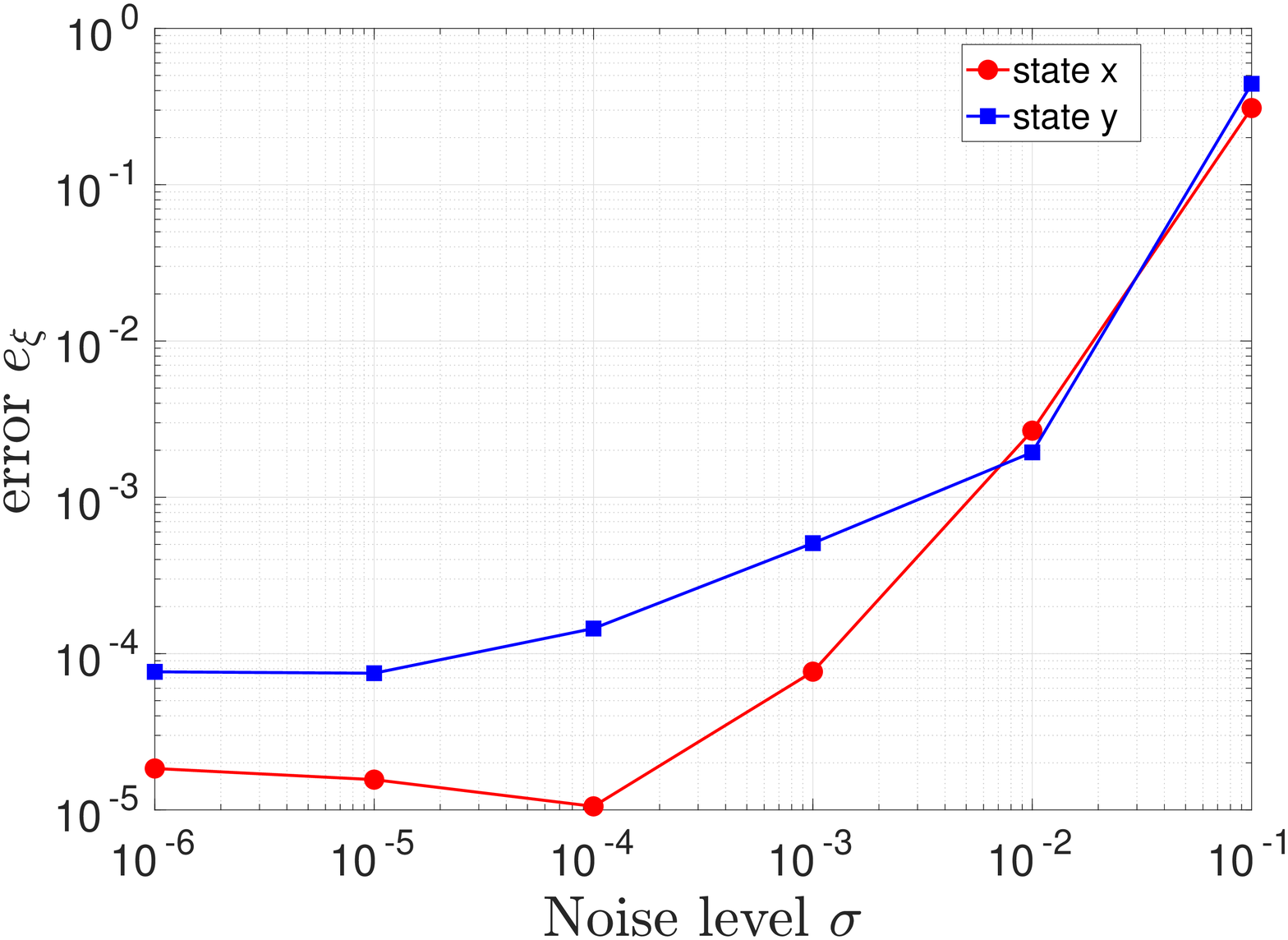}
    \includegraphics[trim = 0 0 0 0, clip,width=.49\textwidth]{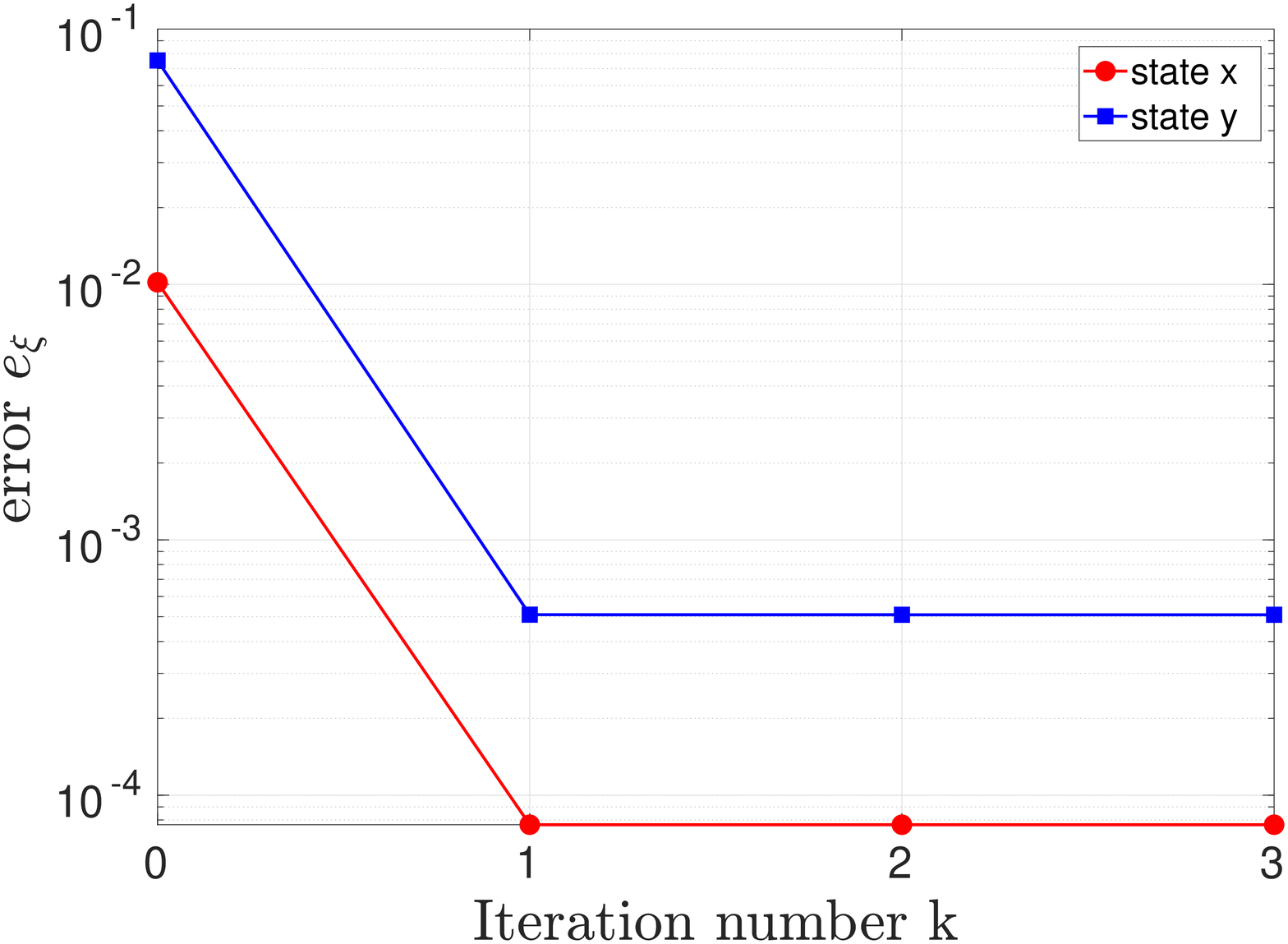}
    \caption{Left: Relative solution error of the Van der Pol oscillator with respect to different noise levels $\sigma$. Right: Relative solution error of the Van der Pol oscillator with respect to iteration $k$ at noise level $\sigma = 10^{-3}$.}
    \label{fig:VanderPolerrorcoeff_and_convergence}
\end{figure}

\begin{figure}[H]
    \centering
    \includegraphics[trim = 0 0 0 0, clip,width=0.6\textwidth]{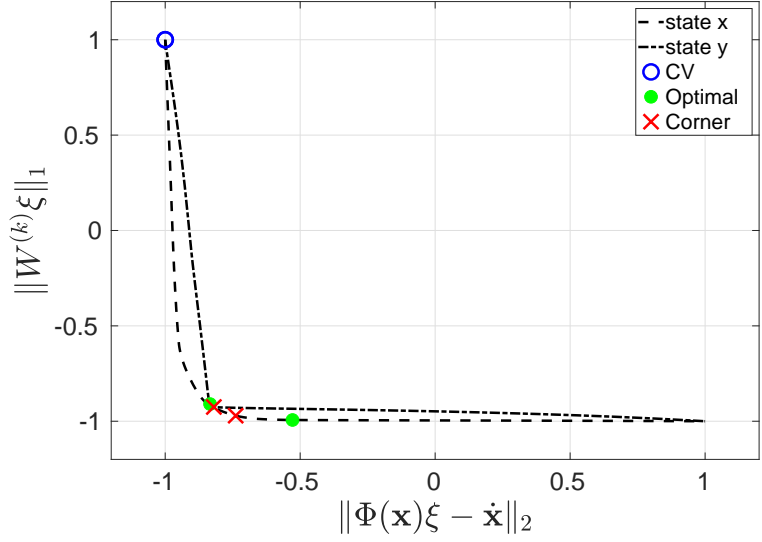}
    \caption{Pareto curves for each state of the Van der Pol oscillator at noise level $\sigma = 10^{-3}$ and iteration 0.}
    \label{fig:VanderPolWBPDNLcurve_it0}
\end{figure}

\begin{figure}[H]
    \centering
    \includegraphics[trim = 0 0 0 0, clip,width=0.63\textwidth]{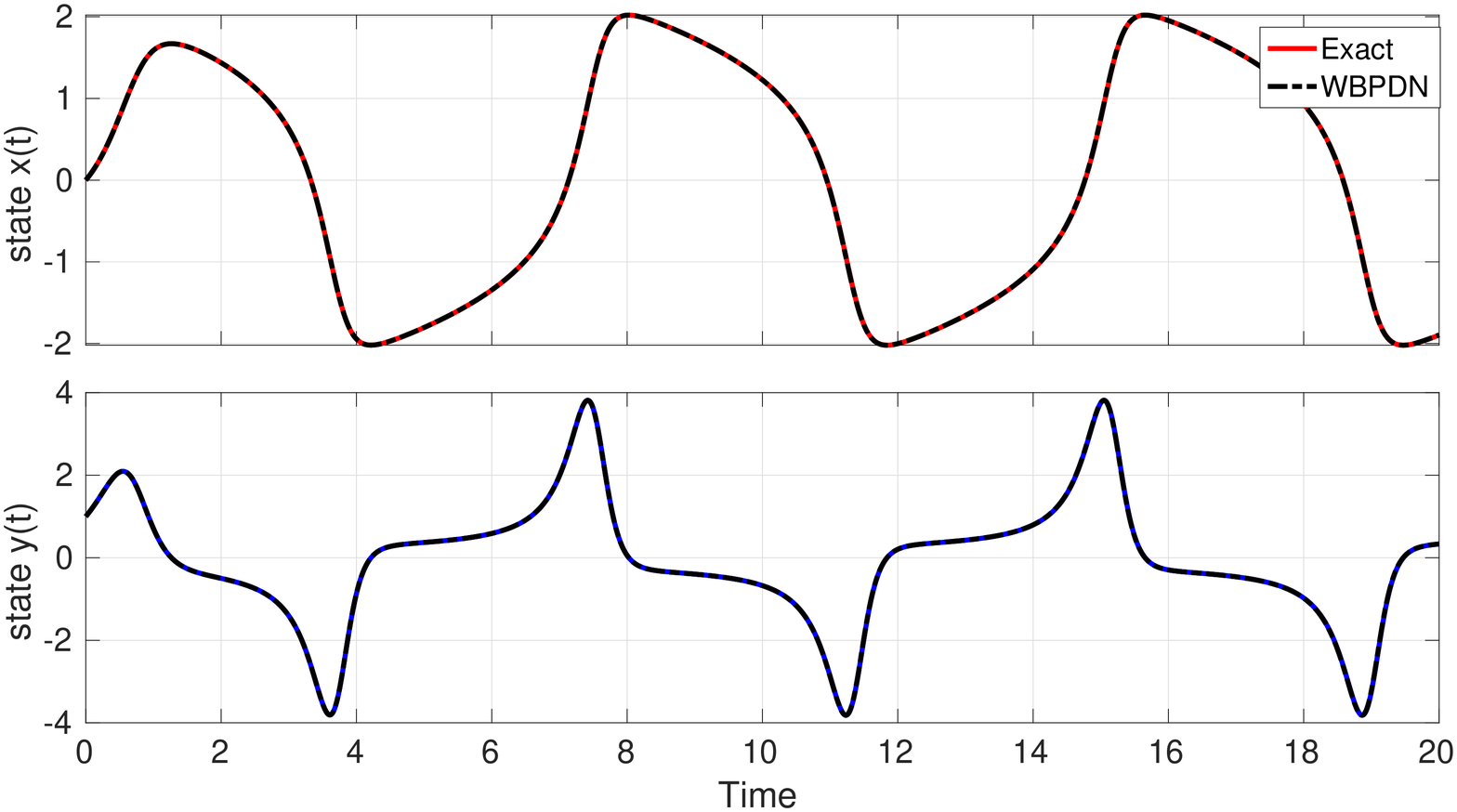}
    \includegraphics[trim = 0 3 0 0, clip,width=0.35\textwidth]{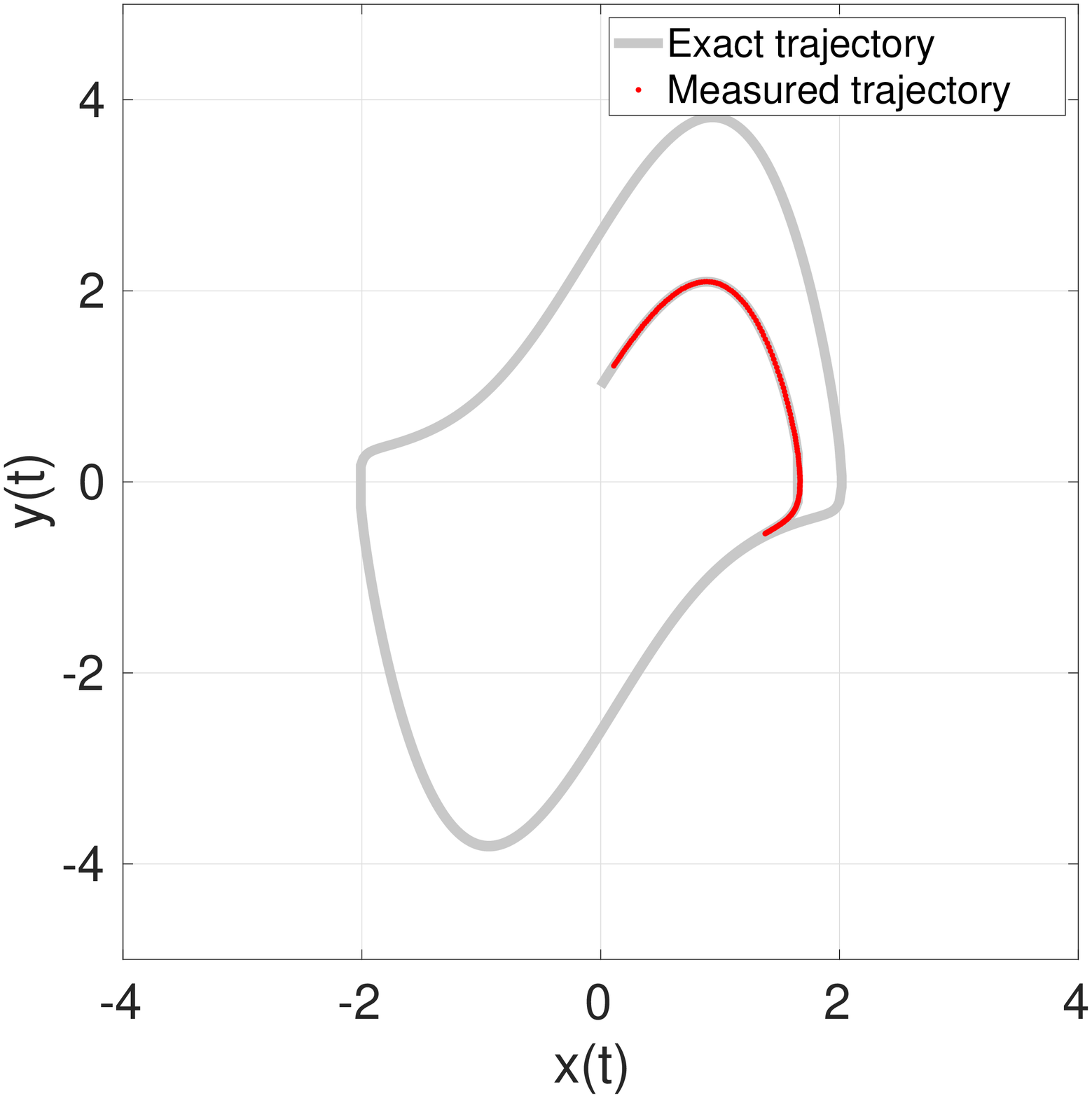}
    \caption{Left: Prediction of Van der Pol state trajectory by the identified model at noise level $\sigma = 10^{-3}$. Right: Exact and measured trajectory for the Van der Pol oscillator at $\sigma = 10^{-3}$.}
    \label{fig:VanderPolWBPDNPredicted}
\end{figure}

The Van der Pol model is a second-order oscillator with a nonlinear damping term. It was originally proposed by Van der Pol as a model to describe the oscillation of a triode in an electrical circuit~\cite{VanDerPol1920}. The Van der Pol model exhibits a limit cycle behavior around the origin. The governing equation for this system is given by the following ODE
\begin{equation}\label{eq:VanDerPolSys}
    \ddot{\zeta} + (\gamma + \varepsilon\zeta^2)\dot{\zeta} + \kappa\zeta  = 0,
\end{equation}
which can be transformed into a first-order system as, 
\begin{subequations}
\begin{alignat}{2}
    \dot{x} = y, &\quad x(0) = x_0,\label{eq:VanDerPolSys_a}\\
    \dot{y} = -\kappa x -\gamma y - \varepsilon x^2y, &\quad y(0) = y_0.\label{eq:VanDerPolSys_b} 
\end{alignat}
\end{subequations}
The parameters of the system ($\ref{eq:VanDerPolSys}$) are set to $\kappa = 1$, $\gamma = 1$ and $\varepsilon = 2$, and the initial conditions to $(x_0,y_0) = (1,0)$. As in the Duffing system, the sparsity of this system is 4, the number of state variables is $n = 2$, and the degree of the polynomial basis is set to $d = 4$, giving $p = 15$ monomial terms. Again, the number of samples used in this example is 200 over 2 time units, ranging from $t = 0.1$ to $t = 2.1$.
The performance of Tikhonov regularization differentiation with respect to different noise levels is shown in Figure~\ref{fig:VanderPolTikDiff} (left), whereas Figure~\ref{fig:VanderPolTikDiff} (right) shows the Pareto curves with corner points matching optimal $\lambda$ values. This case illustrates a difficulty in locating {\it global} corner points instead of {\it local} ones. The Pareto curves for both state variables present two corner points for the range of $\lambda$ used to generate them. The convergence to the wrong corner point can be avoided by selecting the one with higher curvature, by restricting the range of $\lambda$ to some region close to the estimated residual given by the discrepancy principle~\cite{Morozov1966}, or by visually inspecting the Pareto curve plot and picking the corner point manually. Fortunately, the algorithm proposed in~\cite{Cultrera2016} is robust enough to select the right corner in this case.
The relative solution error agrees with the previous cases, as shown in Figure~\ref{fig:VanderPolerrorcoeff_and_convergence}. For low noise levels, the discretization errors on the state derivatives dominate the overall error of the solution. As we increase the noise level, the solution error is dominated by noise and starts to grow rapidly. Again, only one iteration is sufficient to converge and reduce the BPDN error by two orders of magnitude (for $\sigma = 10^{-3}$).
We observe from Figure \ref{fig:VanderPolWBPDNLcurve_it0} that, similar to previous cases, the Pareto curve criterion leads to better estimates of the optimal regularization parameter $\lambda$, as compared to the CV approach. Similar to previous examples, the predicted and exact trajectories match well for $\sigma  = 10^{-3}$; see Figure \ref{fig:VanderPolWBPDNPredicted}. 

\begin{figure}[H]
    \centering
    \includegraphics[trim = 0 0 0 0, clip,width=0.8\textwidth]{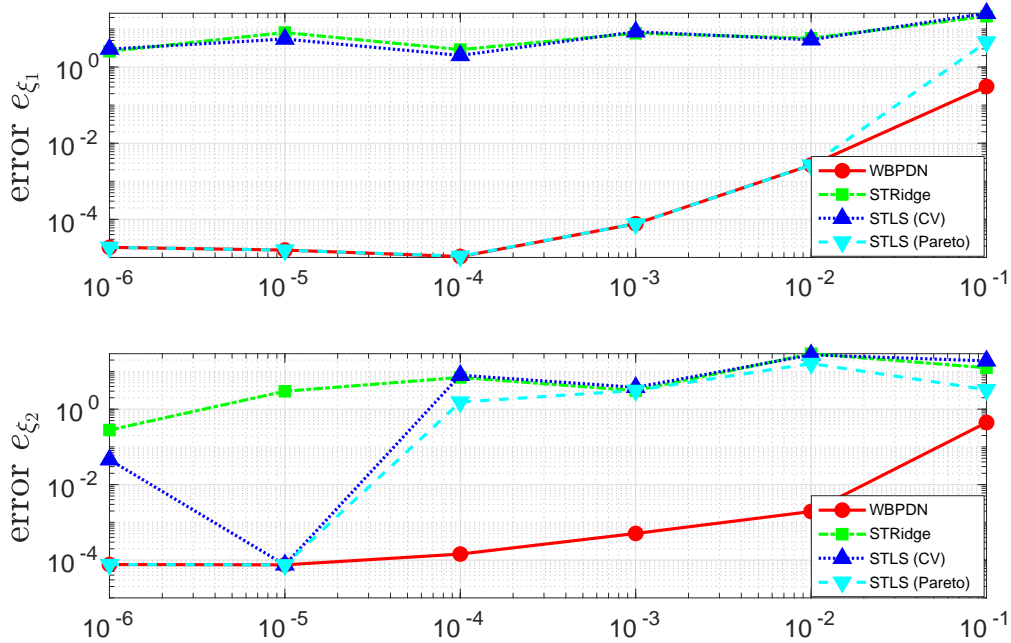}
    \caption{Comparison between WBPDN,  STLS with CV and STLS with Pareto curve algorithms for the Van der Pol oscillator for each state variable. Relative solution error with respect to different noise levels $\sigma$.}
    \label{fig:Lorenz63_WBPDNvsSTRvsSTLS}
\end{figure}

Finally, the comparison among WBPDN, STRidge and STLS is displayed in Figure~\ref{fig:Lorenz63_WBPDNvsSTRvsSTLS}. Similarly to previous examples, WBPDN yields the most accurate solution error. In this example, STRidge and STLS with CV produce inaccurate solutions for almost all noise levels. STLS with Pareto curve yields almost identical results as WBPDN for the $x$ state variable. However, in general, it fails to recover the $y$ state solution accurately.
\subsection{Single degree of freedom spring-mass system}
\label{subsec:springmass}

The single degree of freedom spring-mass system is a second-order harmonic oscillator that, when perturbed from its equilibrium, experiences a restoring force proportional to its displacement. In the absence of damping forces, the total energy of the system is conserved, resulting in a continuous transfer between potential and kinetic energy. The unforced harmonic oscillator is given by
\begin{equation}
    m\ddot{\zeta} + k\zeta = 0,
\end{equation}
where $m$, $k$ and $\zeta$ are the mass, stiffness and displacement, respectively. The system can be transformed to a first order system as
\begin{subequations}
\begin{alignat}{2}
    \dot{x} = y, &\quad x(0) = x_0, \label{eq:spring-mass_a}\\
    \dot{y} = -\omega^2x, &\quad y(0) = y_0, \label{eq:spring-mass_b}
\end{alignat}
\end{subequations}
where $x = \zeta$, $y = \dot{\zeta}$, and $\omega = \sqrt{\frac{k}{m}}$ is the natural frequency of the system. In this example, the mass and stiffness are set to $m = 1 $, $k = 10$, and the initial conditions to $(x_0,y_0) = (1,0)$. The number of state variables is $n = 2$ and the sparsity of this system in the polynomial basis is 2. The number of samples used in this example is 200 over 2 time units, from $t = 0.1$ to $t = 2.1$. 

\begin{figure}[H]
    \centering
    \includegraphics[trim = 330 0 350 0, clip = true,width=0.49\linewidth]{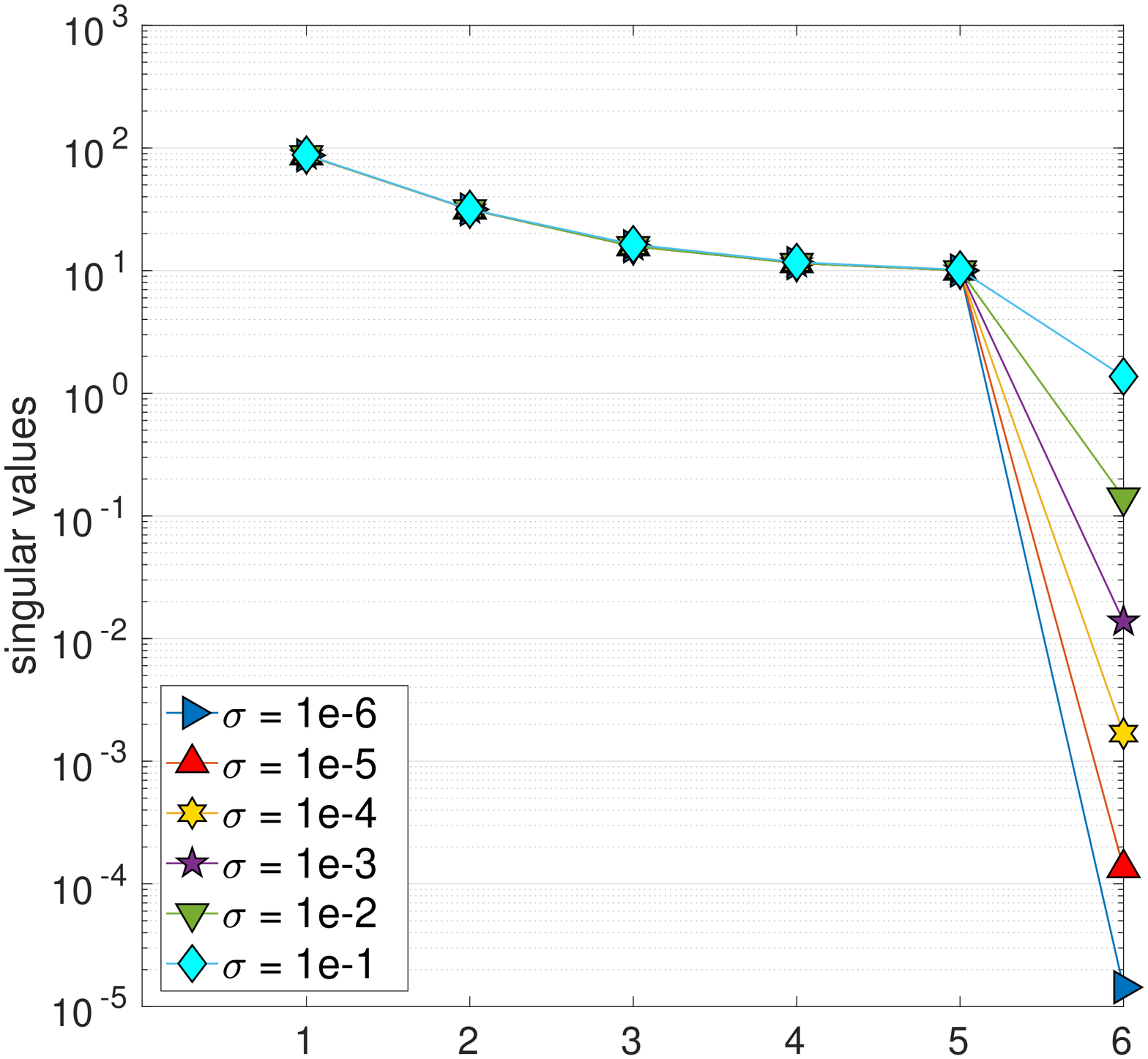}
    \includegraphics[trim = 330 0 350 0, clip = true,width=0.47\linewidth]{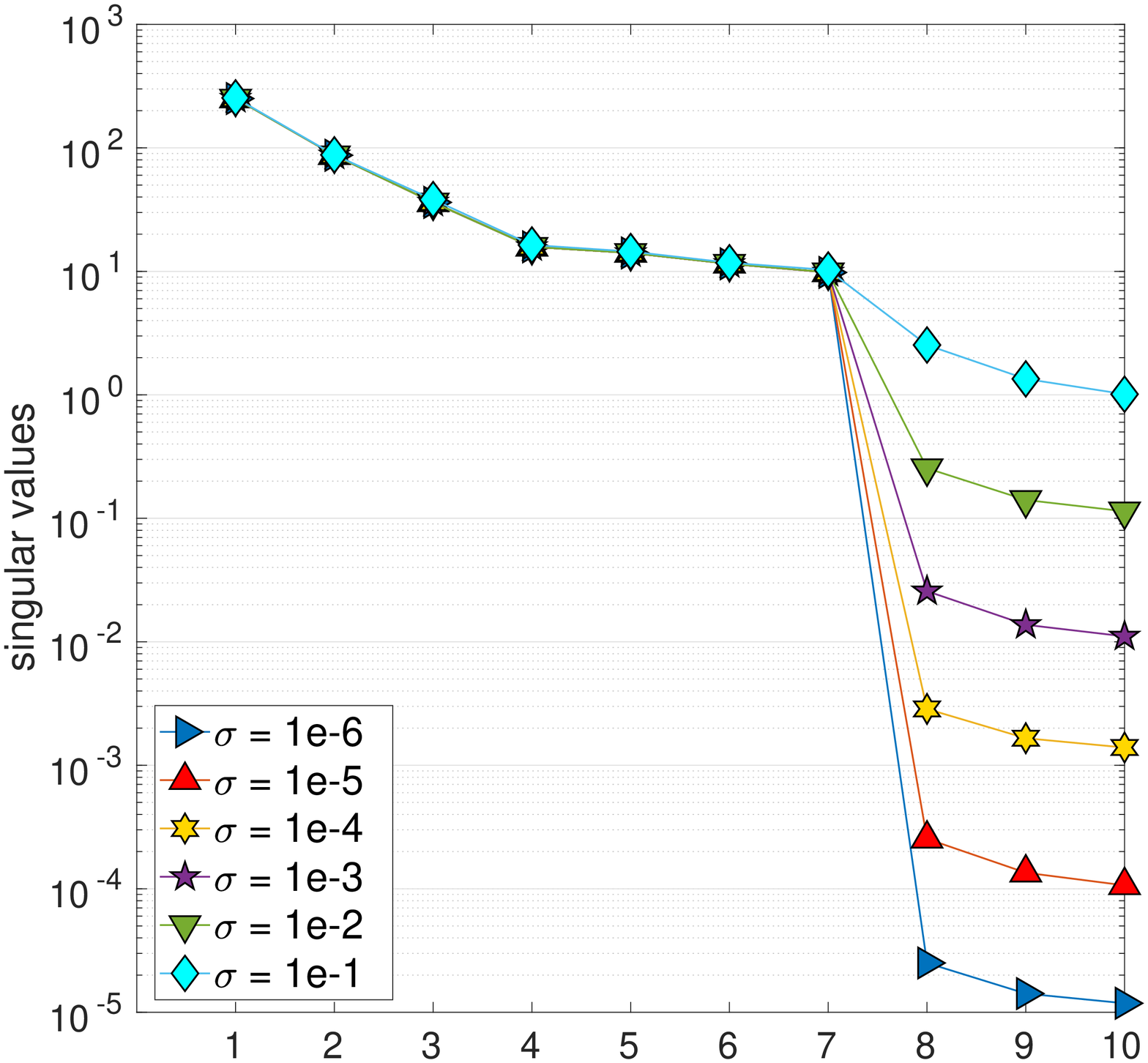}
    \caption{Singular values of $\bm{\Phi}(\mathbf{x})$ for polynomial degree $d = 2$ (left) and $d = 3$ (right) at different noise levels $\sigma$. Note that the gap in the singular values becomes less evident at higher noise levels.}
    \label{fig:SpringMass_svDecay}
\end{figure}

\begin{figure}[H]
    \centering
    \includegraphics[trim = 320 0 320 0, clip = true,width=0.48\linewidth]{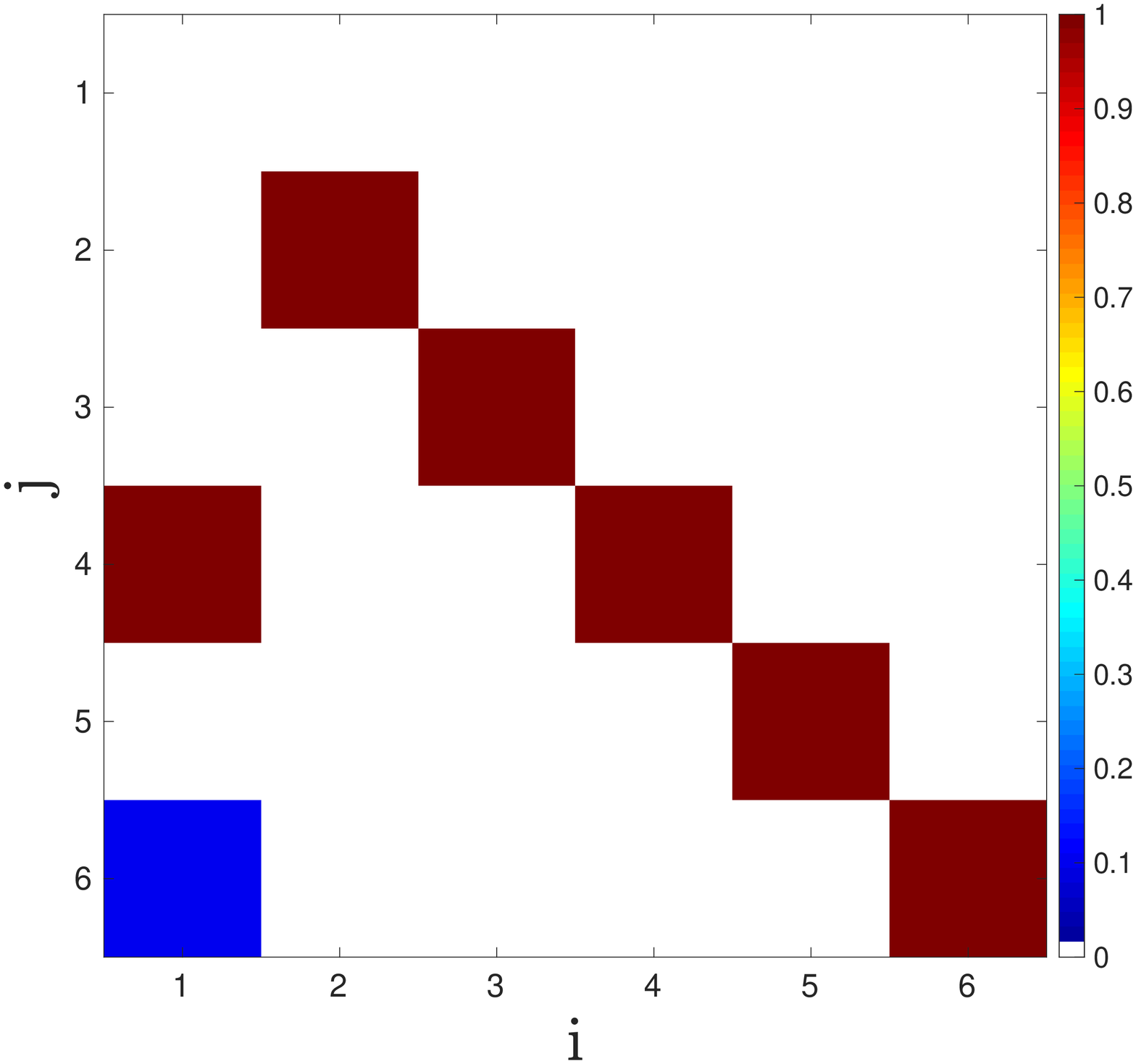}
    \includegraphics[trim = 320 0 320 0, clip = true,width=0.48\linewidth]{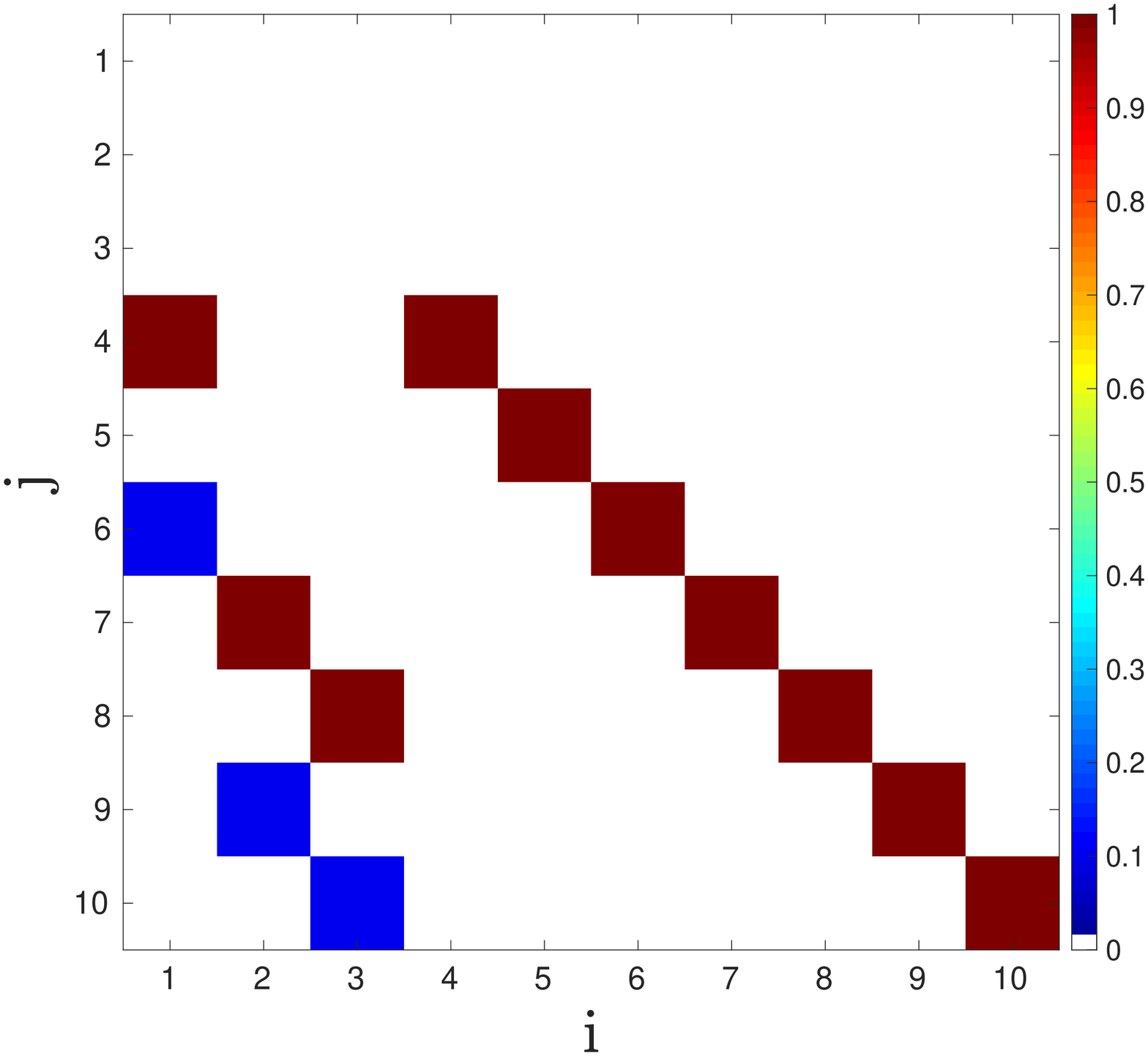}
    \caption{Linear dependence matrix for $d = 2$ (Left) and $d = 3$ (Right) at noise level $\sigma = 10^{-3}$. The horizontal axis is the index $i$ of each column of $\bm{\Phi}(\mathbf{x})$ and the vertical axis is the index $j$ of columns it depends on. White blocks show no dependency and the colored blocks show dependency with a magnitude obtained from the ID coefficient matrix $\mathbf{C}$ in (\ref{eq:rankrfactSC}).}
    \label{fig:SpringMass_LDMatrix_dCases}
\end{figure}

\begin{figure}[H]
    \centering
    \includegraphics[trim = 0 0 0 0, clip,width=0.43\textwidth]{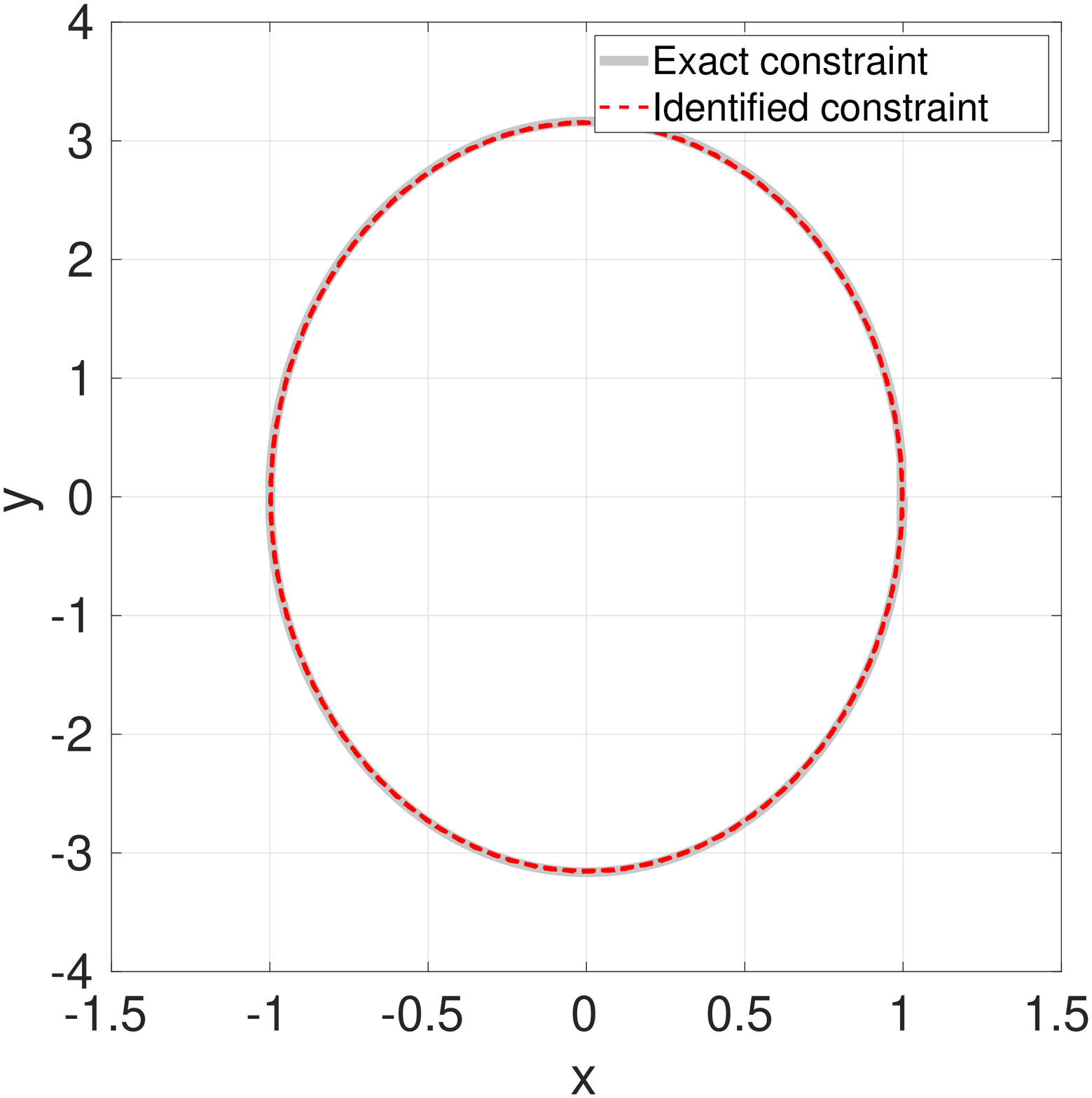}
    \caption{Exact and identified energy constraint for the spring-mass problem with $d = 2$ at noise level $\sigma = 10^{-3}$.}
    \label{fig:SpringMass_constraintplot}
\end{figure}

The total energy  $\mathcal{E}$ is the sum of kinetic $\mathcal{T}$ and potential $\mathcal{V}$ energies given by
\begin{equation}\label{eq:cons_energy_springmass}
    \mathcal{E} =  \mathcal{V} + \mathcal{T} = \frac{1}{2}k x^2 + \frac{1}{2}m y^2,
\end{equation}
which represents the equation of an ellipse centered at the origin and semi-axis given by $a = \sqrt{2\mathcal{E}/k}$ and $b = \sqrt{2\mathcal{E}/m}$. It is straight-forward to verify that the corresponding constraint function $g(\mathbf{x})$ is given by
\begin{equation}\label{eq:constraintmanifold_sdof}
    g(\mathbf{x}) = -1.0 \phi_1(\mathbf{x}) + 1.0 \phi_4(\mathbf{x}) + 0.1 \phi_6(\mathbf{x}),
\end{equation}
where $\phi_i(\mathbf{x}) \in \{1, x, y, x^2, xy, y^2, x^3, x^2y, xy^2, y^3,\dots\},\,\,i = 1,\dots,p$, i.e., $\phi_4(\mathbf{x}) = x^2$.
Figure~\ref{fig:SpringMass_svDecay} displays the singular values of $\bm{\Phi}(\mathbf{x})$ for polynomial degrees $d = 2$ (left) and $d = 3$ (right) to find the true ranks $r=5$ and $r=7$, respectively. As shown, the gap between the 5th and 6th largest singular values in the $d = 2$ case and the 7th and the 6th in the $d = 3$ case becomes smaller as we increase the noise level $\sigma$.
To identify the energy constraint, we then follow the procedure outlined in Algorithm~\ref{alg:ID} in Section~\ref{subsec:constrainedsystems} with the calculated numerical ranks. In the case of $d=2$, the ID of $\bm{\Phi}(\mathbf{x})$ leads to $\mathcal{J}=\{2,\dots,6\}$ as the indices of independent columns, i.e., first column of $\bm{\Phi}(\mathbf{x})$ depends linearly on the rest. This dependency is illustrated in Figure~\ref{fig:SpringMass_LDMatrix_dCases} (left) in the form of a {\it dependency matrix}, where the horizontal axis is the index of each column of $\bm{\Phi}(\mathbf{x})$ and the vertical axis is the index of columns it depends on. White blocks show no dependency and the colored blocks show dependency with a magnitude obtained from the ID coefficient matrix $\mathbf{C}$ in (\ref{eq:rankrfactSC}). As can be observed, the first column of $\bm{\Phi}(\mathbf{x})$ depends on the $4$th and $6$th columns. Following (\ref{eq:eta_recovery}), the recovered constraint function for the case of $d = 2$ and $\sigma = 10^{-3}$ is
\begin{equation}\label{eq:constraintmanifold_sdof_approx}
    g(\mathbf{x}) \approx -1.0000 \phi_1(\mathbf{x}) + 0.9999 \phi_4(\mathbf{x}) + 0.1000 \phi_6(\mathbf{x}),\nonumber
\end{equation}
which closely matches (\ref{eq:constraintmanifold_sdof}), as also depicted in Figure \ref{fig:SpringMass_constraintplot}. Figure~\ref{fig:SpringMass_LDMatrix_dCases} (right) shows the dependency matrix for the case of $d=3$ at $\sigma = 10^{-3}$. Note that increasing the polynomial degree adds additional dependent columns in $\bm{\Phi}(\mathbf{x})$, which are trivial variants of (\ref{eq:constraintmanifold_sdof}). Specifically, one constraint is of the form $x = 0.9998x^3 + 0.1000xy^2$ or equivalently $x(-1+ 0.9998\phi_4 + 0.1000\phi_6)\approx xg(\mathbf{x})\approx 0$ and the other one $y = 1.0001x^2y + 0.1000y^3$ or equivalently 
$y(-1 + 1.0001\phi_4 + 0.1000\phi_6)\approx yg(\mathbf{x})\approx 0$.


\begin{figure}[H]
    \centering
    \includegraphics[trim = 0 0 0 0, clip,width=0.49\textwidth]{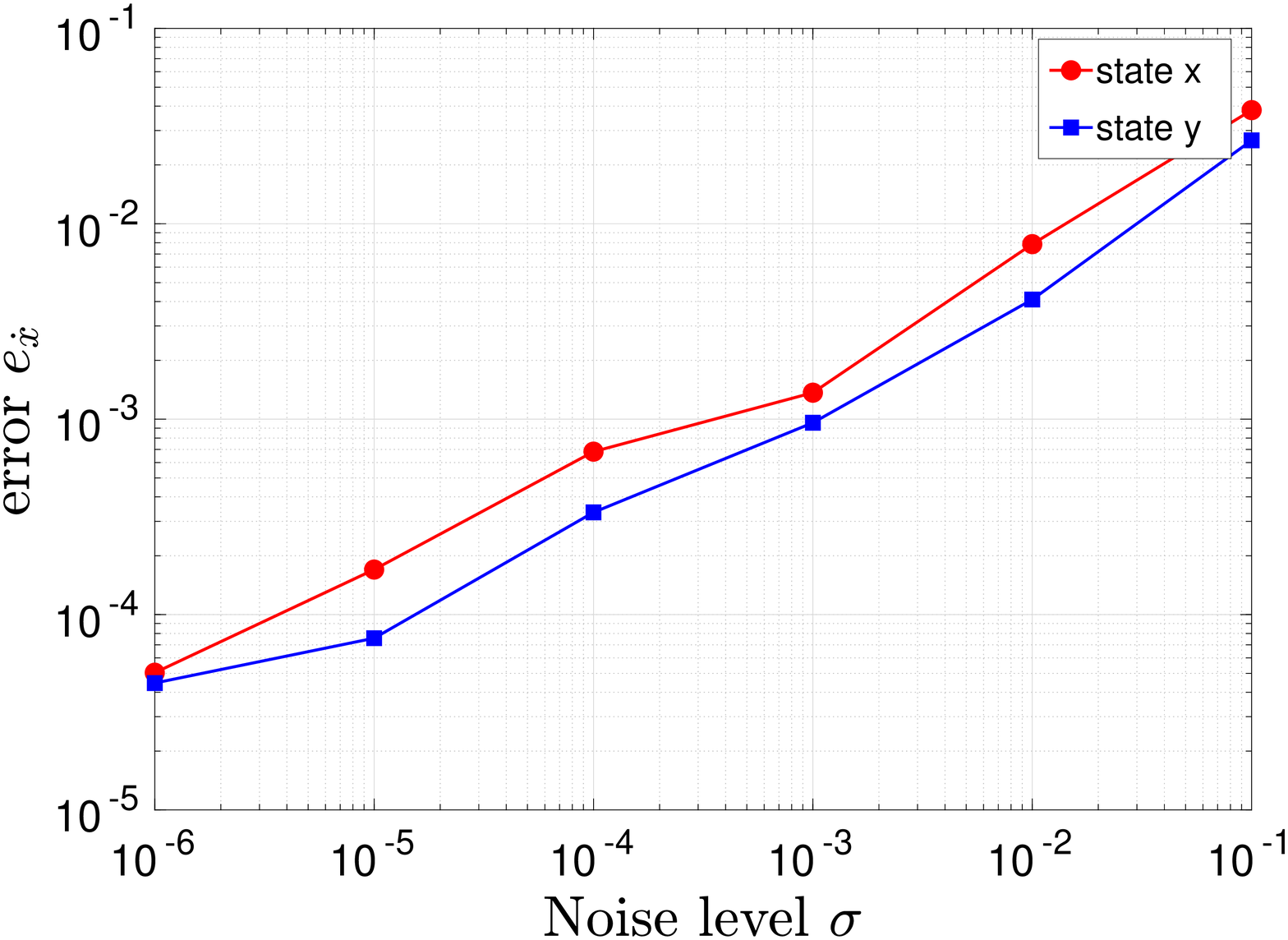}
    \includegraphics[trim = 0 3 0 1, clip,width=0.49\textwidth]{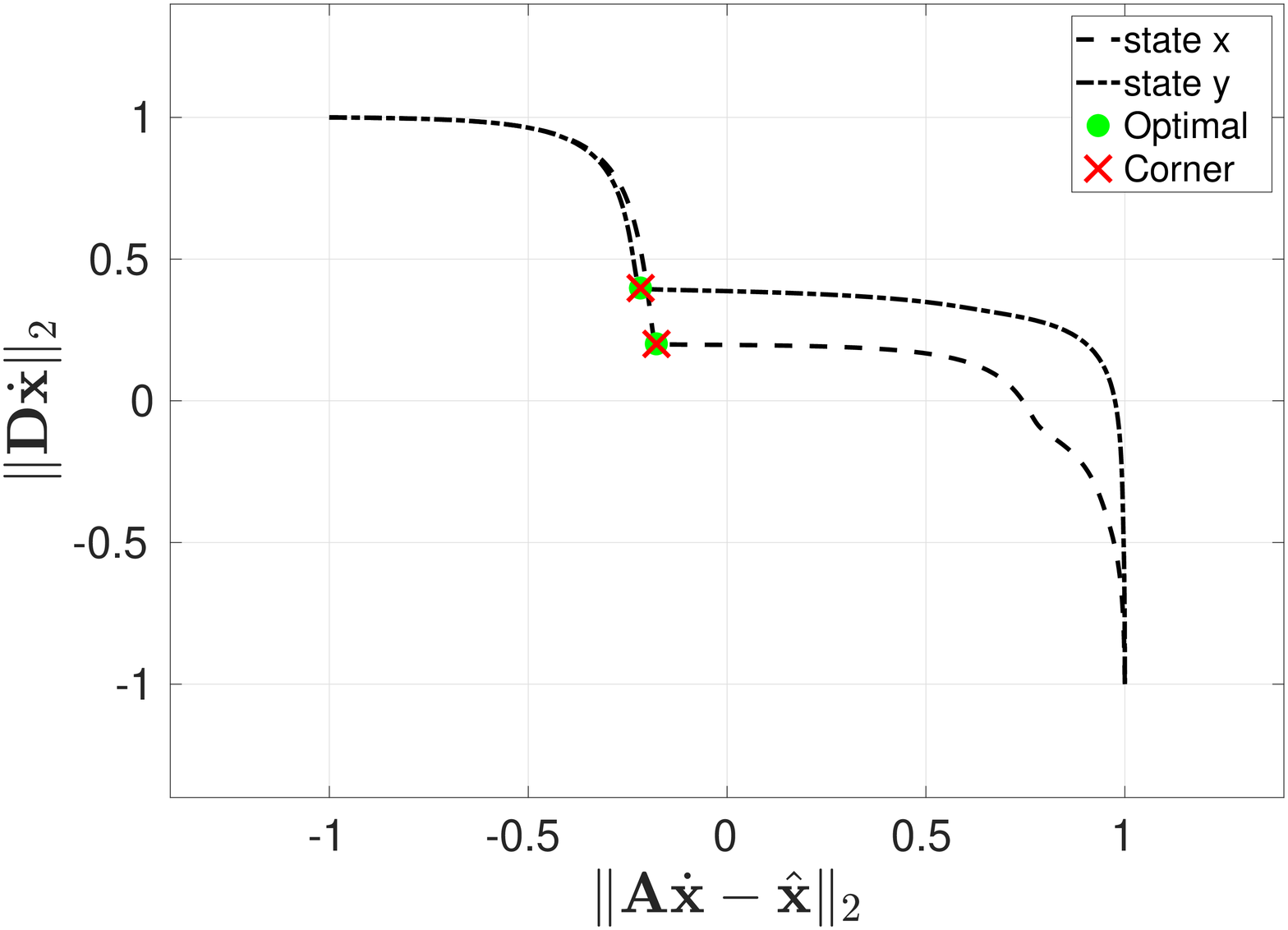}
    \caption{Left: Relative error of the spring-mass state derivatives with respect to different noise levels. Right: Pareto curves for each state of the spring-mass system at noise level $\sigma = 10^{-3}$. The signal-to-noise ratios are: $\text{SNR}_x = 57.30~\text{dB}$ and $\text{SNR}_y = 66.67~\text{dB}$.}
    \label{fig:SpringMassTikDiff}
\end{figure}

\begin{figure}[H]
    \centering
    \includegraphics[trim = 0 0 0 0, clip,width=.49\textwidth]{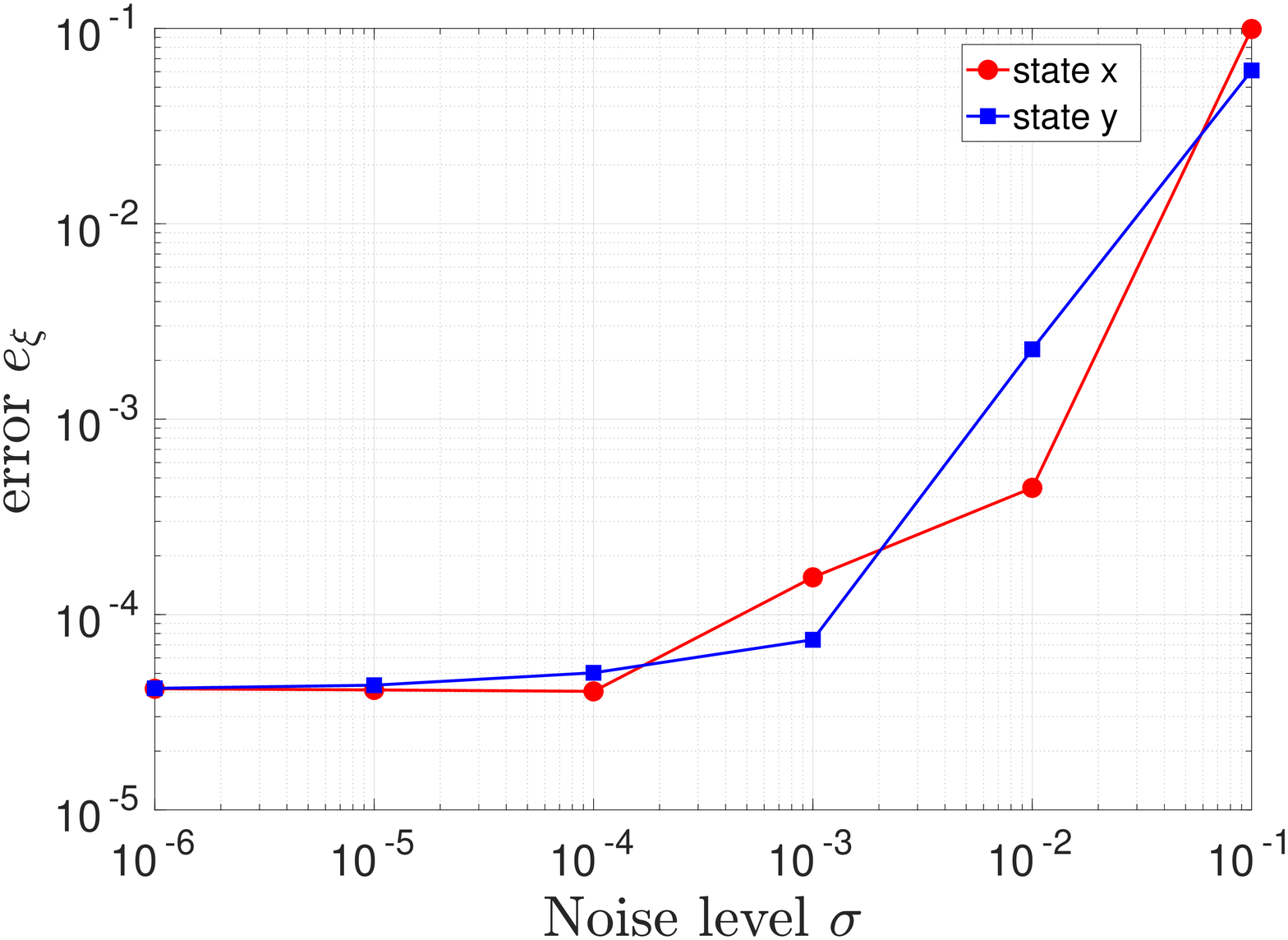}
    \includegraphics[trim = 0 0 0 0, clip,width=.49\textwidth]{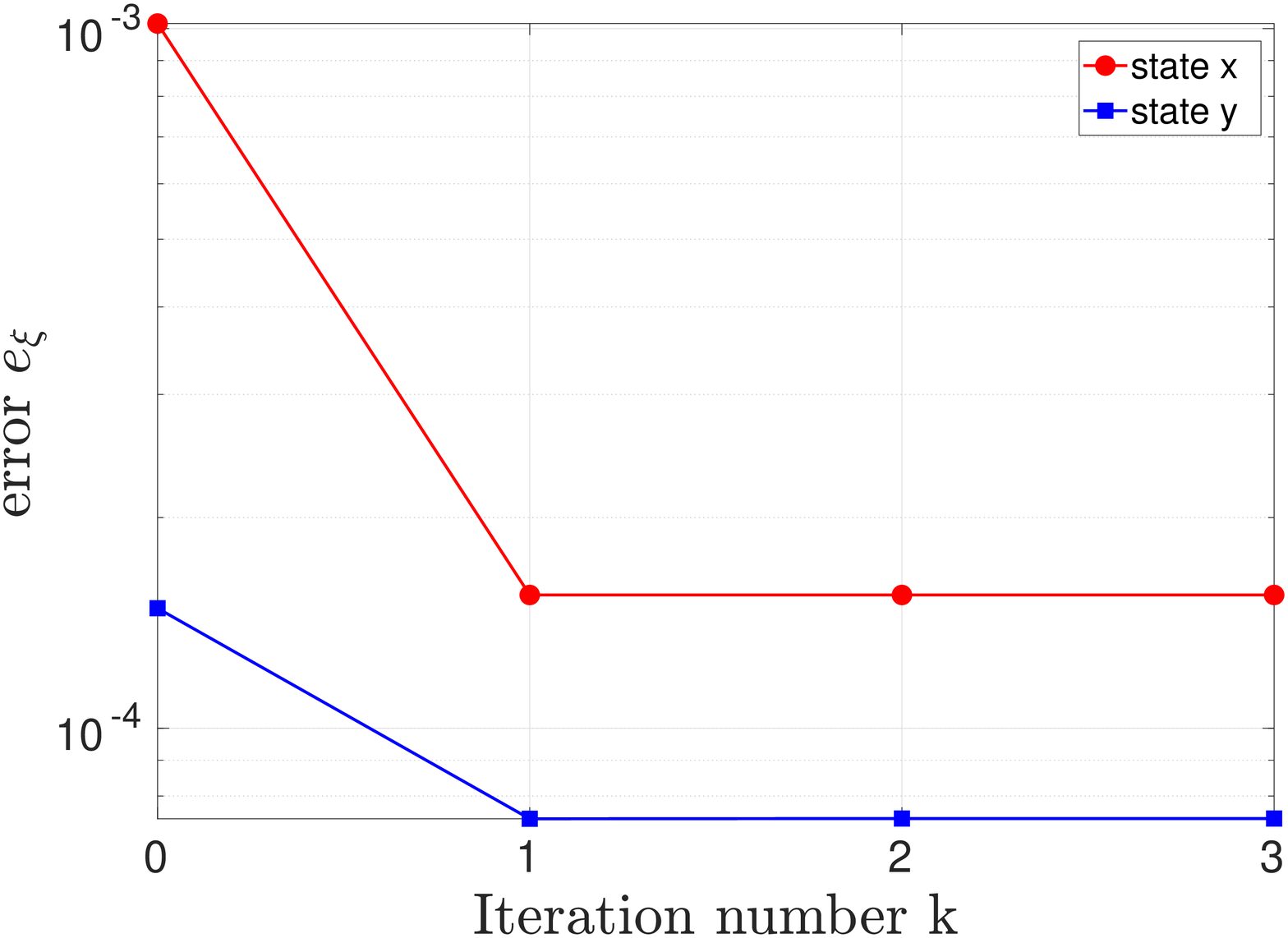}
    \caption{Left: Relative solution error of the spring-mass system with respect to different noise levels. Right: Relative solution error of the spring-mass system with respect to iteration $k$ at noise level $\sigma = 10^{-3}$.}
    \label{fig:SpringMasserrorcoeff_and_convergence}
\end{figure}

We proceed with computing $\bm\xi$ using degree $d=2$ basis by first removing one of the dependent columns of $\bm{\Phi}(\mathbf{x})$ to improve its conditioning. We may chose to eliminate either the $1$st, $4$th, or the $6$th column of $\bm{\Phi}(\mathbf{x})$. With the aim of arriving at a {\it simpler}, i.e., lower order, model of the system, we remove the column associated with the highest polynomial degree, i.e., the $6$th column in the case of $d=2$. By doing so, the condition number of $\bm{\Phi}(\mathbf{x})$ is lowered from $1.46\cdot 10^3$ to $3.16$ for $\sigma = 10^{-3}$.
The error of the state derivatives as well as the Pareto curves for the spring-mass system are illustrated in Figure~\ref{fig:SpringMassTikDiff}. The same trend as in the previous examples is shown in Figure~\ref{fig:SpringMasserrorcoeff_and_convergence} for the relative solution error and the convergence of WBPDN for $\sigma = 10^{-3}$.

\begin{figure}[H]
    \centering
    \includegraphics[trim = 0 0 0 0, clip,width=0.6\textwidth]{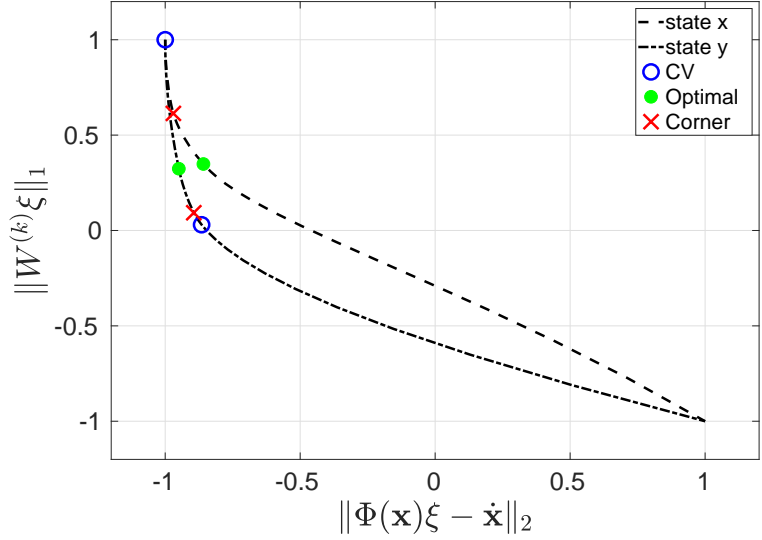}
    \caption{Pareto curves for each state of the spring-mass system at noise level $\sigma = 10^{-3}$ and iteration 0.}
    \label{fig:ParetoCurveWBPDNLcurve_it0}
\end{figure}

\begin{figure}[H]
    \centering
    \includegraphics[trim = 0 0 0 0, clip,width=0.8\textwidth]{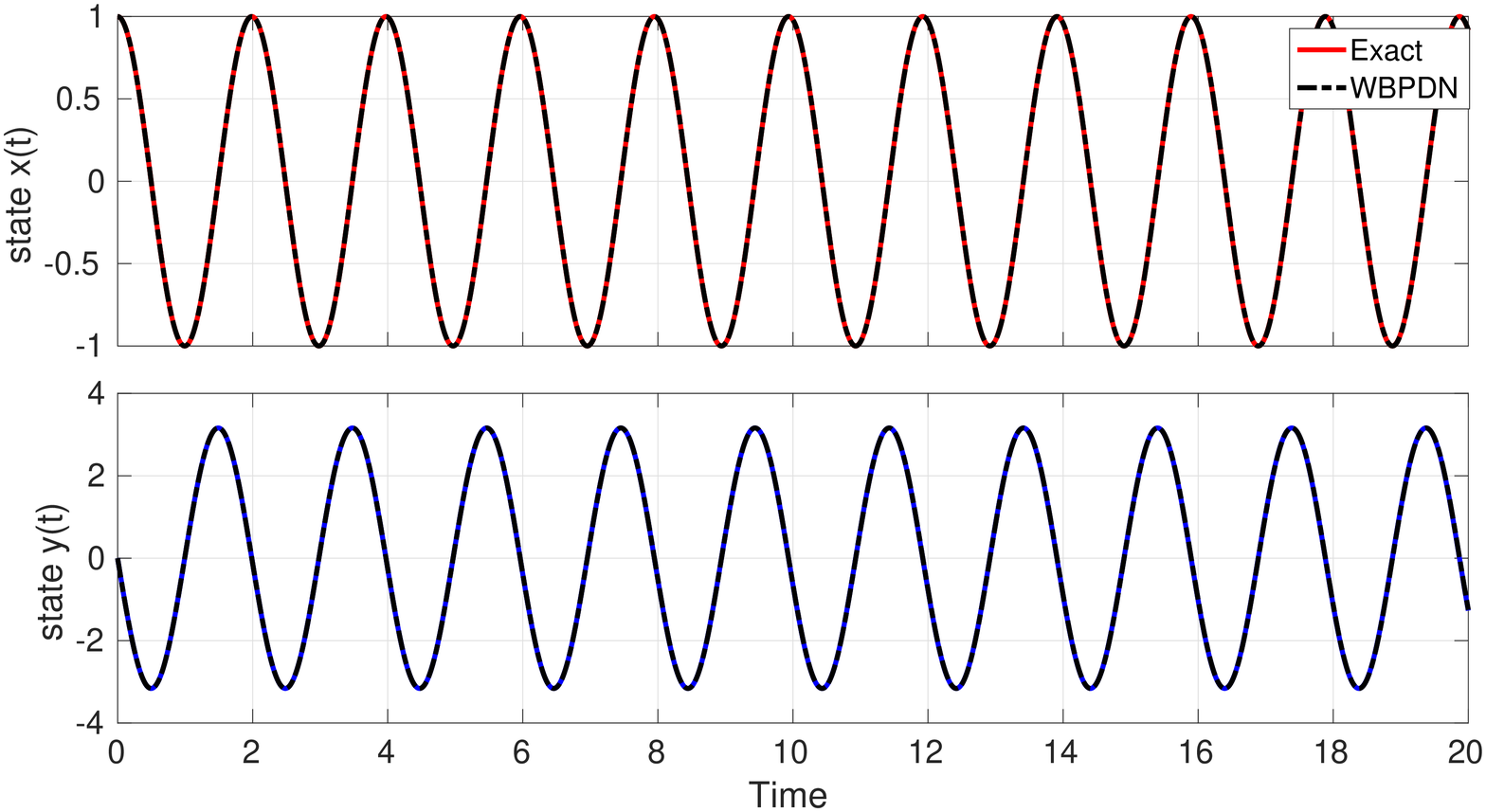}
    \caption{Exact and predicted trajectories of the spring-mass system for $\sigma  = 10^{-3}$.}
    \label{fig:SpringMassWBPDNPredicted}
\end{figure}

The Pareto curves of Figure~\ref{fig:ParetoCurveWBPDNLcurve_it0} are smooth, and the location of the corner point is not as evident as in other examples. In fact, the corner points do not coincide with the optimal ones. This may be caused by the well-conditioning of the truncated $\bm{\Phi}(\mathbf{x})$. Because of this, the solution error does not depend on the regularization parameter as strongly. The exact and predicted state trajectories shown in~Figure \ref{fig:SpringMassWBPDNPredicted} agree well.

\subsection{Euler rigid body dynamics}
\label{subsec:EulerRBD}

\begin{figure}
  \centering
  \includegraphics[trim={300 180 400 110},clip = true, width = 0.7\linewidth]{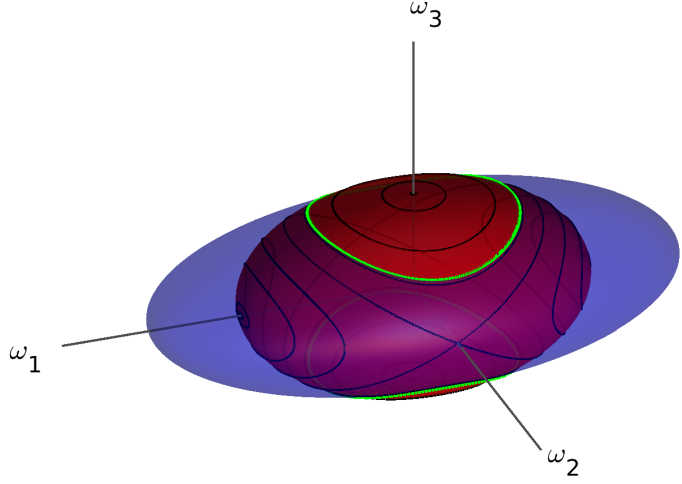}
  \caption{The intersection of energy (red) and angular momentum (blue) constraint ellipsoids define the polhode curve (green), which corresponds to the state trajectory.}
  \label{fig:polhode}
\end{figure}

In classical mechanics, Euler's rotation equations describe the rotation of a rigid body in a rotating reference frame with axes fixed to the body and parallel to the body's principal axes of inertia. These equations are widely known in the spacecraft dynamics community, where attitude performance is essential to meet pointing requirements. For instance, Earth observation satellites must achieve extreme accuracy to point antennas, optical instruments or other remote sensing devices toward specific targets. By improving spacecraft models and estimating physical parameters more accurately, one can enhance the overall performance of the mission. The equations in component form and derived with respect to the principal directions are given by
\begin{subequations}\label{eq:EulerRBD}
\begin{alignat}{3}
    I_1 \dot{\omega}_1 + (I_3 - I_2) \omega_2\omega_3 = \tau_1, &\quad \omega_1(0) = \omega_{1,0},\label{eq:EulerRBD_a}\\
    I_2 \dot{\omega}_2 + (I_1 - I_3) \omega_3\omega_1 = \tau_2, &\quad \omega_2(0) = \omega_{2,0},\label{eq:EulerRBD_b}\\
    I_3 \dot{\omega}_3 + (I_2 - I_1) \omega_1\omega_2 = \tau_3, &\quad \omega_3(0) = \omega_{3,0}.\label{eq:EulerRBD_c}
\end{alignat}
\end{subequations}
where $I_k, \omega_k, \tau_k$, $k=1,2,3$, are the principal moments of inertia, the angular velocities, and the applied external torques, respectively. In this example, we assume no external torques excite the system, i.e., $\tau_1=\tau_2=\tau_3=0$. In the torque-free case, the Euler system (\ref{eq:EulerRBD}) is conservative and satisfies two integrals of motion: conservation of kinetic energy (\ref{eq:cons_kinetic}) and conservation of angular momentum (\ref{eq:cons_angmom})
\begin{subequations}\label{eq:EulerRBD_conservation}
\begin{alignat}{2}
     2\mathcal{T} = I_1 \omega_1^2 + I_2 \omega_2^2 + I_3 \omega_3^2,\label{eq:cons_kinetic}\\
     \Vert\bm{L}\Vert^2_2 = I_1^2 \omega_1^2 + I_2^2 \omega_2^2 + I_3^2 \omega_3^2,\label{eq:cons_angmom}
\end{alignat}
\end{subequations}
where $\mathcal{T}$ and $\bm{L}$ are the kinetic energy and angular momentum vector, respectively. The above constraints represent two ellipsoid-shaped manifolds in the $(\omega_1,\omega_2,\omega_3)$-space whose intersection curve is known as the polhode (Figure~\ref{fig:polhode}). As the true state variables satisfy these two constraints, they lie on their intersection. The purpose of this example is to show that, while the proposed ID approach (Algorithm \ref{alg:ID}) is not able to learn these two constraints individually, it is able to learn their intersection for small noise levels.

\begin{figure}[H]
    \centering
    \includegraphics[trim = 0 0 0 0, clip,width=0.4\textwidth]{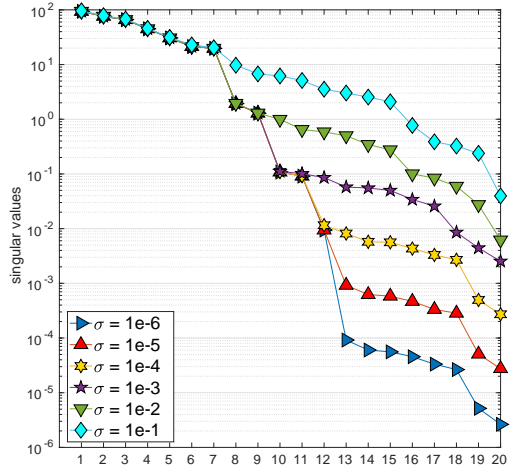}
    \caption{Singular value decay of $\bm{\Phi}(\mathbf{x})$ at different noise levels for Euler Rigid Body Dynamics. In this example there is no clear gap on the singular values of $\bm{\Phi}(\mathbf{x})$. This may lead to the misconception that $\bm{\Phi}(\mathbf{x})$ is full-rank, where, in fact, there exist constraints making some of the columns approximately linearly dependent.}
    \label{fig:svDecay_EulerRBD}
\end{figure}

\begin{figure}[H]
    \centering
    \includegraphics[trim = 320 0 320 0, clip,width=0.48\textwidth]{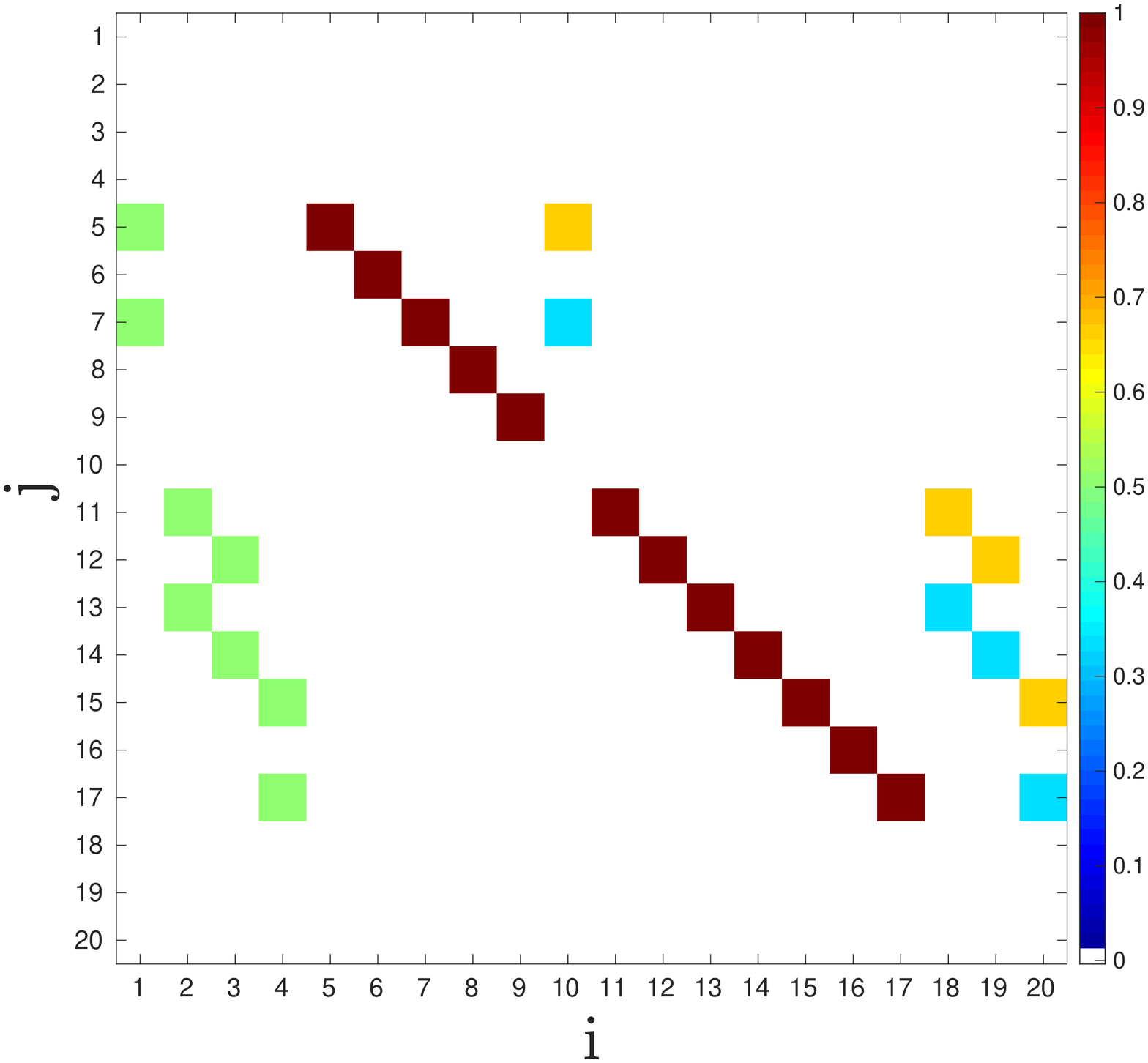}
    \includegraphics[trim = 320 0 320 0, clip,width=0.48\textwidth]{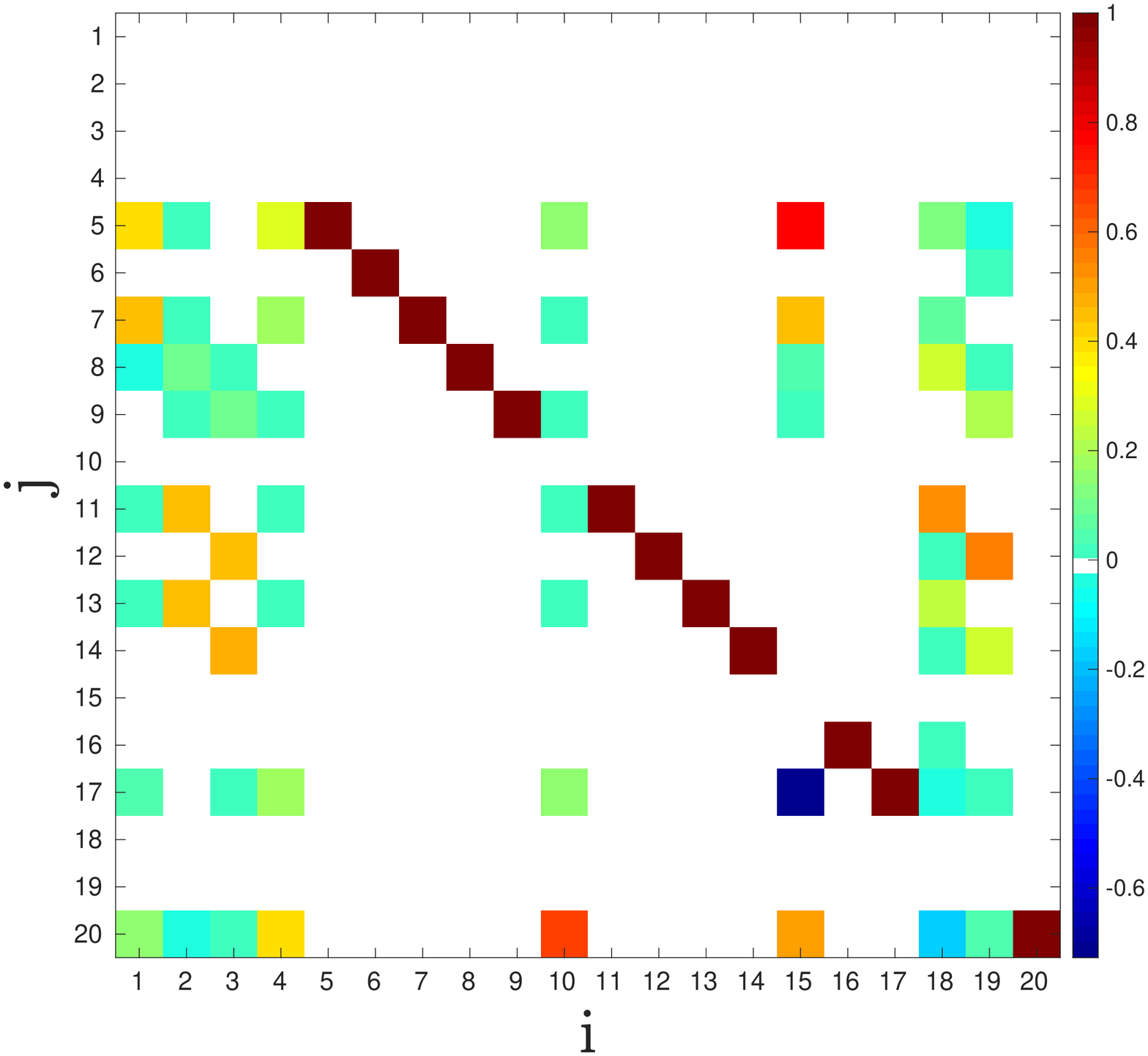}
    \caption{Linear dependence matrix for Euler Rigid Body Dynamics at noise level $\sigma = 10^{-5}$ (Left) and $\sigma = 10^{-3}$ (Right). The horizontal axis is the index $i$ of each column of $\bm{\Phi}(\mathbf{x})$ and the vertical axis is the index $j$ of columns it depends on. White blocks show no dependency and the colored blocks show dependency with a magnitude obtained from the ID coefficient matrix $\mathbf{C}$ in (\ref{eq:rankrfactSC}).} 
    \label{fig:LDMatrix_EulerRBD}
\end{figure}

\begin{figure}[H]
    \centering
    \includegraphics[trim = 300 120 300 50, clip,width=0.65\textwidth]{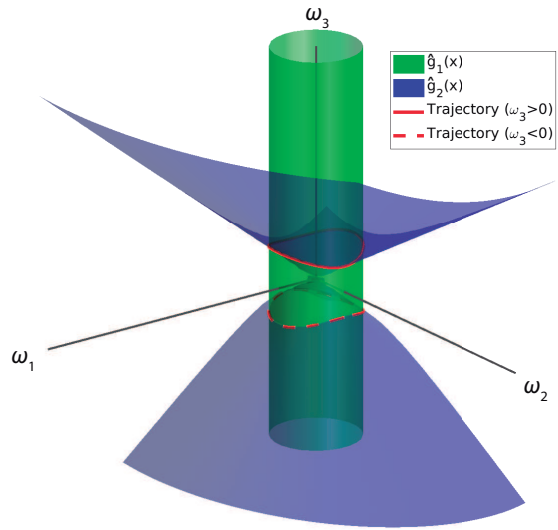}
    \caption{Identified constraints $\hat{g}_1(\mathbf{x})$ and $\hat{g}_2(\mathbf{x})$ for the Euler Rigid Body Dynamics problem with $d = 3$ at noise level $\sigma = 10^{-5}$. The intersection of $\hat{g}_1(\mathbf{x})$ and $\hat{g}_2(\mathbf{x})$ matches well with the exact trajectory of the system shown in solid and dashed red curves.}
    \label{fig:EulerRBD_constraintplot}
\end{figure}

In this example, the moments of inertia are set to $I_1 = 1 $, $I_2 = 2$ and $I_3 = 3$, yielding a tri-inertial body, and the initial conditions to $(\omega_1, \omega_2, \omega_3) = (1,1,1)$. These specific values give $\mathcal{T}=3$ and  $\Vert\bm{L}\Vert^2_2 = 14$. The number of state variables is $n = 3$ and the degree of the polynomial basis is set to $d = 3$, giving $p = 20$ monomial terms. The sparsity of this system in a polynomial basis is 3. We used 1000 samples over 10 time units, from $t = 0.5$ to $t = 10.5$.
The ellipsoid constraint functions associated with (\ref{eq:EulerRBD_conservation}) are expressed as a linear combination of multivariate monomials as
\begin{subequations}\label{eq:EulerRBD_ellipconstraints}
\begin{alignat}{2}
    g_1(\mathbf{x}) = 1\phi_1(\mathbf{x}) - \frac{1}{6}\phi_5(\mathbf{x}) - \frac{1}{3}\phi_7(\mathbf{x}) - \frac{1}{2}\phi_{10}(\mathbf{x});\\
    g_2(\mathbf{x}) = 1\phi_1(\mathbf{x}) - \frac{1}{14}\phi_5(\mathbf{x}) - \frac{2}{7}\phi_7(\mathbf{x}) - \frac{9}{14}\phi_{10}(\mathbf{x}),
\end{alignat}
\end{subequations}
where $\phi_i(\mathbf{x})\in \{1,\omega_1,\omega_2,\omega_3,\omega_1^2,\omega_1\omega_2,\omega_2^2,\omega_1\omega_3,\omega_2\omega_3,\omega_3^2,\dots\}, i=1,\dots,p$. Similar to the previous example, we compute the singular values of $\bm{\Phi}(\mathbf{x})$ for polynomial degree $d = 3$ at different noise levels $\sigma$ to find the numerical rank. As shown in Figure~\ref{fig:svDecay_EulerRBD}, except for the case of $\sigma=10^{-6}$, there is no clear gap in the singular values. This presents an issue when learning the constraint via Algorithm \ref{alg:ID} and  subsequently computing $\bm{\xi}$.
We performed Algorithm \ref{alg:ID} knowing the true rank {\it a priori}, which is 12 in this example. Figure~\ref{fig:LDMatrix_EulerRBD} illustrates the linear dependence matrix for $d = 3$ at two different noise levels $\sigma = 10^{-5}$ (left) and $\sigma = 10^{-3}$ (right). The ID leads to $\mathcal{J} = \{5,6,7,8,9,11,12,13,14,15,16,17\}$ as the indices of the independent columns of $\bm{\Phi}(\mathbf{x})$ for $\sigma = 10^{-5}$, which is the nose level we use for the results presented next. Larger noise levels, lead to several small $\eta_k$ coefficients, which motivate future research in developing better denoising strategies for constraint identification. We select the 1st and 10th columns of $\bm{\Phi}(\mathbf{x})$ corresponding to the subset $\mathcal{I}\backslash\mathcal{J} = \{1,2,3,4,10,18,19,20\}$ of linearly dependent columns to generate the following recovered constraint functions
\begin{subequations}\label{eq:constraintmanifold_mdof_approx}
\begin{alignat}{2}
    \hat{g}_1(\mathbf{x}) = -1.0000\phi_1(\mathbf{x}) + 0.5002\phi_5(\mathbf{x}) + 0.5001\phi_7(\mathbf{x});\nonumber\\
    \hat{g}_2(\mathbf{x}) = 0.6625\phi_5(\mathbf{x}) + 0.3308\phi_7(\mathbf{x})-1.0000\phi_{10}(\mathbf{x}),\nonumber
\end{alignat}
\end{subequations}
also depicted in Figure \ref{fig:EulerRBD_constraintplot}. In obtaining (\ref{eq:constraintmanifold_mdof_approx}), we have thresholded the $C_{m,l}$ coefficients in (\ref{eq:eta_recovery}) with a threshold parameter of size $10^{-3}$. Notice that the recovered constraints $\hat{g}_1(\mathbf{x})$ and $\hat{g}_2(\mathbf{x})$ are different from the exact constraints $g_1(\mathbf{x})$ and $g_2(\mathbf{x})$, respectively, hence the $\ \hat{}\ $ notation for the recovered constraints. This can be observed also by comparing Figures \ref{fig:polhode} and \ref{fig:EulerRBD_constraintplot}. However, the intersection of $\hat{g}_1(\mathbf{x})$ and $\hat{g}_2(\mathbf{x})$ matches well with that of $g_1(\mathbf{x})$ and $g_2(\mathbf{x})$, which is again what constraints the state variables. We note that selecting other columns with indices in $\mathcal{I}\backslash\mathcal{J}$ results in constraints that are trivial variants of $\hat{g}_1(\mathbf{x})$ and $\hat{g}_2(\mathbf{x})$.

This example shows the difficulty in uniquely identifying more than one constraints given by implicit functions of the state variables and using the ID approach. Given a basis $\bm{\Phi}(\mathbf{x})$, there exist a family of different hypersurfaces that contain the state trajectory, defined by their intersection. This is caused by the non-uniqueness of the solution to $\bm{\Phi}(\mathbf{x})\bm{\eta} = \mathbf{0}$; any linear combination of the null space vectors of $\bm{\Phi}(\mathbf{x})$ will also contain the state trajectory. Recovering constraints in the data poses new challenges in system identification. A thorough analysis of the conditions under which constraints can be uniquely identified is still being investigated and will be addressed in a future work.

\begin{figure}[H]
    \centering
    \includegraphics[width=.49\textwidth]{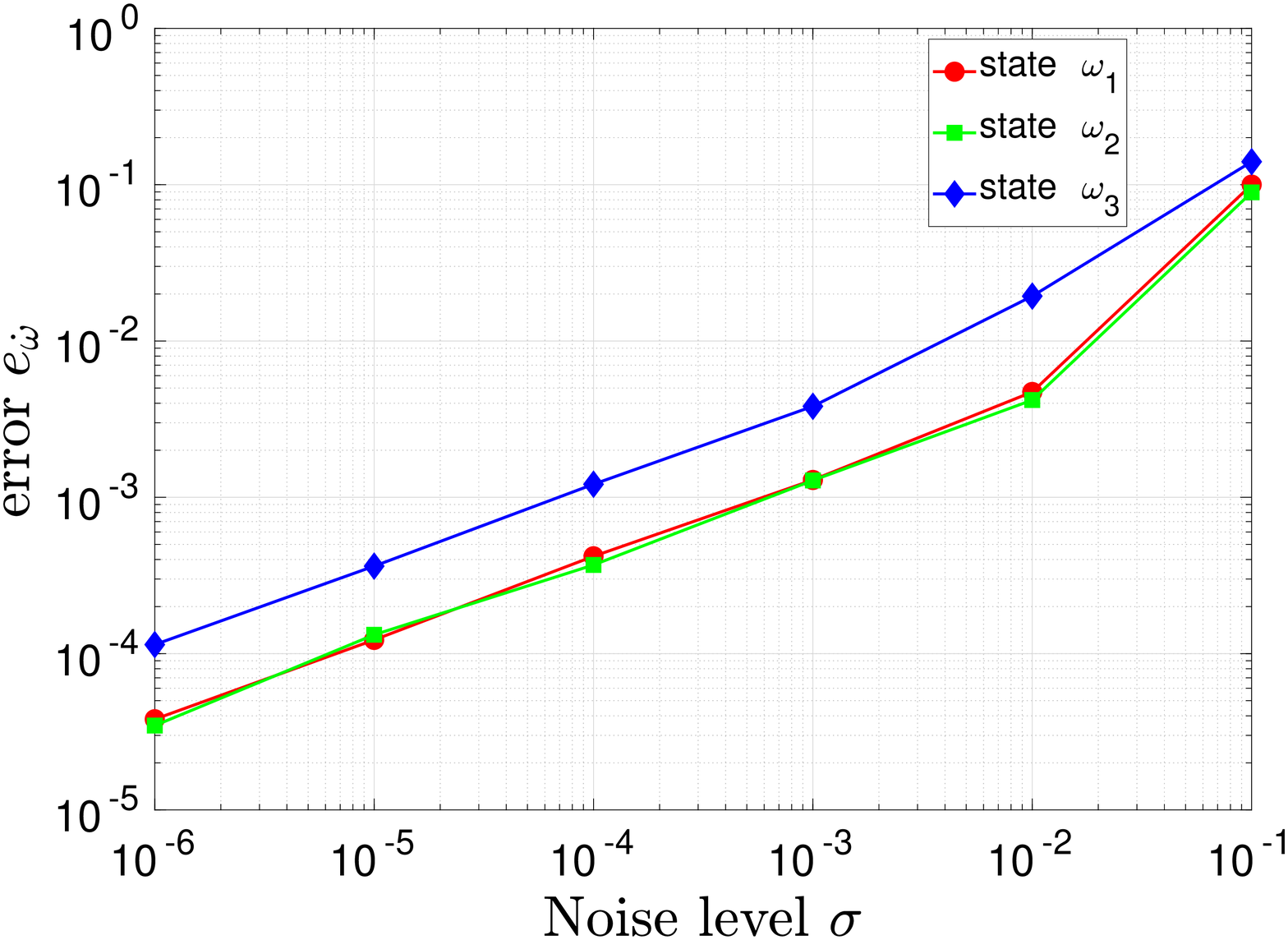}
    \includegraphics[trim = 0 0 0 0, clip,width=.49\textwidth]{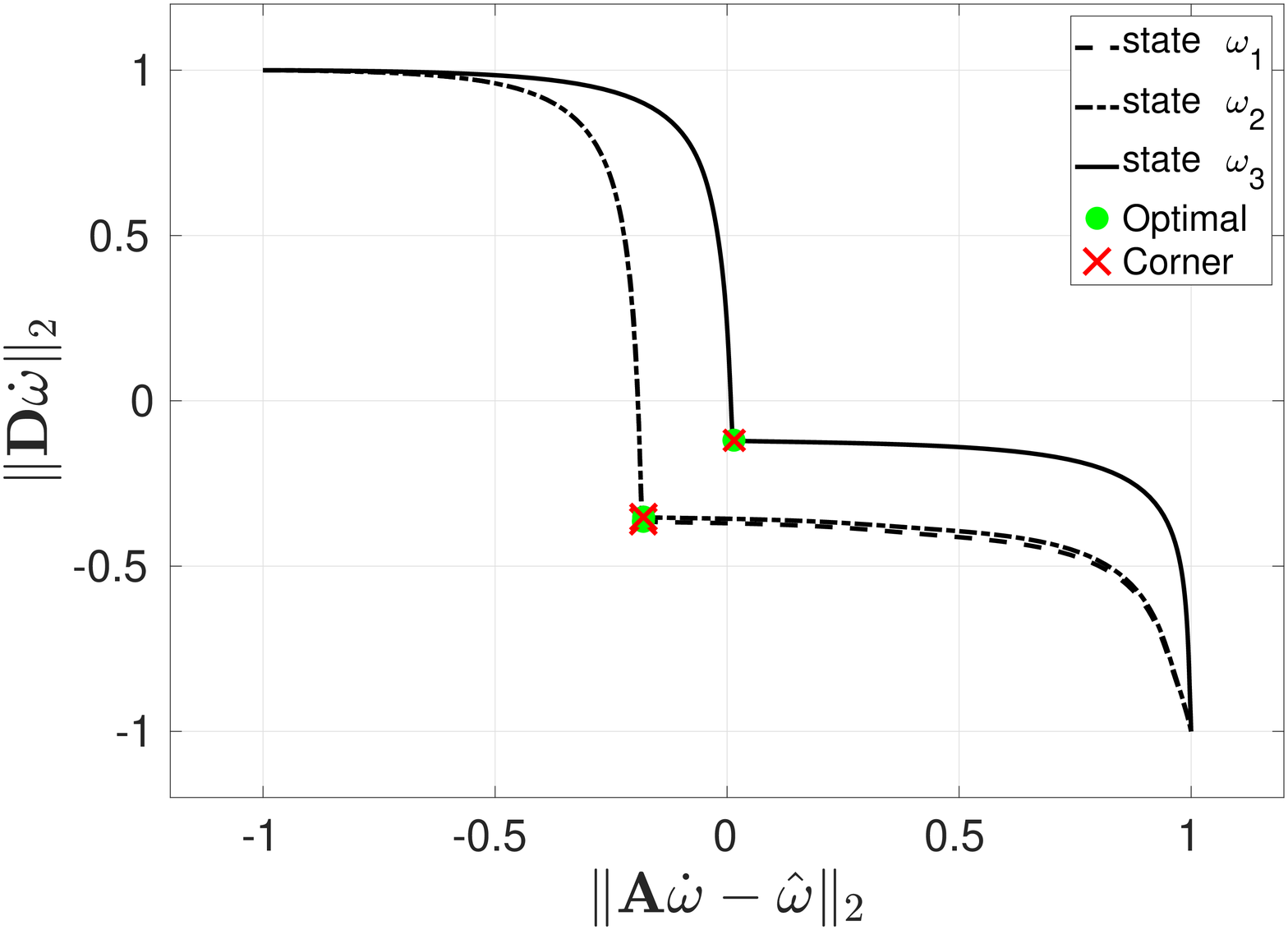}
    \caption{Left: Relative error of the angular accelerations with respect to different noise levels. The signal-to-noise ratios are: $\text{SNR}_{\omega_1} = 58.18~\text{dB}$ and $\text{SNR}_{\omega_2} = 60.68~\text{dB}$, $\text{SNR}_{\omega_3} = 59.75~\text{dB}$. Right: Pareto curves for each angular velocity at noise level $\sigma = 10^{-3}$ }
    \label{fig:EulerRBDDXerror}
\end{figure}

In this case, Tikhonov regularization differentiation also performs well in computing derivatives from noisy measurements for all state variables. The Pareto curves are well defined, and the corners coincide with the optimal regularization parameters, as shown in Figure~\ref{fig:EulerRBDDXerror}.

\begin{figure}[H]
    \centering
    \includegraphics[trim = 230 0 250 0, clip = true,width=.49\textwidth]{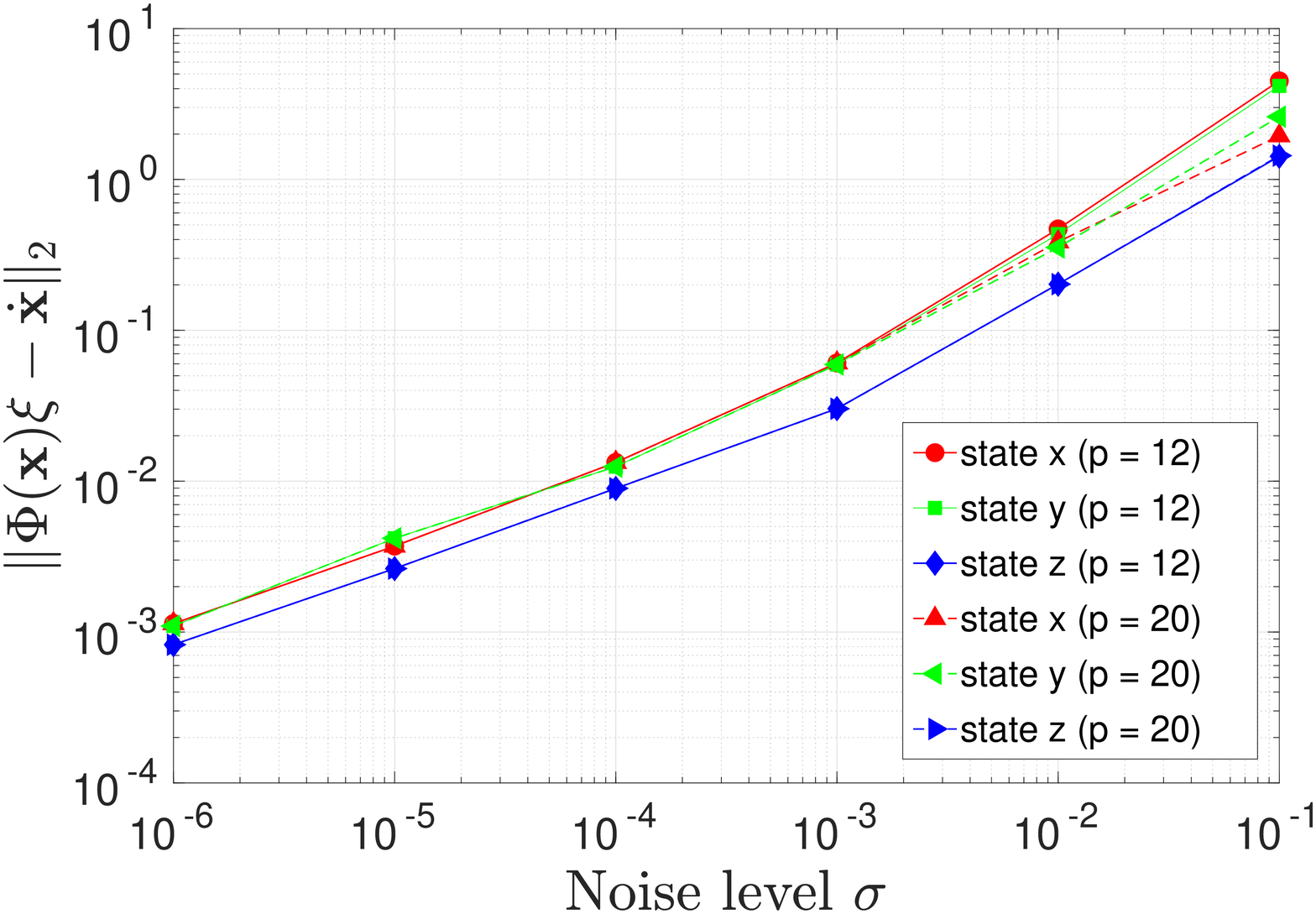}
    \includegraphics[trim = 240 0 250 0, clip = true,width=.49\textwidth]{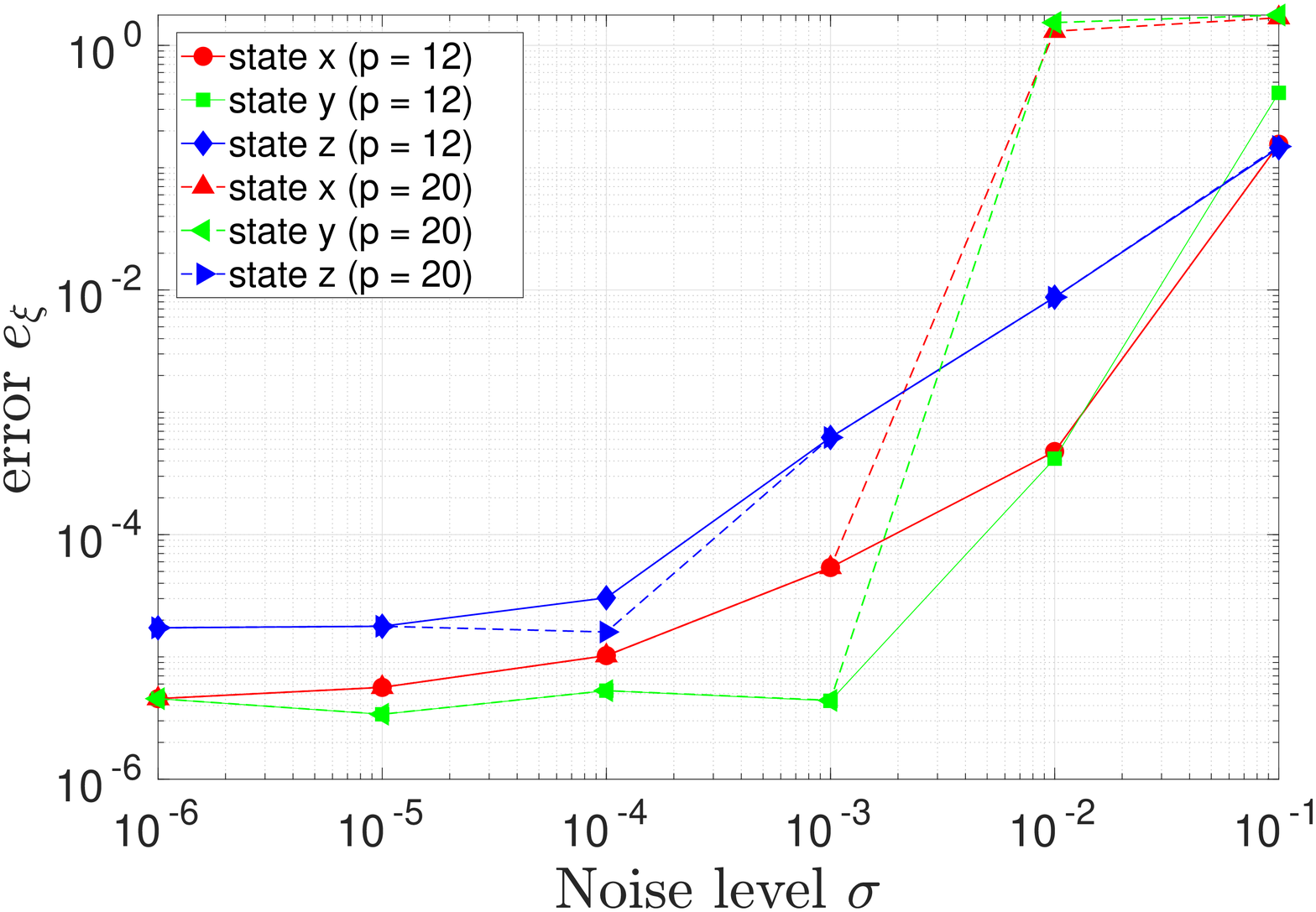}
    \caption{Left: Residual error at different noise levels for $p = 12$ and $p = 20$. Right: Relative solution error at different noise levels for $p = 12$ and $p = 20$. $p = 12$ represents the number of columns in $\bm{\Phi}(\mathbf{x})$ after removing the linear dependent ones, whereas $p = 20$ corresponds to $\bm{\Phi}(\mathbf{x})$ keeping all the columns.}
    \label{fig:EulerRBD_res_error_rankdef}
\end{figure}

Before performing WBPDN, we remove the linearly dependent columns of $\bm{\Phi}({\mathbf{x}})$ in the subset $\mathcal{I}\backslash\mathcal{J}$ to lower the condition number of $\bm{\Phi}(\mathbf{x})$ from $3.40\cdot 10^4$ to $1.96\cdot 10^3$ for $\sigma = 10^{-3}$. The solution error and the convergence agree with the previous examples. We assume that the rank and the dimension of the null space of $\bm{\Phi}({\mathbf{x}})$ are known. If we run the WBPDN on the numerically rank deficient matrix, we may obtain a different $\bm{\xi}$, yielding similar residuals and predicted trajectories caused by the non-uniqueness of the solution due to the ill-conditioning of $\bm{\Phi}(\mathbf{x})$. Figure~ \ref{fig:EulerRBD_res_error_rankdef} displays results for the reduced $\bm{\Phi}(\mathbf{x})$ with  $p = 12$ columns and the original one. For low noise levels both solutions happen to coincide, but for $\sigma = 10^{-2}$ and $\sigma = 10^{-1}$ the solution diverges and the residual remains similar.

\begin{figure}[H]
    \centering
    \includegraphics[trim = 0 0 0 0, clip,width=.49\textwidth]{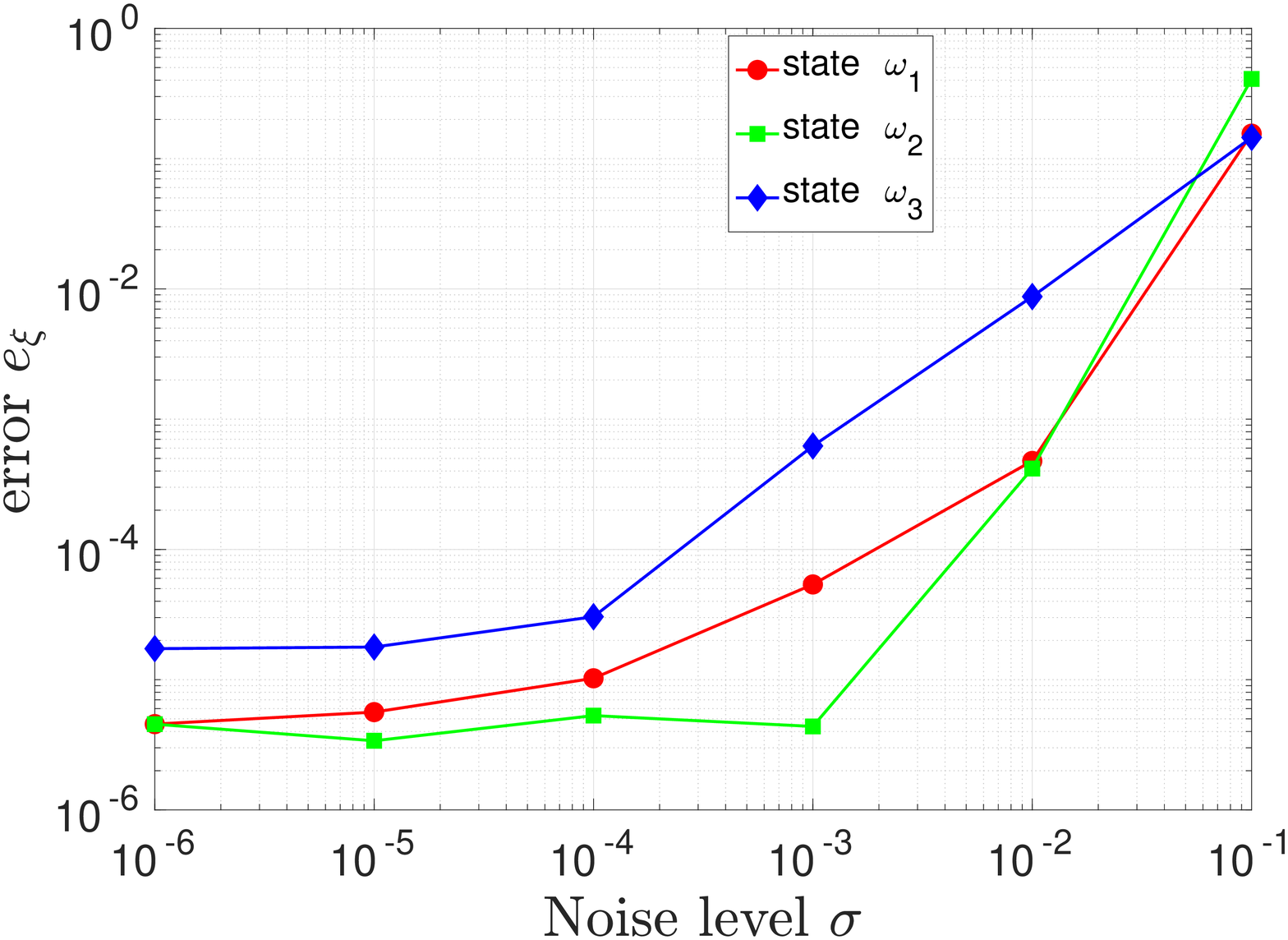}
    \includegraphics[trim = 0 0 0 0, clip,width=.49\textwidth]{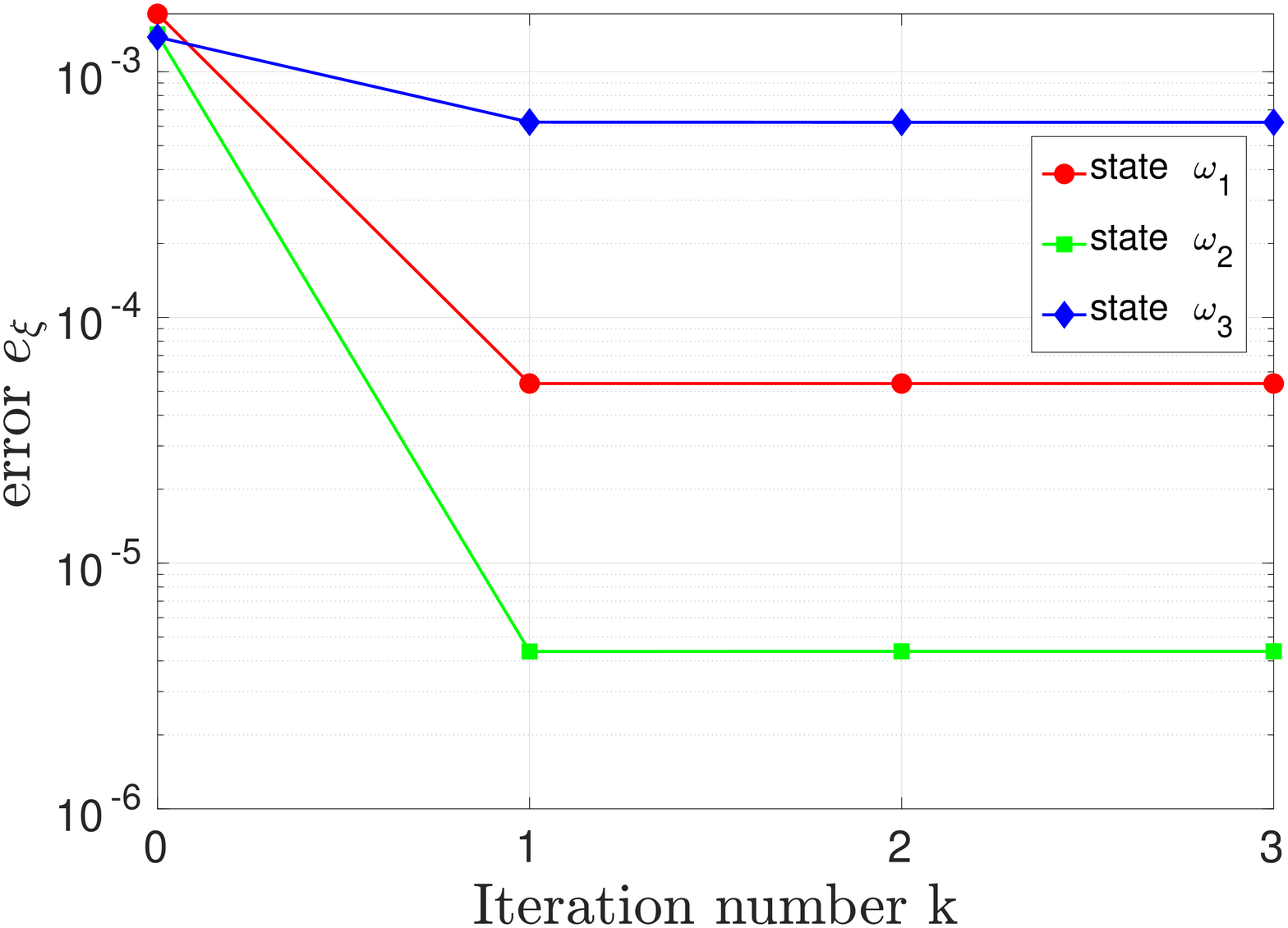}
    \caption{Left: Relative solution error of the Euler rigid body equations with respect to different noise levels. Right: Relative solution error with respect to iteration $k$ at noise level $\sigma = 10^{-3}$.}
    \label{fig:EulerRBDerrorcoeff_and_convergence}
\end{figure}
\begin{figure}[H]
    \centering
    \includegraphics[trim = 0 0 0 0, clip,width=0.6\textwidth]{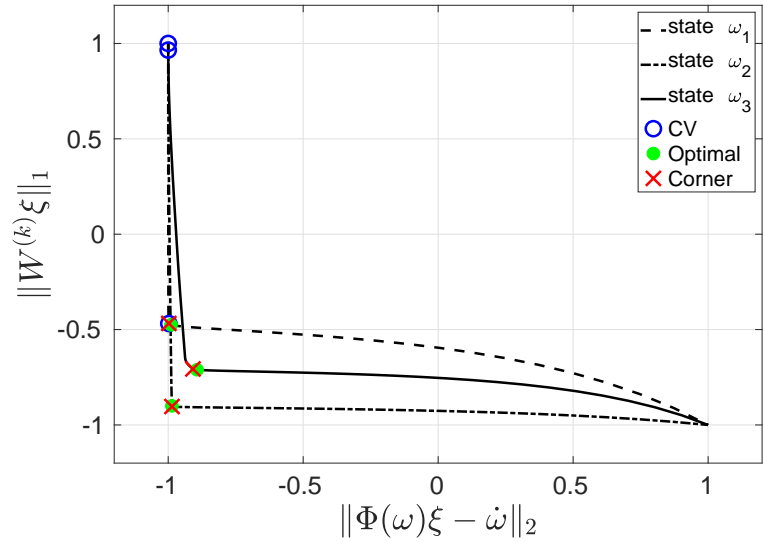}
    \caption{Pareto curves for each state at noise level $\sigma = 10^{-3}$ and iteration 0.}
    \label{fig:EulerRBDWBPDNLcurve_it0}
\end{figure}

\begin{figure}[H]
    \centering
    \includegraphics[width=0.8\textwidth]{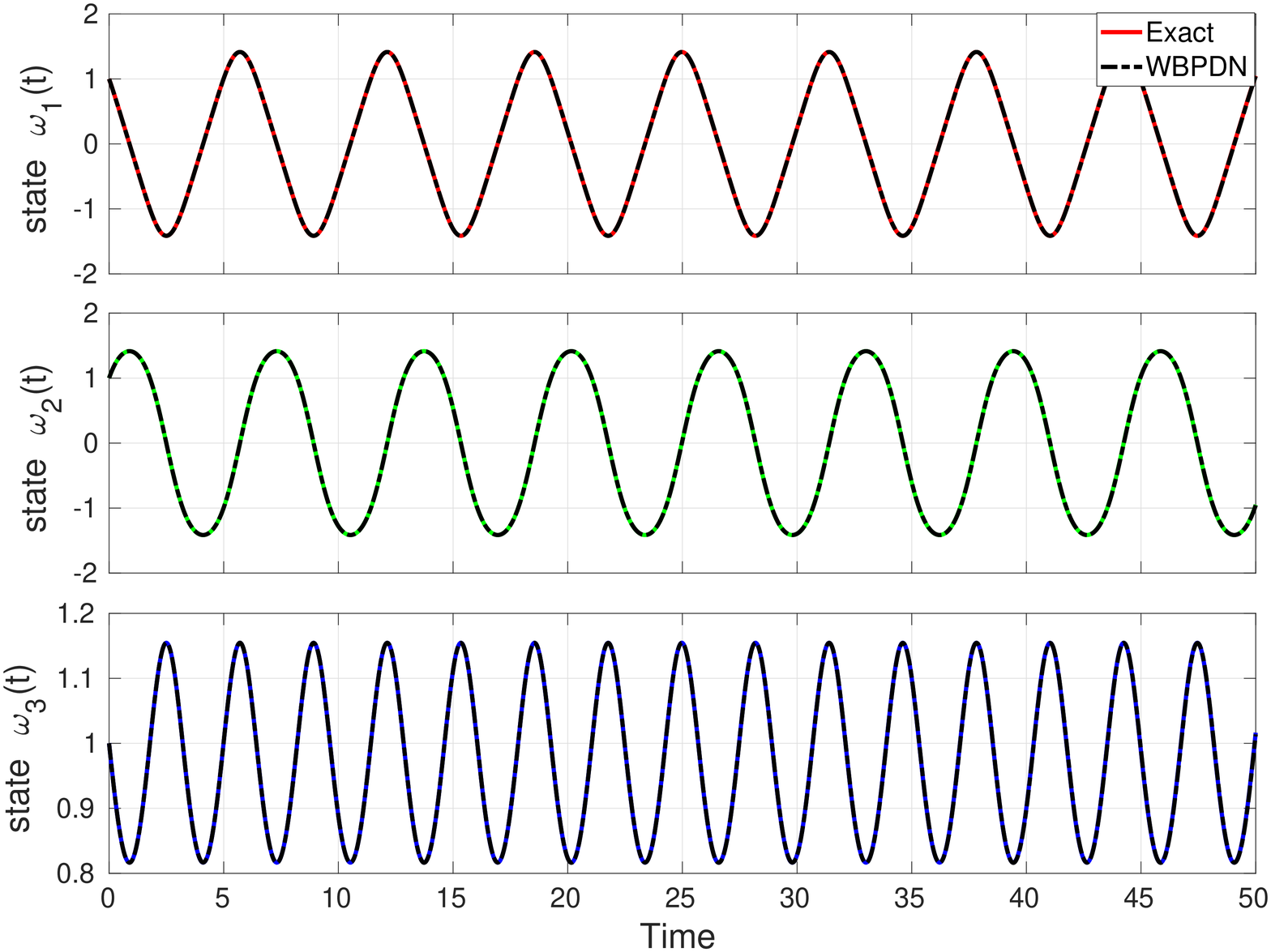}
    \caption{Exact and predicted trajectories of the Euler rigid body dynamics for $\sigma  = 10^{-3}$.}
    \label{fig:EulerRBDPrediction}
\end{figure}

From Figure~\ref{fig:EulerRBDWBPDNLcurve_it0} no significant differences are observed in the Pareto curves with respect to the previous examples. The curves are well-defined and the corner points match the optimal ones. The regularization parameters given by CV only coincide with the optimal ones for the $\omega_1$ state. Similarly, the exact and predicted trajectories agree until $t=50$ (Figure~\ref{fig:EulerRBDPrediction}). 

\section{Discussion and Conclusion}
\label{sec:conclusion}

In summary, the motivation of this work was to improve the accuracy and stability of sparse regression techniques, a.k.a SINDy, for the identification of governing equations of nonlinear dynamical systems from noisy state measurements. As in SINDy, the governing equations are approximated in a multi-variate basis of state variables, here of monomial type. To identify the expansion coefficients,  we proposed a strategy based on the iteratively reweighted Basis Pursuit Denoising (WBPDN) or Adaptive LASSO techniques, previously used in the context of compressed sensing and sparse regression. Penalizing the weighted $\ell_1$-norm of the coefficients via solution-dependent weights enhances sparsity and the stability of the regression problem. The selection of the regularization parameter balancing the sparsity of the recovered coefficients and the proximity of the model to data -- the time derivatives of the state variables -- was done using the Pareto curve criterion. We observed that the corner point of the Pareto curve yielded near optimal selection of regularization parameters, and that the Pareto curve criterion consistently outperforms the commonly used K-fold cross validation. For the range of noise levels we used, the Pareto curves showed well-defined corner points that agreed with regularization parameters that provided minimum solution error. We demonstrated empirically that the WBPDN approach outperformed Sequentially Thresholded Least Squares (STLS) and Sequential Thresholded RidgeRegression (STRidge) techniques previously utilized within the SINDy framework. 

We also addressed the recovery of a single constraint on dynamical systems, e.g., conservation of total energy, given by an implicit function of the state variables. This was achieved by generating the interpolative decomposition (ID) of the sampled basis matrix used to approximate the governing equations. We illustrated the utility of ID in recovering a single constraint for small levels of state measurement noise.

In light of the effectiveness of WBPDN for recovering governing equations of non-linear dynamical systems, it is desirable to extend this work to high-dimensional Partial Differential Equations (PDE) along with novel model reduction techniques to learn the dynamics of multi-physics or multi-scale systems. Robust identification of physical constraints from data -- specifically in the presence of large noise levels -- or the imposition of physical constraints to inform the identification of the governing equations are other directions to further investigate in future work.

\section{Acknowledgements}

This work was supported by the National Science Foundation grant CMMI-1454601.

\section*{Appendix A}
\label{sec:appendix_A}
The discrete derivatives are computed at the midpoints of $(t_i, t_{i+1})$ using the midpoint integration rule
\begin{equation}\label{eq:finitediff}
    x_{i+1} = x_{i} + \dot{x}_{i+1/2}\,\Delta t, \quad i = 1,\dots,m-1.
\end{equation}
The discretization (\ref{eq:finitediff}) yields the linear system
\begin{equation}\label{eq:midpointMat}
    \mathbf{A}\dot{\mathbf{x}} = \hat{\mathbf{x}},
\end{equation}
where $\mathbf{A} \in \R^{(m-1)\times(m-1)}$ is the discrete integral operator, $\dot{\mathbf{x}} \in \R^{m-1}$ are the approximate derivatives at the midpoints, and $\hat{\mathbf{x}} = \{x_j - x_1\}_{j = 2}^{m} \in \R^{m-1}$. Since noisy state variables may amplify the error in the derivatives, we regularize (\ref{eq:midpointMat}) by introducing a differential operator $\mathbf{D}$. The regularized problem then becomes 
\begin{equation}\label{eq:TikRegDiffmin}
    \min_{\dot{\mathbf{x}}} \| \mathbf{A}\dot{\mathbf{x}} - \hat{\mathbf{x}}\|_2^2 + \alpha\|\mathbf{D}\dot{\mathbf{x}}\|_2^2,
\end{equation}
where $\alpha$ is a non-negative parameter that controls the smoothness of the solution $\dot{\mathbf{x}}$. In (\ref{eq:TikRegDiffmin}), $\mathbf{D}$ is defined as
\begin{equation}
    \mathbf{D} = 
    \begin{bmatrix}
    \mathbf{I}, & \mathbf{D}_1, & \mathbf{D}_2
    \end{bmatrix}^T \in \R^{(3m-6) \times (m-1)},
\end{equation}
where $\mathbf{I}$ denotes the identity matrix, and $\mathbf{D}_1$ and $\mathbf{D}_2$ are the discrete first and second difference operators. The minimization problem (\ref{eq:TikRegDiffmin}) is convex, differentiable, and admits a closed-form solution. The $\alpha$ parameter modulates the balance between the fidelity to the data and the smoothness of the derivative. For a given value of $\alpha$, the optimal solution is given by
\begin{equation}
    \dot{\mathbf{x}}_{\alpha} = (\mathbf{A}^T\mathbf{A} + \alpha \mathbf{D}^T\mathbf{D})^{-1}\mathbf{A}^T \hat{\mathbf{x}}.
\end{equation}
We found that Tikhonov regularization differentiation was robust, computationally efficient and easy to implement. We remark that the accuracy of the state derivatives depends on the sample size $m$, sampling rate $\Delta t$ and level of noise $\sigma$ in the data. Assuming the Nyquist-Shannon sampling theorem~\cite{shannon1949communication} is satisfied and the noise is zero-mean independent and identically distributed, the midpoint integration yields a discrete approximation error (bias) $\sim \mathcal{O}(\Delta t^2)$ and noise amplification (variance) that scales as $\sim \mathcal{O}(\frac{\sigma^2}{\Delta t^2})$~\cite{spall2005introduction}. In our numerical experiments, we observed that the condition number of $\mathbf{A}$ scales linearly with the sample size; the larger the condition number, the larger the least squares error may be. Tikhonov regularization attempts to lower the condition number of  $\mathbf{A}$ and reduce the effect of noise by appropriately selecting $\alpha$. However, it introduces bias in the computed solutions \cite{Hansen2010}.

In our experience, the L-curve criterion for choosing $\alpha$ yields the best results. In the case of Tikhonov regularization, Regi{\'n}ska~\cite{Reginska1996} proves that the log-log L-curve is always strictly convex at its ends for $\alpha \leq \sigma_n$ and $\alpha \geq \sigma_1$ ($\sigma_1$ and $\sigma_n$ being the largest and smallest singular values of $\mathbf{A}$, respectively). Let $\mathbf{u}_i$, $i = 1,\dots,m-1$, denote the right singular vectors of $\mathbf{A}$ associated with the $i$th largest singular value of $\mathbf{A}$. The L-curve can be concave if $|\mathbf{u}_i^T \hat{\mathbf{x}}|$ are monotonic with respect to $i$ or are constant~\cite{Hansen1999}.

\begin{remark}
Numerical differentiation may produce inaccurate results around the boundaries of a function if boundary conditions are unknown. This is the case in the present application where the time derivatives of states are not known at the ends of the training time period. As such, we here compute the time derivatives over a time interval that is 5\% larger than the intended training time span (from each side), but use the state data and estimated time derivatives over the original training time span.
\end{remark}


\bibliography{mybibfile}

\end{document}